\documentclass[manuscript,screen]{acmart}
\usepackage[utf8]{inputenc}
\usepackage{longtable}
\usepackage{booktabs}
\usepackage[table,dvipsnames]{xcolor}
\usepackage{multirow}
\usepackage{caption}
\usepackage{tabularx} 

\usepackage{array}    
\usepackage{subcaption}
\usepackage{enumitem}
\usepackage{balance}
\usepackage{wrapfig}
\usepackage[most]{tcolorbox}

\usepackage{duckuments}
\usepackage{arydshln}
\usepackage{amsmath}
\usepackage{needspace}
\usepackage{float}
\usepackage{graphicx}
\newcolumntype{Y}{>{\centering\arraybackslash}p{.45cm}}

\AtBeginDocument{%
  \providecommand\BibTeX{{%
    \normalfont B\kern-0.5em{\scshape i\kern-0.25em b}\kern-0.8em\TeX}}}

\setcopyright{acmcopyright}
\copyrightyear{2025}
\acmYear{2025}
\acmDOI{}

\definecolor{gold}{rgb}{1.0, 0.88, 0.0}

\acmISBN{}

\begin{document}

\title{\textsc{TraceHiding}: Scalable Machine Unlearning for Mobility Data}

\author{Ali Faraji}
\orcid{0000-0002-2439-8493}
\affiliation{%
  \institution{York University}
  \city{Toronto}
  \state{Ontario}
  \country{Canada}}
\email{faraji@yorku.ca}

\author{Manos Papagelis}
\orcid{0000-0003-0138-2541}
\affiliation{%
  \institution{York University}
  \city{Toronto}
  \state{Ontario}
  \country{Canada}}
\email{papaggel@eecs.yorku.ca}

\renewcommand{\shortauthors}{Faraji \& Papagelis}

\begin{abstract}
This work introduces \textbf{\textsc{TraceHiding}}, a scalable, importance-aware \textbf{machine unlearning} framework for mobility trajectory data. Motivated by privacy regulations such as GDPR and CCPA granting users \textit{``the right to be forgotten,''} \textsc{TraceHiding} removes specified user trajectories from trained deep models without full retraining. It combines a \textbf{hierarchical data-driven importance scoring scheme} with \textbf{teacher–student distillation}. Importance scores—computed at \textit{token}, \textit{trajectory}, and \textit{user} levels from statistical properties (\textit{coverage diversity}, \textit{entropy}, \textit{length})—quantify each training sample’s impact, enabling targeted forgetting of high-impact data while preserving common patterns. The student model \textbf{retains knowledge} on remaining data and \textbf{unlearns} targeted trajectories through an importance-weighted loss that amplifies forgetting signals for unique samples and attenuates them for frequent ones. We validate on \textbf{Trajectory–User Linking (TUL)} tasks across three real-world higher-order mobility datasets (\textsc{HO-Rome}, \textsc{HO-Geolife}, \textsc{HO-NYC}) and multiple architectures (GRU, LSTM, BERT, ModernBERT, GCN-TULHOR), against strong unlearning baselines including SCRUB, \textsc{NegGrad}, \textsc{NegGrad+}, \textsc{Bad-T}, and \textsc{Finetuning}. Experiments under \textbf{uniform} and \textbf{targeted} user deletion show \textsc{TraceHiding}, especially its entropy-based variant, achieves superior \textbf{unlearning accuracy}, competitive \textbf{membership inference attack (MIA)} resilience, and up to \textbf{40× speedup} over retraining with minimal test accuracy loss. Results highlight robustness to adversarial deletion of high-information users and consistent performance across models. To our knowledge, this is \textbf{the first systematic study of machine unlearning for trajectory data}, providing a reproducible pipeline with public code and preprocessing tools. By integrating domain-aware importance estimation with efficient unlearning, \textsc{TraceHiding} advances privacy-preserving mobility analytics and offers a scalable blueprint for responsible AI deployment in spatiotemporal applications.

\end{abstract}



\received{21 September 2025}
\received[revised]{TBD}
\received[accepted]{TBD}

\maketitle

\section{Introduction}
\label{sec:introduction}

\subsection{Motivation}

The rapid proliferation of location-aware technologies, such as GPS navigation systems, mobile location-based services, and pervasive sensing platforms, has led to unprecedented volumes of trajectory data capturing the movement of individuals, vehicles, and other entities. These spatiotemporal traces are the backbone of numerous intelligent systems, enabling applications ranging from traffic optimization and urban planning to mobility recommendation and epidemic modeling. Yet, the same data that powers innovation also encodes sensitive ``\textit{who–was–where–when}'' information in model parameters, creating the potential for privacy violations, unauthorized behavioral profiling, or even the reproduction of location-bound copyrighted material \cite{Kanza2024GeospatialDataOwnership}.
Growing regulatory frameworks, such as the EU General Data Protection Regulation (GDPR~\cite{regulation2018art}) and the California Consumer Privacy Act (CCPA), enshrine the \textit{``right to be forgotten,''} requiring that individuals can request the removal of their personal data from learned models. \textbf{Machine unlearning} has emerged as a computational paradigm to meet this obligation: instead of retraining models from scratch, it seeks to selectively remove the influence of specific data points while preserving performance on the remaining dataset. However, unlearning research has predominantly focused on image, text, and tabular data, leaving mobility and trajectory domains underexplored, despite their high privacy sensitivity and unique structural properties.

\subsection{Problem of Interest, Limitations of Current State of the Art, and Challenges}

In this work, we focus on the problem of \textbf{machine unlearning for trajectory data} in the context of \textbf{trajectory classification} and, specifically, \textbf{Trajectory–User Linking (TUL)}. Here, given a sequence of spatiotemporal tokens representing a user’s movement history, the model predicts the identity of the user. This task underpins various mobility analytics capabilities, but also poses acute privacy risks: an accurate TUL model can re-identify individuals from anonymized traces. When a deletion request is issued, the challenge is to erase the statistical footprint of all trajectories associated with the target user without degrading the utility of the neural model on retained data, thus ensuring compliance with privacy regulations and preserving public trust in mobility-driven AI systems.
Existing unlearning methods fall broadly into \textbf{exact retraining}, \textbf{approximate unlearning}, and \textbf{analytic unlearning}. Exact retraining provides strong guarantees but is computationally prohibitive for large-scale models and datasets. Approximate methods, such as gradient reversal, teacher–student distillation, or noise-based parameter perturbation, offer efficiency but often ignore the heterogeneous influence of individual samples, leading to utility loss or incomplete forgetting. Analytic approaches, while efficient, apply only to restricted model classes and do not generalize to deep architectures recently used for trajectory mining.
In addition, trajectory data present \textbf{domain-specific challenges} absent in other modalities:
\begin{itemize}
    \item \textbf{Spatiotemporal correlation.} Removing a trajectory can perturb learned spatial and temporal patterns shared across multiple users.
    \item \textbf{Representation entanglement.} Internal embeddings often capture shared behavioral profiles, making it difficult to isolate and erase user-specific contributions.
    \item \textbf{Sequential dependencies.} Models capture movement patterns through non-linear, long-range temporal interactions, which complicates selective forgetting.
    \item \textbf{Scalability.} Real-world trajectory datasets are large, and naively adapting existing methods can lead to prohibitive runtimes.
\end{itemize}
These challenges mean that methods which treat all data points equally or ignore structural uniqueness are prone to either \textbf{over-forgetting} (damaging model utility) or \textbf{under-forgetting} (failing privacy requirements).

\subsection{Our Approach \& Contributions}

We present \textbf{\textsc{TraceHiding}}, the first importance-aware machine unlearning framework tailored to trajectory data. \textsc{TraceHiding} extends teacher–student unlearning architectures with a \textbf{hierarchical}, \textbf{data-driven importance scoring mechanism} that quantifies the influence of individual \textit{tokens}, \textit{trajectories}, and \textit{users}, independent of any specific model parameters. The unlearning process is then guided by these scores, selectively amplifying forgetting for highly unique or privacy-critical trajectories while preserving common patterns.
In brief, our main contributions are:
\begin{itemize}
    \item \textbf{Importance-Aware Unlearning Framework.} We introduce \textsc{TraceHiding}, which integrates hierarchical importance scores into the unlearning loss to target high-impact trajectories. An overview of the proposed algorithmic framework is shown in Figure~\ref{fig:unlearning_overview}.
    \item \textbf{Hierarchical Importance Scoring.} We define scalable, model-agnostic metrics at the token, trajectory, and user levels (e.g., coverage diversity, entropy, length, uniqueness) and unify them into normalized importance scores.
    \item \textbf{Teacher–Student Distillation with Importance Weighting.} We adapt the distillation process to simultaneously maximize forgetting on the unlearning set and retain knowledge on the remaining set, with per-sample weighting driven by importance scores.
    \item \textbf{Comprehensive Empirical Validation.} We conduct a comprehensive empirical evaluation across three large-scale trajectory datasets (\textsc{HO-Rome}, \textsc{HO-Geolife}, \textsc{HO-NYC}) and diverse architectures (\textsc{GRU}, \textsc{LSTM}, \textsc{BERT}, \textsc{ModernBERT}, \textsc{GCN-TULHOR}) and demonstrate that \textsc{TraceHiding} consistently outperforms state-of-the-art machine unlearning baselines (\textsc{NegGrad}, \textsc{NegGrad+}, \textsc{SCRUB}, \textsc{Bad-T}, \textsc{Finetuning}) in unlearning accuracy, utility preservation, and runtime efficiency.
    \item \textbf{Open-Source Resources.} We release source code, preprocessing pipelines, and benchmarks to support reproducibility and further research, as follows:
    \begin{tcolorbox}[colback=gray!10, colframe=black, title=\textsc{TraceHiding} Project Repository, sharp corners=south]
    Official source code of \textsc{TraceHiding} models:
    
    \smallskip
    
    \centering
    \href{https://github.com/alifa98/TraceHiding}{\texttt{https://github.com/alifa98/TraceHiding}}
    
    \smallskip
    
    This repository includes all relevant scripts, model definitions, data processing tools, and evaluation utilities used throughout the research.
    \end{tcolorbox}
    
    \begin{tcolorbox}[colback=gray!10, colframe=black, title=\textsc{Point2Hex}: Data Preprocessing Tools Repository, sharp corners=south]
    Official source code of relevant trajectory data representations:
    
    \smallskip
    
    \centering
    \href{https://github.com/alifa98/point2hex}{\texttt{https://github.com/alifa98/point2hex}}
    
    \smallskip
    
    This companion repository includes tools for processing and converting raw mobility data into higher-order representations, including map matching, point-to-hexagon conversion, and visualization.
    \end{tcolorbox}
\end{itemize}

\subsection{Paper Organization}

The remainder of this paper is organized as follows. Section~\ref{sec:related-work} reviews relevant literature in trajectory mining and machine unlearning. Section~\ref{sec:data-representation} introduces our data representation, tokenization scheme, and the formulation of the TUL task. Section~\ref{sec:problem} formalizes the machine unlearning problem in the context of trajectory classification. Section~\ref{sec:methodology} presents the proposed \textsc{TraceHiding} framework, detailing the hierarchical importance score computation, the teacher–student architecture, and the design of the loss function. Section~\ref{sec:experiments} outlines the experimental setup, datasets, and evaluation metrics, and reports the results along with comparative analyses and ablation studies. Finally, Section~\ref{sec:conclusion} concludes the paper with a summary of contributions and potential directions for future research.

\section{Related Work}
\label{sec:related-work}

We review related work on \textit{trajectory classification} and \textit{machine unlearning}, and then \textit{position our contribution}.

\subsection{Trajectory Classification}

Trajectory data classification involves analyzing spatiotemporal sequences to uncover patterns, make predictions, or associate trajectories with users. Among its many applications, such as intelligent transportation systems \cite{arasteh2022network, nematichari2022evaluating}, urban planning \cite{sawas2019versatile, pechlivanoglou2019efficient, mehmood2020learning}, environmental monitoring, and public health \cite{pechlivanoglou2022epidemic, pechlivanoglou2022microscopic, yanin2023optimal, alix2022mobility}, privacy concerns have grown due to the identifiable nature of movement data \cite{Kanza2024GeospatialDataOwnership}.
Deep learning methods have become the dominant approach for trajectory analysis \cite{wang2020deep, atluri2018spatio, zheng2015trajectory}. Recurrent neural networks (RNNs), particularly Long Short-Term Memory (LSTM) networks, have proven effective for capturing temporal dependencies in trajectory sequences \cite{si2022pedestrain,zhang2018spatial}. More recent approaches leverage attention mechanisms and transformer architectures to capture long-range dependencies in trajectory sequences, enabling better understanding of movement patterns across different temporal scales \cite{nadiri2025trajlearn}.
The complexity of trajectory data presents unique challenges for neural architectures. Unlike traditional time series, trajectories embed both spatial and temporal information, requiring models to understand geographical constraints, road networks, and human mobility patterns. Graph neural networks (GNNs) have emerged as particularly suitable for trajectory analysis, as they can naturally represent the underlying road network structure and capture spatial relationships between different locations \cite{xu2022adaptive}.
Several efforts focus on preprocessing and analyzing trajectories, including trajectory similarity and prediction \cite{xue2022auxmoblcast, nadiri2025trajlearn}, clustering \cite{han2022clustering}, classification \cite{miao2020tul, mahmoud2023tul}, simplification \cite{wang2021simplification, alix2023pathletrl, alix2024pathletrlplusplus}, outlier detection \cite{koetsier2022anomaly}, and imputation \cite{musleh2023kamel}. These preprocessing steps often involve feature engineering to extract meaningful representations from raw GPS coordinates, including speed profiles, turning angles, stop detection, and semantic location information.
In particular, \emph{trajectory-user linking} (TUL) attempts to associate anonymous trajectories with known individuals, raising serious data governance challenges \cite{miao2020tul, mahmoud2023tul}. TUL methods typically rely on extracting unique behavioral signatures from movement patterns, such as frequently visited locations, travel time preferences, and route choices. The effectiveness of these methods in identifying individuals from seemingly anonymous data demonstrates the inherent privacy risks in trajectory data, making the need for effective unlearning techniques particularly acute \cite{Kanza2024GeospatialDataOwnership}.

\subsection{Machine Unlearning: Concepts and Approaches}

Machine unlearning is the task of removing the influence of specific training data from a trained model. We review existing unlearning approaches including \emph{exact}, \emph{approximate}, and \emph{analytic}, along with unlearning evaluation methods.

\subsubsection{Exact Unlearning}

Exact unlearning involves retraining a model from scratch after removing the data to be forgotten. This provides theoretical guarantees but is computationally expensive and impractical for large-scale models.
One well-known framework is SISA (Sharded, Isolated, Sliced, and Aggregated) training by Bourtoule et al. \cite{bourtoule2021machine}, which organizes training into independent shards to facilitate selective retraining. SISA reduces unlearning costs by limiting retraining to affected shards, but requires careful design of data partitioning strategies. The method faces particular challenges with trajectory data, where temporal dependencies and spatial correlations make clean partitioning difficult.
To address exact unlearning in graph settings, Chen et al. introduced \textit{GraphEraser} \cite{Min2022GraphUnlearning}, which uses graph-aware partitioning and aggregation techniques. GraphEraser employs community detection algorithms to create balanced shards while preserving graph structure, and introduces learning-based aggregation to combine predictions from multiple shard models. While promising for graph data, the approach assumes clear structural boundaries that may not exist in trajectory datasets, where individual trajectories may span multiple geographical regions or time periods.

\subsubsection{Approximate Unlearning}

Approximate unlearning avoids full retraining by estimating how to remove data influence, trading precision for efficiency. A useful taxonomy divides this into \textbf{strong unlearning}, which aims to make parameters match the retrained model, and \textbf{weak unlearning}, which focuses on adjusting outputs to remove memorized traces \cite{xu2023machine, xu2024machine}. We next summarize representative methods for approximate unlearning.

\begin{itemize}
        \item \textit{Parameter Perturbation Methods}. Several methods directly modify model parameters to achieve unlearning. Foster et al. \cite{foster2024fast} propose Selective Synaptic Dampening (SSD), which uses the Fisher Information Matrix to identify parameters most important to the forget set and dampens them proportionally. This retrain-free approach offers significant speedup but may struggle with highly correlated features common in trajectory data, where spatial and temporal features are inherently interdependent. Golatkar et al. \cite{golatkar2020eternal} introduce a scrubbing approach that leverages the stability of stochastic gradient descent to selectively remove information. Their method uses a Forgetting Lagrangian that balances retained data performance with information removal.
        \item \textit{Gradient-Based and Noise Injection Methods}. Tarun et al. \cite{tarun2023fast} propose error-maximizing noise generation for unlearning, which efficiently removes influence without revisiting the full training dataset. Their framework scales well to multiple classes but relies on the assumption that adding targeted noise can mask learned patterns. For trajectory data, this approach faces challenges because movement patterns involve complex spatiotemporal correlations that may be difficult to mask with simple noise injection. Huang et al. \cite{huang2024unified} present a unified gradient-based framework that decomposes updates into forgetting, retaining, and saliency-modulated components. They introduce a fast-slow parameter update strategy for efficient Hessian approximation, making the method more scalable. However, the framework's effectiveness depends on accurately identifying which parameters encode information about specific data points.
        \item \textit{Structure-Aware and Sparsity-Based Methods}. Jia et al. \cite{jia2023sparsification} explore model sparsification as a pathway to simplify unlearning. Their approach first induces sparsity through pruning, then applies unlearning techniques to the sparse model. While this reduces computational complexity, it may be problematic for trajectory models where important spatiotemporal relationships might be encoded in connections that appear less critical during pruning.Zhao et al. \cite{zhao2024what} investigate factors that influence unlearning difficulty, identifying entanglement between forget and retain sets and the degree of memorization as key challenges. Their Refined-Unlearning Meta-algorithm (RUM) addresses these by refining forget sets into homogeneous subsets.    %
        \item \textit{Teacher-Student Architectures}. A related line of work leverages \textbf{teacher-student architectures} for controlled forgetting. These methods offer promising directions for trajectory unlearning due to their ability to selectively transfer knowledge while filtering out sensitive information. Chundawat et al. \cite{chundawat2023can} propose using two competent and incompetent teachers in a knowledge distillation framework to induce selective forgetting. Their method supports various granularities of forgetting, from single classes to random subsets, making it potentially suitable for trajectory applications where forgetting requirements might vary from individual trajectories to entire user patterns. SCRUB \cite{kurmanji2024towards} introduces a contrastive optimization objective within the teacher-student framework, using a rewinding procedure to fine-tune the forgetting process. The method's ability to maintain utility while achieving high forget quality makes it attractive for trajectory applications, where preserving general mobility patterns while forgetting specific user traces is crucial. Chundawat et al. \cite{chundawat2023zero} extend this work to zero-shot scenarios where no original training data is available. Their gated knowledge transfer (GKT) framework filters out information related to the forget class during knowledge transfer, which could be adapted for trajectory scenarios where access to original training data is restricted due to privacy regulations. 
\end{itemize}
Our method, \textsc{TraceHiding}, extends the teacher–student architecture to a novel weak unlearning framework for trajectory data. It incorporates a hierarchical importance-scoring scheme to selectively remove sensitive trajectories while preserving model utility and frequent patterns.

\subsubsection{Analytic Unlearning}

Analytic unlearning techniques rely on models with closed-form solutions. For example, Cao and Yang \cite{Cao2015ForgetWithUnlearning} proposed transforming models into summation form, enabling deletion by subtracting cached statistics. Similarly, Brophy and Lowd \cite{pmlr-v139-brophy21a} introduced DaRE forests, allowing local updates within decision trees.
These methods are efficient and exact but apply only to limited model classes. They are not compatible with neural architectures commonly used for trajectory classification, and thus are not suitable for our setting.

\medskip\noindent\textbf{Evaluation of Machine Unlearning}. The evaluation of machine unlearning methods is multifaceted, typically revolving around three key aspects: the model's remaining utility, the effectiveness of the forgetting process, and the computational efficiency of the method.
Early work by Cao et al.~\cite{Cao2015ForgetWithUnlearning} established the foundational metrics of \textbf{completeness} and \textbf{timeliness}. Completeness requires the unlearned model to be indistinguishable from a model retrained from scratch without the data-to-be-forgotten, while timeliness measures the speed of the unlearning process. The concept of completeness underpins many subsequent forgetting metrics. The SISA framework \cite{bourtoule2021machine} is evaluated on \textbf{time to unlearn}, \textbf{model accuracy}, and the resulting accuracy degradation, explicitly balancing efficiency and utility. Similarly, Chen et al.~\cite{Min2022GraphUnlearning} focus on \textbf{Unlearning Efficiency} (speed relative to retraining) and \textbf{Model Utility} (measured by the Micro F1 Score).
More recent works have proposed sophisticated metrics for evaluating forgetting, especially for privacy. Jia et al.~\cite{jia2023sparsification} provide a holistic evaluation framework including \textbf{Unlearning Accuracy (UA)} on the forgotten set, \textbf{Remaining Accuracy (RA)} on the retained set, and, critically, the efficacy against \textbf{Membership Inference Attacks (MIA)} to quantify privacy. To circumvent the high computational cost of retraining for comparison, novel proxy metrics have been introduced. Chundawat et al.~\cite{chundawat2023can} proposed Zero Retrain Forgetting (ZRF), which measures the randomness of the unlearned model's predictions on the forgotten data. In a similar vein, the Anamnesis Index (AIN)~\cite{chundawat2023zero} evaluates forgetting by measuring the time it takes for an unlearned model to relearn the forgotten information, where a longer relearning time signifies more effective unlearning.

\subsection{Research Gap and Literature Positioning}

Existing research has advanced trajectory classification and machine unlearning independently, \textbf{yet their intersection remains unaddressed}. While trajectory mining reveals spatiotemporal patterns and unlearning techniques enable data removal, no prior work considers how to forget trajectory data in neural models. This omission is critical, as trajectory data introduces correlated dependencies, varying privacy requirements, and the need to preserve aggregate patterns essential for applications such as traffic prediction.  
\textsc{TraceHiding} directly fills this gap. By modeling the hierarchical structure of trajectories and applying importance scoring, it provides a weak approximate unlearning framework that balances forgetting accuracy with pattern preservation. In doing so, it extends machine unlearning to the mobility domain and positions itself as the first approach to make trajectory-based learning simultaneously privacy-compliant and utility-preserving.
\section{Trajectory Representation and Task Formulation}
\label{sec:data-representation}

This section explains how raw spatiotemporal trajectory data are processed into a representation suitable for machine learning tasks, with a focus on trajectory classification. We first introduce key concepts, then describe the tokenization method we use.
Finally, we formally define the main predictive task studied in this work: \emph{Trajectory-User Linking} (TUL).

\subsection{Trajectory Representation}
\label{sec:trajectory-representation}

We begin with the standard definition of a raw spatiotemporal trajectory.

\begin{definition}
\label{def:st_trajectory}
\textbf{(Spatiotemporal Trajectory \(T_c\))}
A trajectory is a \emph{temporally} ordered sequence of spatiotemporal points \(p\) that belong to a user (class) \(c \in C\). Each point is a tuple \((x, t)\), where \(x \in \mathbb{R}^2\) represents a geographic location (e.g., GPS coordinates) and \(t \in \mathbb{R}_{\geq 0}\) is the timestamp. Formally,
\[
T_c = \big( (x_1, t_1), (x_2, t_2), \dots, (x_l, t_l) \big), \quad \text{with } t_1 < t_2 < \dots < t_l.
\]
\end{definition}
\noindent Raw spatiotemporal trajectories pose challenges for direct use in deep learning models. They are often noisy, high-dimensional, and irregularly sampled in space and time. To create representations that models can use, researchers often apply discretization or tokenization techniques.
For our experiments, we use the recently introduced \textbf{\textsc{Point2Hex}} dataset~\cite{faraji2023point}. \textsc{Point2Hex} provides higher-order mobility flow data, representing trajectories as sequences of hexagons on a hexagon-tessellated map. 
In brief, the GPS point sequences from the raw trajectory dataset are transformed into tokens through a two-step process:
\begin{enumerate}
    \item \textbf{Map Tessellation:} Partition the continuous spatial domain \(\mathbb{R}^2\) into a finite set of non-overlapping spatial cells, referred to as \emph{hexagons} or \emph{blocks}, denoted by the set \(\mathcal{B}\). Hexagonal tessellations provide a uniform, isotropic grid that minimizes distance bias and edge effects, improving spatial analysis accuracy compared to square grids.
    \item \textbf{Mapping GPS points to cells:} Assign each GPS point \((x_i, t_i)\) in \(T_c\) to the ID of the cell \(b \in \mathcal{B}\) such that \(x_i \in b\). The resulting cell ID becomes a token in the transformed sequence.
\end{enumerate}
\begin{figure}[t]
    \centering
    \includegraphics[width=\linewidth]{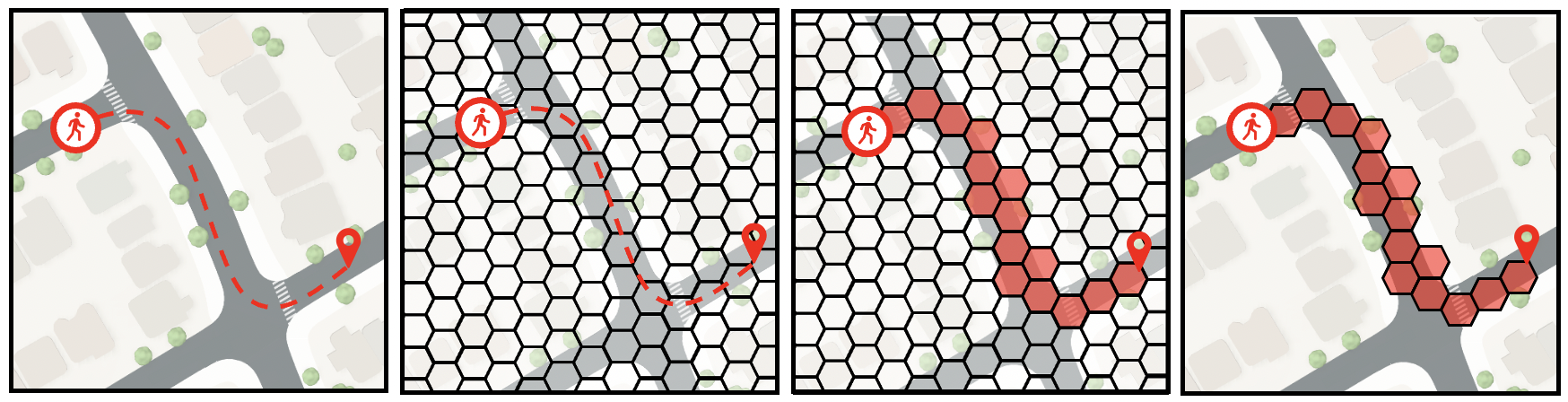}
    \caption[Path over hex grid; each hexagon becomes a token]{Visualization of the trajectory tokenization process. From left to right: (1) A pedestrian trajectory is first defined in continuous geographic space, following roads and crossings between the origin (walker symbol) and the destination (pin symbol). (2) A hexagonal tiling is overlaid on the map, enabling spatial discretization of the continuous trajectory. (3) The original path is projected onto the hexagonal grid, where the trajectory is represented as a sequence of traversed hexagonal cells. (4) The final trajectory representation emerges as a discrete path encoded by a connected chain of hexagons (tokens), which preserves the shape and direction of the original trajectory while enabling efficient tokenization.}
    \label{fig:hexagon_traj}
\end{figure}
\begin{figure}[t]
\includegraphics[width=\textwidth]{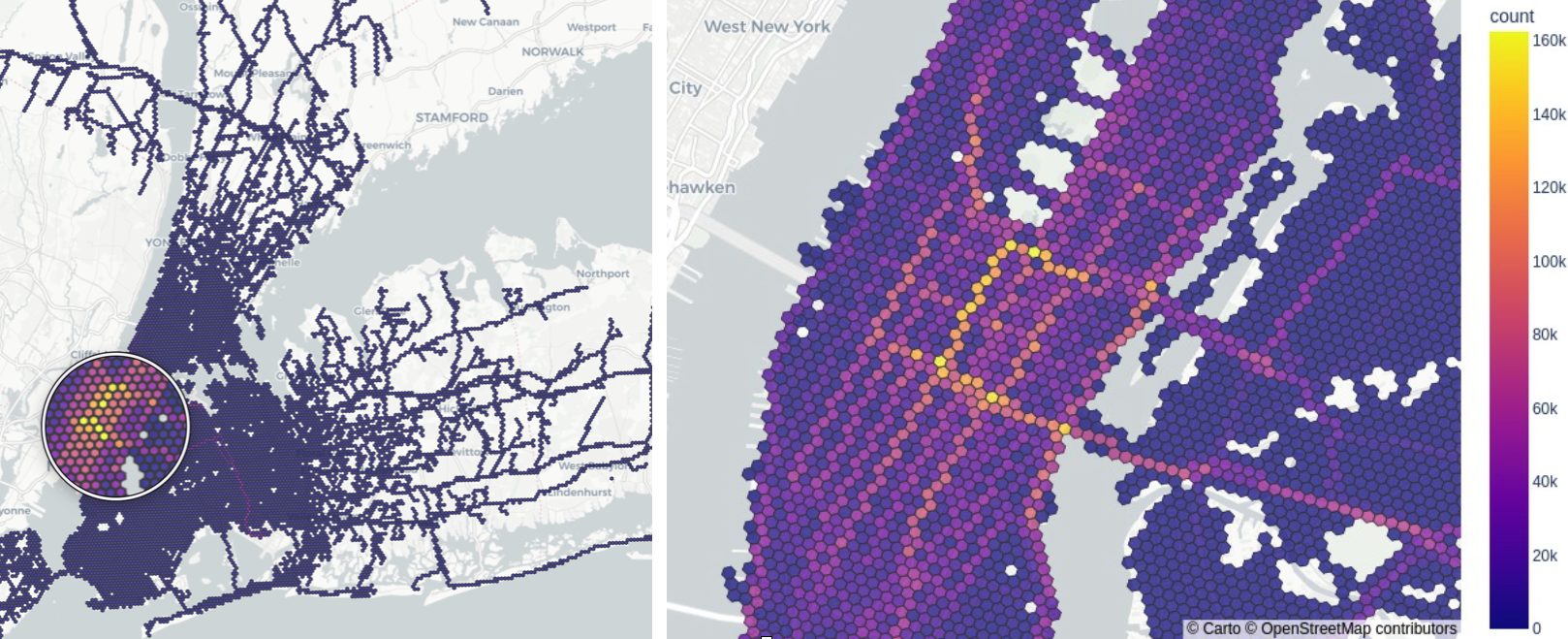} 
 \caption[Heatmap of hexagon-sequence paths on tessellated map]{A heatmap visualization using hexagonal tessellation to represent trajectory counts across the map. Each hexagon aggregates data within its area, highlighting spatial variations in hit rates. In this figure, denser regions such as central Manhattan show significantly higher counts (yellow) compared to peripheral or suburban areas with lower activity (purple/blue).}
\label{fig:traj_nyc} 
\end{figure}
Figure~\ref{fig:hexagon_traj} illustrates how a continuous path is transformed into a sequence of discrete tokens (i.e., hexagons). This transformation enables us to redefine a trajectory as a sequence over a finite vocabulary \(\mathcal{B}\), suitable for  modeling. An example of a tessellated map along with a trajectory heatmap is shown in Figure~\ref{fig:traj_nyc}.

\begin{definition}
\label{def:ho_trajectory}
\textbf{(Higher-Order Trajectory \(\mathcal{T}_c\))}
For user \(c \in C\), a higher-order trajectory \(\mathcal{T}_c\) is a \emph{temporally} ordered sequence of discrete tokens from \(\mathcal{B}\), where each token represents a visited region (e.g., a hexagonal cell). Formally,
\[
\mathcal{T}_c = (b_1, b_2, \dots, b_l), \quad \text{where } b_i \in \mathcal{B},\ \forall i,\ \text{and } \mathcal{T}_c \in \mathcal{B}^*.
\]
\end{definition}

\noindent The same idea applies to \textbf{higher-order check-in datasets}. For example, check-ins at points of interest in location-based social networks can be mapped to discrete regions. Each check-in (or its enclosing region) becomes a token, and the time-ordered list of tokens forms the trajectory \(\mathcal{T}_c\).
Mapping raw spatiotemporal data to tokens through hexagonal tessellation yields a compact, noise-resilient representation that standardizes trajectories despite irregular sampling and generalizes across diverse mobility data. In our study, we use the following higher-order mobility flow datasets:  
\begin{itemize}
    \item \textbf{\textsc{Ho-Rome}} \cite{bracciale2014crawdadrome}: Recorded traces of taxi cabs in Rome, Italy.
    \item \textbf{\textsc{Ho-Geolife}} \cite{zheng2008geolife, zheng2009geolife, zheng2010geolife}: GPS traces collected from multiple users during outdoor activities, covering a total distance of $\sim$1.2M km.
    \item \textbf{\textsc{Ho-Foursquare-NYC}} \cite{yang2014foursquare}: Points of interest (\textsc{PoI}) check-ins recorded by mobile devices in New York City.
\end{itemize}
Table~\ref{tab:ho-dataset-stats} summarizes the descriptive statistics of these datasets.
\begin{table}[t]
    \centering
    \caption{Statistics of the higher-order mobility flow datasets.}
    \label{tab:ho-dataset-stats}
      \begin{tabular*}{\linewidth}{@{\extracolsep{\fill}} l r r l c c @{}}
        \toprule
        \textbf{\textsc{Dataset}} & \textbf{\# \textsc{Users}} & \textbf{\# \textsc{Trajectories}} & \textsc{Time Period} & \textbf{\textsc{Type}} & \textbf{\textsc{Resolution}}  \\
        \midrule
        \textbf{\textsc{Ho-Rome}} & 315 & 5,873 & 02/01/14 -- 03/02/14 &  \textsc{Gps} traces & 8 \\
        \textbf{\textsc{Ho-Geolife}} & 57 & 2,100 & 04/01/07 -- 10/31/11 & \textsc{Gps} traces & 8 \\
        \textbf{\textsc{Ho-NYC}} & 1,083 & 49,983 & 04/12/12 -- 02/16/13 & Visits & 9 \\
      \bottomrule 
    \end{tabular*}
\end{table}

\subsection{Trajectory-User Linking (TUL) Task}
\label{sec:tul_task}

Using the higher-order representation (Definition~\ref{def:ho_trajectory}), we define a key task in mobility analytics and privacy research. A fundamental challenge in this domain is the ability to re-identify users from anonymized or abstracted movement data. This motivates the following task.

\medskip\noindent \textbf{Task Definition: Trajectory-User Linking (TUL).}
Let \(\mathcal{T} = (t_1, t_2, \dots, t_l)\) be an observed trajectory, where each \(t_i\) denotes a token in the trajectory (as defined in Definition~\ref{def:ho_trajectory}), and let \(C\) be a finite set of known users (or classes). The goal of the \emph{Trajectory-User Linking (TUL)} task is to identify the user (or class) \(c \in C\) who most likely generated \(\mathcal{T}\), based on a learned function \(M\).
Formally, we define the model
\[
M : \mathcal{B}^* \to \Delta(C) ,
\]
where \(\Delta(C)\) is the probability simplex over \(C\), i.e., the set of all probability distributions over users. The model \(M\) is trained from a dataset of known trajectory-user pairs \(\{(\mathcal{T}, c)\}_{c \in C}\), and the output \(M(\mathcal{T})\) represents the model's belief over the identity of the user who generated \(\mathcal{T}\).
The predicted user is then given by:
\[
\hat{c} = \arg\max_{c \in C} M(\mathcal{T})_c ,
\]
where \(M(\mathcal{T})_c\) denotes the probability assigned to user \(c\) by the model \(M\).

\section{The Problem}
\label{sec:problem}

In this section, we establish the mathematical foundations of the paper, introducing the core terminology, symbols, and conventions, with a complete glossary provided in Table~\ref{tab:notations}. We then formalize the central problem, specifying the inputs, desired outputs, and key structural or computational constraints.

\begin{definition}
\textbf{(Training Dataset \(\mathcal{D}_t\))}
The training dataset is the collection of higher-order trajectory data used to train a machine learning model for trajectory classification. It consists of sequences \(\mathcal{T}\), each of varying length \(l\), representing trajectories associated with a user \(c\).
Formally, 
\[
\mathcal{D}_t = \bigcup_{c \in C} \mathcal{T}_c, \quad \text{where } \mathcal{T}_c \subseteq \mathcal{B}^* .
\]
\end{definition}

\noindent Training dataset serves as the foundation for the model's learning process, enabling it to adjust its parameters to accurately classify trajectories before any unlearning is applied.

\begin{definition}
\textbf{(Forgetting / Unlearning Dataset \(\mathcal{D}_u\))}
The forgetting/unlearning dataset consists of specific higher-order trajectory data targeted for removal from the machine learning model's knowledge. This dataset is a subset of the training dataset (\(\mathcal{D}_u \subseteq \mathcal{D}_t\)) and includes trajectories associated with the corresponding users.
\end{definition}

\noindent The terms ``Unlearning Dataset'' and ``Forgetting Dataset'' can be used interchangeably. We mostly use the term ``Unlearning Dataset'' throughout this work.

\begin{definition}
\textbf{(Learning Process \(\mathcal{A}\))}
The learning process \(\mathcal{A}(\cdot)\) is the mechanism through which a machine learning model is trained on a trajectory dataset.
Formally, it is a function
\[
\mathcal{A} : \mathcal{P}(\mathcal{B}^*) \to \mathcal{F}, \quad \mathcal{A}(\mathcal{D}_t) = M_t,
\]
where \(\mathcal{P}(\mathcal{B}^*)\) denotes the power set of finite sequences over the token set \(\mathcal{B}\), and \(\mathcal{F}\) is the space of models (functions) mapping trajectories to user identities:
\[
M_t : \mathcal{B}^* \to C.
\]
The model is parameterized by weights \(W_t\), optimized during training for trajectory classification.
\end{definition}

\begin{definition}
\textbf{(Remaining Dataset \(\mathcal{D}_r\))}
The remaining dataset is the portion of the original trajectory training dataset that remains after the unlearning dataset has been removed. It represents the trajectory data on which the model should retain its knowledge and continue to perform classification accurately.
Formally,
\[
\mathcal{D}_r = \mathcal{D}_t \setminus \mathcal{D}_u.
\]
If a model is retrained on this remaining dataset using the learning process, the resulting model is
\[
M_r = \mathcal{A}(\mathcal{D}_r),
\]
with corresponding parameters \(W_r\).
\end{definition}

\begin{table}[t]
    \centering
    \begin{tabular}{c p{0.85\linewidth}}
        \toprule[1.2pt]
         \textbf{Symbol} & \textbf{Description}  \\
         \midrule
         \(C\) & Set of all users \\
         \(\mathcal{T}_c\) & A trajectory that belongs to the user \(c \in C\) \\
         \(l\) & Trajectory length \\
         \(\mathcal{D}_t\) & Training dataset \\
         \(\mathcal{D}_t^c\) & Set of all trajectories in \(\mathcal{D}_t\) that belong to user \(c\) \\
         \(\mathcal{D}_u\) & Unlearning/Forgetting dataset \\
         \(\mathcal{D}_r\) & Remaining dataset \\
         \(M_t\) & Originally trained model on \(\mathcal{D}_t\) \\
         \(M_u\) & Unlearned model \\
         \(M_r\) & Trained model on \(\mathcal{D}_r\) \\
         \(W_t\) & Weights of \(M_t\) trained on \(\mathcal{D}_t\) \\
         \(W_u\) & Weights of unlearned model \(M_u\) \\
         \(W_r\) & Weights of \(M_r\) trained on \(\mathcal{D}_r\) \\
         \(\mathcal{A}(\cdot)\) & Learning process \\
         \(\mathcal{U}(\cdot)\) & Unlearning process \\
                  \(\mathcal{B}\) & Set of all tokens (block) in the dataset \\
         \(\mathcal{B}^*\) & Set of all finite-length sequences over the token set \(\mathcal{B}\), i.e., \(\mathcal{B}^* = \bigcup_{l=1}^{\infty} \mathcal{B}^l\) \\
         \bottomrule[1.2pt]
    \end{tabular}
    \caption{Summary of notations.}
    \label{tab:notations}
\end{table}

\noindent
\textbf{Problem Statement of Unlearning in Trajectory Classification (\(\mathcal{U}\)).}
The unlearning process \(\mathcal{U}(\cdot)\) is designed to remove the influence of the unlearning dataset, which consists of specific trajectory sequences, from a trained model.
Formally, it is a function
\[
\mathcal{U} : \mathcal{F} \times \mathcal{P}(\mathcal{B}^*) \times \mathcal{P}(\mathcal{B}^*) \to \mathcal{F}, \quad \mathcal{U}(M_t, \mathcal{D}_t, \mathcal{D}_u) = M_u,
\]
where \(M_u : \mathcal{B}^* \to C\) is the resulting model after unlearning, and ideally satisfies:
\[
M_u \approx \mathcal{A}(\mathcal{D}_r).
\]
The model \(M_u\) has adjusted weights \(W_u\) that no longer retain knowledge of the unlearned trajectories.
Obviously, a trivial solution for this problem is to retrain the model \(M_t\) from scratch on the remaining data \(\mathcal{D}_r\), but we are trying to avoid full retraining as it is impractical in most of the large models. In practice, \(M_u\) and \(M_r\) may not be identical for every unlearning process.
\section{An Algorithmic Framework for Unlearning Trajectory Mobility Data}
\label{sec:methodology}

\begin{figure}[t]
    \centering
    \includegraphics[width=\linewidth]{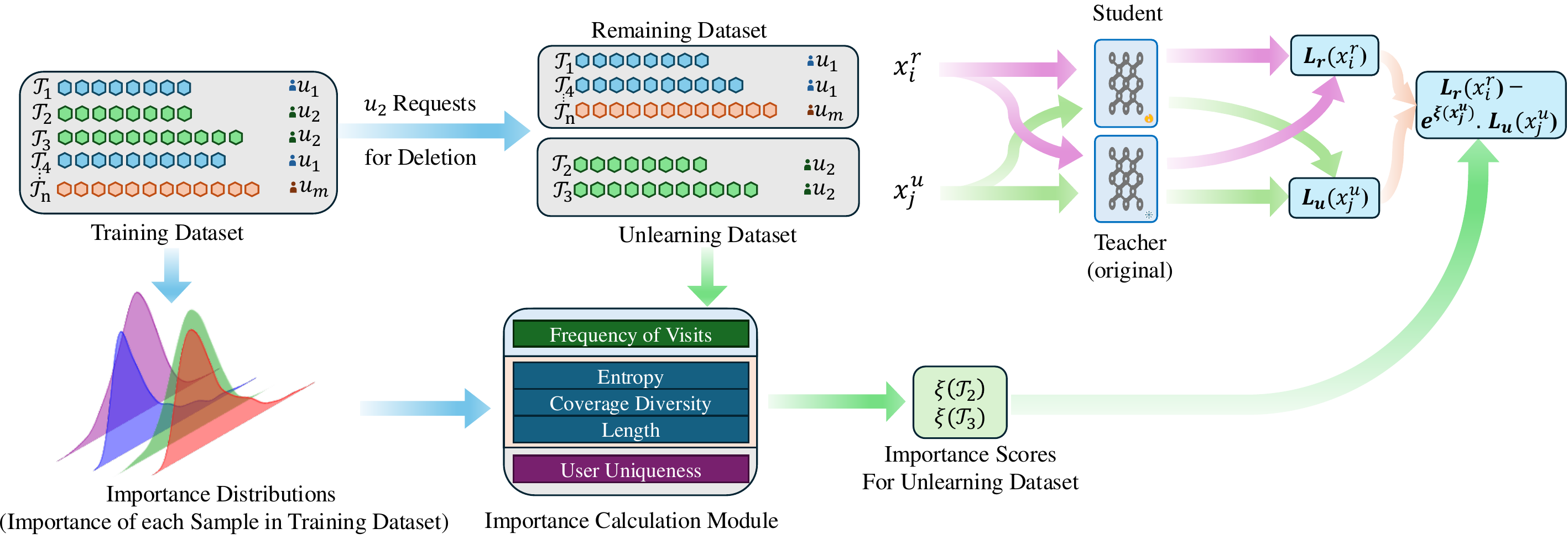}
    \caption[Overview of selective unlearning guided by teacher model using dual loss]{This diagram shows our unlearning process where the \textit{Student Model} is trained to forget data from the \textit{Unlearning Dataset} (\(x_j^u \in \mathcal{D}_u\)) while retaining knowledge from the \textit{Remaining Dataset} (\(x_i^r \in \mathcal{D}_r\)). The \textit{Teacher Model}  guides this process, with two loss functions: \(L_r\), ensuring retention of \(x_i^r\), and \(L_u\), promoting unlearning of \(x_j^u\). The combined loss directs gradient calculation and then backpropagation for \textit{student model}, allowing selective forgetting based on sample importance \(\xi(x)\).}
    \label{fig:unlearning_overview}
\end{figure}

Machine unlearning aims to remove the influence of specific data points from trained models while preserving overall utility. Popular approaches, such as \textsc{SCRUB} \cite{kurmanji2024towards} and \textsc{Bad-T} \cite{chundawat2023can}, leverage teacher–student frameworks to achieve this goal. \textsc{SCRUB} trains a student model to maximize error on the unlearning dataset while minimizing it on the retained data, whereas \textsc{Bad-T} employs two teachers: a competent teacher to provide correct information for retained data and an incompetent teacher to induce forgetting by introducing noise into the unlearning set. While effective for traditional supervised learning tasks, \textbf{these methods assume that all data points contribute equally}, which limits their applicability to more complex domains.
Trajectory datasets, in particular, contain rich spatiotemporal dependencies where certain trajectories, or even sub-trajectories, can have disproportionately high influence on model predictions. Treating all data points uniformly can lead to \textbf{inefficient forgetting}, either overfitting to unimportant patterns or discarding critical structures in the data. To overcome this limitation, we introduce a novel \textbf{trajectory-level importance scoring mechanism} that fundamentally extends the frameworks of Kurmanji et al. \cite{kurmanji2024towards} and Chundawat et al. \cite{chundawat2023can}, enabling adaptive control over each trajectory’s contribution to the unlearning process and achieving more precise, efficient, and targeted forgetting.
By weighting trajectories according to their estimated influence, our approach enables \textbf{targeted unlearning} of high-impact or sensitive trajectories while better preserving global patterns and model utility. Our proposed algorithmic framework integrates \textbf{three tightly coupled components}: 
\begin{itemize}
    \item A data-driven importance scoring module.
    \item A teacher–student learning architecture. 
    \item A dedicated loss function that balances selective forgetting with predictive performance.
\end{itemize}
\noindent Figure~\ref{fig:unlearning_overview} illustrates the interaction between these components, and the remainder of this section details each in turn.

\subsection{Data-driven Importance Score (\(\xi\))}
We begin by assigning data points a scalar importance score \(\xi(x)\) that reflects their expected contribution to the learning objective. Rather than relying on ad-hoc heuristics, we exploit the wealth of \emph{historical} trajectory data to quantify this contribution. Specifically, these scores are computed based purely on properties of the data, such as visit frequency, spatial coverage, entropy, and user uniqueness, without reference to any specific model. By mining these statistics directly from the corpus, we derive importance scores that reflect real, observed influence patterns, grounded in the data distribution itself. We define the importance score \(\xi(x)\) at different levels of granularity (token, trajectory, and user) and then explain how these can be used to influence the unlearning process.

\begin{definition}[Importance Score \(\xi\)]\label{def:importance_score}
Let \(\mathcal{D}_t = \{x_i\}_{i=1}^n\) be the training dataset, where each \(x_i\) is a trajectory. An importance score \(\xi(x)\) function is a mapping
\[
\xi : \mathcal{D}_t \to \mathbb{R},
\]
that assigns each trajectory \(x_i \in \mathcal{D}_t\) a real-valued score \(\xi(x_i)\), reflecting its estimated influence on the learning process for the task of user linking.
\end{definition}

\noindent This score ultimately governs how each trajectory contributes to the unlearning objective, as defined in the loss function. While we compute importance at different structural levels (tokens, trajectories, users), all scores are ultimately applied at the trajectory level because the model’s input and loss functions operate on full trajectories.
This score serves as a proxy for the contribution of a data point to the performance or behavior of a model trained on the training dataset. The concept of importance in trajectory data can be analyzed at various \textbf{hierarchical levels}, including:
\begin{itemize}
    \item Token Level (e.g., Hexagon Level),
    \item Trajectory Level (across the entire trajectory), and
    \item User Level (aggregating multiple trajectories associated with a user).
\end{itemize}

\noindent These levels of importance are interrelated hierarchically. In cases where importance is defined at the user level, all trajectories associated with that user receive the same score. In cases where token-level scores are defined (e.g., token frequency), they are aggregated across the sequence to produce a trajectory-level score. In all cases, trajectory-level scores are used as inputs to the loss function as each sample (data point) is a trajectory.
To make scores from different criteria consistent and comparable, we apply a general \emph{min-max normalization} (defined in Equation~\ref{eq:normalized_xi}) after computing each raw score. This standardizes all scores to the \([0, 1]\) range. See Section~\ref{subsec:normalization} for the full normalization method.

\subsubsection{Token Level Importance} \label{subsec:importance_score_token}
At the finest granularity, we attach an importance score to each \textbf{token}, a small spatial segment along a trajectory, so that its local contribution to the training set is quantified. We introduce a token level importance measure based on token frequency, which counts how often a given token appears across all trajectories.

\smallskip\noindent \textbf{i) Frequency of Token}: Let \(f(b)\) denote the frequency of token \(b\),

\begin{equation}
    \label{eq:frequency_of_visits}
    f(b) = \bigm| \{ \mathcal{T} \in \mathcal{D}_t \mid b \in \mathcal{T} \} \bigm|.
\end{equation}
The importance of each token is defined as the inverse of its frequency, \(1/f(b)\), so that seldom-seen tokens are deemed more significant.
In the context of our trajectory datasets, \(f(b)\) can be interpreted as the number of visits to the corresponding hexagon, and these visit counts will later serve as building blocks for trajectory level importance.

\smallskip\noindent \textbf{Rationale.} 
Intuitively, locations that are visited rarely reveal atypical movement patterns and therefore carry more information about the diversity of the data than locations that are traversed routinely.

\subsubsection{Trajectory Level Importance} \label{subsec:importance_score_trajectory}
At the trajectory level, we calculate importance scores for each trajectory, considering the entire sequence of tokens as a single unit to evaluate its influence on the data distribution and its contribution to learning the dataset.
We consider three trajectory level importance metrics: 
(i) \textbf{Coverage Diversity}, 
(ii) \textbf{Information-theoretic importance} (entropy), and 
(iii) \textbf{Trajectory Length}.
At the end, we derive a unified trajectory level importance score that aggregates all of the trajectory level importance scores.

\smallskip\noindent \textbf{i) Coverage Diversity}: Let \(D(x)\) denote the number of unique blocks covered by trajectory \(x \in \mathcal{T}\):
\begin{equation}
    \label{eq:coverage_diversity}
    D(x) = \bigm| \left\{  b \mid b \in x  \right\} \bigm|.
\end{equation}

\noindent We can define the importance score \(\xi(x)\) related to the coverage of a trajectory \(x\) as follows:
\begin{equation}
    \label{eq:coverage_importance}
    \xi_{\text{coverage}}(x) = \frac{D(x)}{\lvert\mathcal{B}\rvert},
\end{equation}
where \(|\mathcal{B}|\) is the total number of unique blocks in the dataset.

\smallskip\noindent\textbf{Rationale.} 
A trajectory that visits many distinct blocks injects proportionally more spatial information and thus exerts a larger influence on the model’s parameters.

\smallskip\noindent \textbf{ii) Information-theoretic (Entropy-based Importance)}:  
We define the importance score \(\xi(x)\) of a trajectory \(x\) as the inverse of its entropy, calculated over its bigrams:
\begin{equation}
    \label{eq:entropy_importance}
    \xi_{\text{entropy}} (x) =  \frac{1}{H(x)},
\end{equation}
where he entropy \(H(x)\) of a trajectory \(x = (t_1, \dots, t_n)\) is defined as:
\begin{equation}
\label{eq:entropy}
    H(x) = - \sum_{b \in \text{bigram}(x)} p(b) \log_2 p(b),
\end{equation}
where \(p(b)\) is the empirical probability of bigram \(b\) occurring in trajectory \(x\):
\begin{equation}
\label{eq:probab_frequency}
    p(b) = \frac{\text{count}(b)}{\sum_{b' \in \text{bigram}(x)} \text{count}(b')}.
\end{equation}

\noindent We define the set of bigrams for a sequence \(x = t_1, t_2, \dots, t_n)\) as:
\begin{equation}
\label{eq:bigram}
    \text{bigram}(x) = \{ (t_i, t_{i+1}) \mid 1 \leq i < n \}.
\end{equation}

\smallskip\noindent \textbf{Rationale.} 
Entropy captures the unpredictability of a trajectory. A high-entropy sequence is more random and harder to model, implying it contributes less distinctive signal. Conversely, low-entropy sequences are more structured and provide clearer, more learnable patterns. We use the inverse of entropy to emphasize low-entropy (high-signal) trajectories in the unlearning process.
The rationale behind choosing bigram is that bigram entropy captures local sequential dependencies more effectively than unigram frequency. It strikes a balance between expressive structure and computational efficiency. Higher-order n-grams are computationally expensive and prone to sparsity.

\smallskip\noindent \textbf{iii) Trajectory Length}: Let \(L(x)\) denote the length of a trajectory \(x\), which is defined as the number of elements (or blocks) in the sequence,
\begin{equation}
    \label{eq:trajectory_length}
    L(x) = |x|.
\end{equation}

\noindent We can define the importance score \(\xi(x)\) related to the length of a trajectory \(x\) as follows:
\begin{equation}
    \label{eq:length_importance}
        \xi_{\text{length}}(x) = L(x),
\end{equation}
where, \(L(x)\) is the length of the trajectory \(x\).

\smallskip\noindent \textbf{Rationale.} 
In sequence-based models, every token in a trajectory contributes a gradient update; therefore, longer trajectories inject more total signal during training and wield greater influence over the learned parameters. By weighting trajectories in proportion to their length, we obtain a direct proxy for their cumulative impact on the model.

\smallskip\noindent\textbf{iv) Unified trajectory importance score}:
To evaluate the overall importance of a trajectory \(x \in \mathcal{D}_t\), we define a single unified importance score that combines the three individual scores: Coverage Diversity, Entropy, and Length. This score is computed as a weighted combination of the normalized individual scores:

\begin{equation}
    \label{eq:single_importance_score}
    \xi_{\text{unified}}(x) = \alpha \cdot \xi_{\text{coverage}}(x) + \beta \cdot \xi_{\text{entropy}}(x) + \gamma \cdot \xi_{\text{length}}(x),
\end{equation}
where the components are defined as follows:
\begin{itemize}
    \item \(\xi_{\text{coverage}}(x)\): the normalized (Equation~\ref{eq:normalized_xi}) importance score based on coverage diversity, as defined in Equation~\ref{eq:coverage_importance},
    \item \(\xi_{\text{entropy}}(x)\): the normalized (Equation~\ref{eq:normalized_xi}) importance score based on entropy, as defined in Equation~\ref{eq:entropy_importance},
    \item \(\xi_{\text{length}}(x)\): the normalized (Equation~\ref{eq:normalized_xi}) importance score based on trajectory length, as defined in Equation~\ref{eq:length_importance}.
\end{itemize}

\noindent The coefficients \(\alpha\), \(\beta\), and \(\gamma\) are non-negative weights that allow adjusting the relative contribution of each component to the overall score. These weights can be determined based on the specific application or dataset characteristics, with the condition
\begin{equation}
    \label{eq:weights_condition}
    \alpha + \beta + \gamma = 1.
\end{equation}

\smallskip\noindent \textbf{Rationale.} 
This single importance score provides a comprehensive measure of a trajectory's significance by considering its coverage of unique blocks, structural randomness (via entropy), and length. By combining these aspects, it captures both the diversity and complexity of the trajectory, as well as its overall scale, providing a holistic metric for evaluating its contribution to the dataset.
In Section~\ref{sec:unified_importance_weights}, we discuss how to define a unified version of importance and the method to weight each term within the unified version automatically.

\subsubsection{User Level Importance} \label{subsec:importance_score_user}
At the highest level, we determine the importance of each user by aggregating the contributions of all their trajectories. This provides insight into the overall influence of a user’s data on the model. By calculating user level importance through the aggregation of individual trajectory scores, we can capture a holistic view of the user’s impact on the dataset.
Various aggregation strategies, such as averaging, maximum selection, or median selection, offer flexibility to emphasize different aspects of a user's influence. Additionally, user level importance such as \textit{user Uniqueness}, help quantify the distinctiveness and contribution of a user’s data in comparison to others, shedding light on their unique role in the dataset.
We introduce two metrics to quantify a user’s contribution:
(i) \textbf{user uniqueness} and
(ii) \textbf{entropy-based influence}.
Analogous to the trajectory level scores, we subsequently merge these metrics into a single, unified user level importance score.

\smallskip\noindent\textbf{i) User Uniqueness}: User uniqueness is quantified based on the set of blocks in their trajectories that do not appear in the trajectories of any other user. Let \(\mathcal{B}_c = \left\{ b \in \mathcal{T} \;\middle|\; \mathcal{T} \in \mathcal{D}_t^c \right\}\), denoting the set of all blocks contained in all trajectories of user \(c\), and let \(\mathcal{B}_{-c} = \left\{ b \in \mathcal{T} \;\middle|\; \mathcal{T} \in \mathcal{D}_t^{c'}, \; \forall c' \neq c \right\}\), denoting the set of all blocks from the trajectories of all users other than \(c\). Then, the uniqueness score for user \(c\) is defined as
\[
    \text{Uniq}(c) = \left| \mathcal{B}_c \setminus \mathcal{B}_{-c} \right|.
\]
\noindent The importance score \(\xi(x)\) of a trajectory \(x\) which belongs to user \(c\) can then be defined as

\begin{equation}
    \label{eq:user_unique_importance}
\xi_{\text{unique}}(x) = \text{Uniq}(c).
\end{equation}

\smallskip\noindent \textbf{Rationale.} 
Trajectories that contain location blocks rarely (or never) visited by other users inject highly individual-specific signals into the model, skewing its parameters toward that user’s behaviour.  When the objective is \emph{machine unlearning}, erasing such distinctive data points is crucial: eliminating a user’s unique blocks most effectively retracts their influence while preserving population-level patterns contributed by common routes.  The uniqueness-based score therefore serves as a principled proxy for a trajectory’s marginal contribution, and by extension, for the impact its removal will have on the trained model.

\smallskip\noindent\textbf{ii) User Entropy-Based Influence:}  
To evaluate the randomness or predictability of a user’s trajectory set, we aggregate the entropy scores of their individual trajectories. The user level entropy importance is defined as the inverse of the aggregated entropy:
\begin{equation}
    \label{eq:user_entropy_importance}
    \xi_{\text{entropy}}(c) = \frac{1}{\sum_{x \in \mathcal{D}_t^c} H(x)},
\end{equation}
where \(H(x)\) is the entropy of trajectory \(x\), as defined in Equation~\ref{eq:entropy}. By summing the entropies of all trajectories for a user, we can assess how structured or random their overall trajectory set is, with lower aggregated entropy indicating higher importance.
The rationale is analogous to the information-theoretic score defined at the trajectory level.

\smallskip\noindent\textbf{iii) Unified user importance score}:
To evaluate a user's overall importance, we define a single unified user importance score, \(\xi(c)\), by combining the previously defined metrics: User Uniqueness and User Entropy-Based Influence.
This score is computed as a weighted linear combination:

\begin{equation}
    \label{eq:single_user_importance_score}
    \xi(c) = \eta \cdot \xi_{\text{unique}}(c) + \lambda \cdot \xi_{\text{entropy}}(c),
\end{equation}
where, \(\xi_{\text{unique}}(c)\) is the uniqueness of the user’s trajectory set, as defined in Equation~\ref{eq:user_unique_importance}, and \(\xi_{\text{entropy}}(c)\) is the user’s entropy-based importance, as defined in Equation~\ref{eq:user_entropy_importance}.
The coefficients \(\eta\) and \(\lambda\) are non-negative weights that control the relative contribution of each metric. These weights can be adjusted to reflect the emphasis on different aspects of user level importance, with the condition
\begin{equation}
    \label{eq:user_weights_condition}
    \eta + \lambda = 1.
\end{equation}

\noindent The unified user importance score offers a comprehensive evaluation of a user's overall contribution to the dataset, combining aspects of their data’s uniqueness and structural randomness. By adjusting the weights, \(\xi(c)\) can be tailored to highlight specific facets of user influence in the dataset.

\subsubsection{Normalization}
\label{subsec:normalization}
There are different ways of defining the importance score \(\xi(x)\), and we explored some options in Sections~\ref{subsec:importance_score_token}, \ref{subsec:importance_score_trajectory}, and \ref{subsec:importance_score_user}.
To make the importance score \(\xi(x)\) consistent and comparable across different score criteria, we normalize it using a min-max normalization technique.
Min-max normalization ensures that the scores lie within a standardized range, typically between 0 and 1, and can be computed as follows:
\begin{equation}
    \label{eq:normalized_xi}
    \xi_{norm} = \frac{\xi - \xi_{min}}{\xi_{max} - \xi_{min}},
\end{equation}
where \(\xi_{min}\) and \(\xi_{max}\) are the minimum and maximum values of the raw importance score \(\xi\) in the dataset, respectively. By applying min-max normalization, the importance scores are rescaled proportionally within the range [0, 1], preserving the relative differences between data points while removing the influence of differing scales or units of the raw scores.
This normalization is particularly useful when the raw importance scores are derived from different metrics or measures with varying units or scales. It ensures that the resulting scores are dimensionless and comparable, making it easier to interpret and analyze the relative significance across different importance scores.

\subsection{Teacher-Student Model}
The unlearning process utilizes a teacher-student framework. The \textbf{Teacher model}, denoted as \(M_t\), is the original model that has been trained on the dataset \(\mathcal{D}_t\). The output of this model for a given input \(x\) is represented by \(M_t(x)\). During the unlearning procedure, the Teacher model's primary role is to provide reference outputs. Specifically, forward passes are performed on the Teacher model for batches of both the unlearning data \(\mathcal{D}_u\) and the remaining data \(\mathcal{D}_r\). The outputs generated by \(M_t\) for these data batches are then used to calculate the loss function (defined in Equation~\ref{eq:lossfn}), which guides the update of the Student model. Crucially, the weights of the Teacher model (\(M_t\)) remain frozen and are not changed throughout the entire unlearning procedure.
The \textbf{Student model}, denoted as \(M_u\), begins as an identical copy of the originally trained Teacher model \(M_t\) (also trained on \(\mathcal{D}_t\)). The fundamental distinction between the \textit{student} and the \textit{teacher} lies in their roles and behavior during the unlearning phase. While the Teacher model's weights are fixed, the Student model's weights are actively updated. This update mechanism relies on the loss calculated using the Teacher's outputs, which is then back-propagated through the Student model (\(M_u\)) to adjust its weights.
The purpose of this teacher-student interaction is \textbf{twofold}:
\textit{First}, by calculating a loss based on the Teacher's output for the unlearning data \(\mathcal{D}_u\) (specifically, using the negative of the standard loss as implied by the unlearning objective), we compel the Student model (\(M_u\)) to deviate from the original model's behavior on this data. This process guides the Student model to effectively ``forget'' these targeted samples.
\textit{Second}, by calculating a loss based on the Teacher's output for the remaining data \(\mathcal{D}_r\), we reinforce the Student model's ability to perform correctly on the data it is intended to retain. This step is vital to prevent the degradation of the model's accuracy on samples not designated for unlearning.

The loss function used for updating the Student model (as detailed in Equation~\ref{eq:lossfn}) incorporates an importance score (defined in Equation~\ref{eq:entropy_importance}) as a multiplicative factor for the loss values derived from the unlearning samples \(\mathcal{D}_u\). This score quantifies the importance of an input (e.g., a user trajectory). The rationale for including this factor is to avoid excessively forcing the model to forget information that is common or abundant within the dataset.
For instance, consider a scenario where an unlearning sample pertains to a user's trajectory that includes driving along a straight stretch of a major highway. This pattern, a straight path on a highway, is likely a very common occurrence within the broader dataset, representing a fundamental driving behavior the model should generally understand. If this common pattern were treated with the same unlearning intensity as a highly unique trajectory, the model might be excessively penalized for recognizing and predicting straight highway driving. This could lead to a degradation in its ability to handle other, unrelated instances of highway driving for users whose data is not being unlearned.
To mitigate this, the importance score comes into play. For such a common pattern (the straight highway path), its calculated importance score would be relatively low. Consequently, when this sample is processed during unlearning, its contribution to the unlearning loss (the signal telling the model to ``forget'') is dampened by this lower score. The model is still nudged to disassociate the specific unlearned trajectory from its memory, but it's not forced to aggressively unlearn the general concept of ``driving straight on a highway.''
This nuanced approach, therefore, serves a critical purpose: it helps prevent the unlearning process from inadvertently degrading the model's performance on general, non-unique knowledge that is widely represented in the data. By down-weighting the unlearning impact of common patterns, the system can more effectively focus its efforts on targeting and removing truly unique or sensitive data points for which unlearning is paramount, while preserving the model’s valuable learned representations of common scenarios. This ensures that the model remains robust and accurate for the vast majority of data it is expected to handle, even after specific information has been unlearned.

\subsection{Loss Function}

We denote the outputs of the \textit{Student model} and \textit{Teacher model} for an input trajectory \(x\) as \(M_u(x)\) and \(M_t(x)\), respectively.
The total loss for a sample \(x\) is defined as
\begin{equation}
    \label{eq:lossfn}
    \mathcal{L}(x) = 
    \begin{cases}
    L_r(x), & x \in \mathcal{D}_r \\
    - (e^{\xi_{\text{norm}}(x)} - 1) \cdot L_u(x), & x \in \mathcal{D}_u
    \end{cases}.
\end{equation}

\noindent This loss encourages the Student model to retain knowledge on \(\mathcal{D}_r\) while forgetting \(\mathcal{D}_u\), scaled by the normalized importance score.
The unlearning loss \(L_u(x)\) is defined as
\begin{equation}
    \label{eq:unlearning_loss}
    L_u(x) = D_{KL}(M_u(x) \parallel M_t(x)),
\end{equation}
and the remaining data loss \(L_r(x)\) is
\begin{equation}
    \label{eq:remaining_loss}
    L_r(x) = c_1 \cdot D_{KL}(M_u(x) \parallel M_t(x)) + c_2 \cdot \text{CE}(\mathbf{y}, M_u(x)),
\end{equation}
where \(c_1, c_2 \in \mathbb{R}_{\geq 0}\) are weighting coefficients to adjust the loss values to the same magnitude., and \(\text{CE}(\cdot, \cdot)\) is the cross-entropy
\begin{equation}
    \label{eq:cross_entropy}
    \text{CE}(\mathbf{y}, \mathbf{\hat{y}}) = -\sum_{i=1}^{|C|} y_i \log(\hat{y}_i).
\end{equation}

\noindent The importance score \(\xi(x_i)\) is defined in definition \ref{def:importance_score} and we normalize it using Equation~\ref{eq:normalized_xi}. Also, here we are using an exponential weighting scheme for incorporating the the importance score \(\xi(x)\) into the loss function.
Using the exponential term \((e^{\xi_{norm}(x)} - 1)\) as an importance score weight in Equation~\ref{eq:lossfn} is highly beneficial for targeted machine unlearning because it creates a non-linear scaling mechanism that strategically focuses the forgetting process. This exponential factor strongly amplifies the unlearning signal for samples deemed highly important or unique (\(\xi_{norm}(x) \gg 0\)), ensuring they receive significant pressure to be forgotten by the model. Conversely, for samples with low importance (\(\xi_{norm}(x) \approx 0\)), the factor approaches zero, effectively dampening the unlearning signal and preventing the model from inadvertently forgetting common knowledge or patterns that should be retained. 

\section{Evaluation and Results}
\label{sec:experiments}

In this section, we present a comprehensive experimental evaluation of \textsc{TraceHiding}, our proposed machine unlearning method. We begin by outlining the key research questions guiding our analysis in Section~\ref{sec:questions}.
Next, we describe the experimental setup in Section~\ref{sec:setup}, including the datasets, model architectures, baseline methods used for comparison, and evaluation metrics.
Then we start answering the research questions, discussing the selection and integration of importance scores  in Section~\ref{sec:RQ_importance_Score}.
Following this, we present core experimental studies for unlearning in Section~\ref{sec:eval_uniform_sampling}. Each study is conducted across multiple datasets and model architectures to evaluate generalizability and includes detailed comparisons against several established baseline methods.
Furthermore, we analyze the robustness across different sampling for deletion in Section~\ref{sec:RQ_deletion_sampling}, and trade-off between accuracy and speed in Section~\ref{sec:RQ_tradeoffs_speed}.

\subsection{Research Questions}
\label{sec:questions}

Our evaluation focuses on a machine unlearning scenario where users request the deletion of their data. Specifically, we address the case of \textbf{complete user-level unlearning}, where \textbf{all trajectory data associated with a given user} are entirely removed upon request. This differs fundamentally from approaches that delete only random or isolated trajectory segments. Our ``\textbf{User Deletion}'' strategy ensures that any identifiable traces of a user's movement patterns are comprehensively erased. By unlearning the entire history of trajectories linked to a user, our method aims to prevent the reconstruction of behavioral patterns, thereby \textbf{reducing re-identification risks} and \textbf{strengthening user privacy protection}. Before presenting our experimental evaluation, we first outline the key\textbf{ research questions (RQs)} that guide this study:
\begin{itemize}[leftmargin=*, labelsep=0.5em]
    \item RQ1. \textbf{Designing Optimal Importance Scoring}. How can we systematically define an importance score that reflects each trajectory’s contribution to the model, and how do different formulations impact the unlearning process?
    \item RQ3. \textbf{Advancing Unlearning Methodologies}. In what ways does our proposed approach extend the capabilities of existing baseline unlearning methods and enable more effective and reliable trajectory data removal?
    \item RQ3. \textbf{Impact of Sampling Strategies}. How do uniform and targeted sampling strategies, along with varying sample sizes, affect unlearning effectiveness and privacy guarantees?
    \item RQ4. \textbf{Balancing Accuracy and Efficiency}. What are the fundamental trade-offs between achieving high unlearning accuracy and maintaining computational efficiency in large-scale trajectory data settings?
\end{itemize}
\noindent These questions are aimed at understanding how the concept of sample importance can be harnessed to improve the design and performance of machine unlearning techniques. By systematically addressing these questions, we seek to uncover not only the strengths and limitations of our proposed method, but also the broader trade-offs involved in balancing unlearning effectiveness, computational efficiency, and data utility. Therefore, this section distills the extensive appendix tables
(Tables~\ref{tab:results_rome_uniform_sampling}–\ref{tab:results_NYC_targeted_sampling})
into concise comparisons that answer our four research questions (\textbf{RQ1–RQ4}).

\subsection{Experimental Setup} \label{sec:setup}
To thoroughly evaluate the efficacy and robustness of our proposed unlearning methodology, we design a comprehensive experimental setup that captures various realistic scenarios and challenges. This section details the core components of our experiments, including the specific deletion scenarios considered, the user sampling strategies employed to select data for deletion, the datasets used for evaluation, the diverse trajectory classification models tested, and the metrics used to assess unlearning performance. Together, these components form a rigorous framework to systematically analyze how well unlearning methods perform under different conditions and model architectures.
We begin by defining the deletion scenarios to establish the context of data removal requests. Next, we describe the strategies used to sample users for deletion, highlighting both uniform and targeted approaches. We then introduce the datasets and their characteristics, followed by a presentation of the classification models leveraged to validate our approach across various architectures. Finally, we conclude with the evaluation metrics that quantify both the completeness and efficiency of the unlearning process.

\subsubsection{Deletion Scenarios}

In this work, we focus specifically on \textit{user deletion}, which entails removing all data associated with an individual user from the model. Unlike approaches that randomly select data points for unlearning, our scenario assumes that a user initiates a deletion request, and consequently, the entire dataset corresponding to that user must be unlearned. This user-centric deletion reflects practical privacy demands and regulatory requirements, such as those outlined by data protection laws, where users have the right to request the complete removal of their personal data.
Although our primary focus is on user level deletion, the proposed methods can be naturally extended to handle random or partial deletion scenarios. This flexibility ensures that the framework remains applicable across a range of unlearning tasks, whether targeting specific data points, segments, or users as a whole.

\subsubsection{Sampling Strategies for Deletion}

We employ \textbf{two distinct user sampling strategies to select candidates for deletion}, each serving a specific purpose in evaluating the robustness of unlearning methods:
\begin{itemize}
    \item \textbf{Uniform Sampling}. Designed to sample users uniformly at random (i.e., with equal probability). This strategy provides a baseline scenario, ensuring that deletions occur without bias or preference towards any particular user group. Uniform sampling allows us to assess the performance of deletion methods under conditions that simulate arbitrary and unbiased removal requests.
    \item \textbf{Targeted Sampling}. Designed to sample users proportionally to their aggregated entropy values. Entropy, as a measure of uncertainty or information content, is computed for each user based on their trajectories. By sampling users according to these entropy values, we focus deletion efforts on users contributing higher informational complexity or variability. This strategy enables us to examine how deletion methods perform when faced with more challenging or influential user data, thereby providing insights into their effectiveness under targeted unlearning scenarios.
\end{itemize}

\noindent For each dataset, we conduct experiments using sample sizes of 1\%, 5\%, 10\%, and 20\% to evaluate the unlearning performance at varying scales of data removal. To ensure robustness and reliability of the results, each experiment is repeated across 5 independent runs, allowing us to account for variability due to random initialization or sampling. The final reported metrics are averaged over these runs, providing a more stable and statistically meaningful assessment of the unlearning methods. This rigorous evaluation protocol helps to minimize the influence of random chance and supports confident conclusions about method effectiveness across different deletion magnitudes.

\subsubsection{Datasets} \label{sec:datasets}
We briefly describe each dataset and their statistics in Table~\ref{tab:ho-dataset-stats} in Section~\ref{sec:data-representation}.
We are using two different types of trajectories: one is 1) the continuous path of users, and the other is 2) the sequence of check-ins that the user recorded. We treat both as sequences of tokens to capture the spatial-temporal patterns of user movement, regardless of the type of trajectory. The datasets are as follows: 

\noindent\textbf{\textsc{Ho-Rome}} \cite{bracciale2014crawdadrome, faraji2023point} consists of recorded mobility traces of taxis operating in Rome, Italy,

\noindent\textbf{\textsc{Ho-Geolife}} \cite{zheng2008geolife, zheng2009geolife, zheng2010geolife, faraji2023point} consists of recorded mobility traces of individuals, covering a total distance of $\sim$1.2M Km, and

\noindent\textbf{\textsc{Ho-NYC}} \cite{yang2014foursquare, faraji2023point} consists of \textsc{Poi} check-ins recorded by the phones of several users in the city of New York.

\subsubsection{Trajectory Classification Models} \label{sec:models}
We are testing various models for this task.
Our goal is to evaluate how well the unlearning process works with each model. 
By doing this, we can make sure that our proposed method is effective, no matter which model and architecture is used.
We conduct our experiments using the following four models:
\textbf{GRU} \cite{chung2014empirical}, \textbf{LSTM} \cite{hochreiter1997long}, \textbf{BERT} \cite{devlin2019bert}, \textbf{ModernBERT} \cite{warner2024smarter}, 
and \textbf{GCN-TULHOR} \cite{tran2025gcntulhortrajectoryuserlinkingleveraging}.
GRU and LSTM are classic recurrent neural network variants designed to capture long-range dependencies in sequential data. BERT and ModernBERT, conversely, are transformer-based architectures that leverage self-attention to achieve state-of-the-art performance across various language tasks. 
GCN-TULHOR is an extension of TULHOR \cite{mahmoud2023tul} that integrates Graph Convolutional Networks (GCNs) into the transformer framework to handle more complex structural information.
We include these models because they span the major families of sequence modeling, RNN, transformer-based, 
as well as a hybrid GCN-transformer approach
thereby ensuring our methodology is tested across a diverse and representative set of architectures.
This choice highlights that our proposed methodology is model-agnostic and addresses an orthogonal problem independent of any specific neural architecture.

\subsubsection{Unlearning Methods}
\label{sec:tracehidings}
\label{sec:baselines}

We introduce variants of our method, \textsc{TraceHiding}, which arise from different parameter choices, such as the importance score functions used for unlearning.

\smallskip\noindent\textbf{\textsc{TraceHiding (Ent)}.}
In this variant, we employ the entropy-based importance function, given by Equation~\ref{eq:entropy_importance}, to determine the significance of individual samples, which is a guide for unlearning.

\smallskip\noindent\textbf{\textsc{TraceHiding (C.D.)}.}
In this variant, we utilize the coverage diversity-based importance function, defined by Equation~\ref{eq:coverage_importance}, to assess sample significance, thereby informing the method about importance of samples for unlearning.

\smallskip\noindent\textbf{\textsc{TraceHiding (Uni.)}.}
In this variant, we employ a unified equation to calculate the importance score (Equation~\ref{eq:single_importance_score}). The selection of appropriate weights within this equation is discussed in Section~\ref{sec:unified_importance_weights}.

Furthermore, we use the following baselines for comparative analysis with our method. These methods are well-known in machine unlearning for classification and have appeared in most previous studies.

\smallskip\noindent\textbf{\textsc{NegGrad} \cite{kurmangi2024machine}.} 
This method applies gradient ascent exclusively to $\mathcal{D}_u$ (or equivalently, gradient descent on $\mathcal{D}_u$ with the gradient negated, which is why it is often called \textsc{NegGrad}). The underlying idea is to try to ``delete'' $\mathcal{D}_u$ by maximizing the loss on that data, effectively aiming to ``reverse'' the learning process the network previously performed to learn that data.

\smallskip\noindent\textbf{\textsc{NegGrad+} \cite{kurmangi2024machine}.}
\textsc{NegGrad} might reduce the performance on the remaining dataset $\mathcal{D}_r$ because it lacks motivation to preserve useful information while conducting gradient ascent. Essentially, if the remaining data is similar to the unlearning dataset, \textsc{NegGrad}’s goal of maximizing the loss on $\mathcal{D}_u$ could inadvertently increase the loss on parts of $\mathcal{D}_r$, which is undesirable.
To tackle this issue, it employs a more robust approach that performs gradient ascent on the unlearning dataset (like \textsc{NegGrad}) and gradient descent on the remaining dataset (similar to Fine-tuning). This method aims to achieve a balance by maximizing the loss on $\mathcal{D}_u$ while minimizing it on $\mathcal{D}_r$.

\smallskip\noindent\textbf{\textsc{SCRUB} \cite{kurmanji2024towards}.}
\textsc{SCRUB} uses a teacher-student approach. The teacher is the original model, trained on the full dataset. The student starts with the teacher's knowledge and then learns to keep only the information relevant to the remaining dataset. The student has two main goals: to become similar to the teacher for the remaining dataset and to become different from the teacher for the dataset to be forgotten. This way, information about the unlearning dataset is effectively removed. \textsc{SCRUB} aims to balance removing information about the unlearning dataset while keeping information about the remaining dataset. It does this through both positive and negative knowledge transfer from the original model.

\smallskip\noindent\textbf{\textsc{Finetuning}.}
This method is finetuning on the remaining dataset $\mathcal{D}_r$.
This method focuses on the remaining data, revisiting it to ensure we prioritize it over the data that hasn't undergone fine-tuning. By doing so, we concentrate on improving the parts that need more attention and forgetting the data that doesn't participate in fine-tuning.

\smallskip\noindent\textbf{\textsc{Bad-T} \cite{chundawat2023can}.} 
This work introduces a novel machine unlearning technique through a teacher-student framework. The core of this approach involves two distinct types of teachers: a competent teacher, which is a fully trained model that encapsulates the complete knowledge from the training data, and an incompetent teacher, which is a randomly initialized model that lacks any meaningful knowledge. The student model, initially carrying the knowledge of the competent teacher, undergoes a selective unlearning process guided by these two teachers.
This method tries to achieve unlearning by minimizing a loss on the student’s outputs and those of the incompetent teacher for the unlearning data, effectively erasing this information. Simultaneously, the loss is minimized between the student’s outputs and those of the competent teacher for the retain data, preserving essential knowledge.

\subsubsection{Evaluation Metrics}

Machine unlearning methods are typically evaluated across three key dimensions: \textbf{(i) utility}, the extent to which the original model’s performance is preserved; \textbf{(ii) completeness}, the degree to which the influence of deleted data is fully removed, ensuring consistency between the unlearned model and a retrained model; and \textbf{(iii) timeliness}, the efficiency of the unlearning process compared to full retraining. Based on these criteria, we adopt the following evaluation metrics.

\medskip\noindent\textbf{A. Accuracy}. Measures the overall correctness of the model by dividing the number of correct predictions (both true positives and true negatives) by the total number of cases examined:

\begin{equation}
\label{eq:accuracy}
    \text{Accuracy} = \frac{TP + TN}{TP + TN + FP + FN}.
\end{equation}

\noindent We calculate accuracy on different splits of the dataset:
\begin{itemize}
    \item Remaining dataset accuracy (\textbf{RA}).
    \item Test dataset accuracy (\textbf{TA}).
    \item Unlearning dataset accuracy (\textbf{UA}).
\end{itemize}
The Unlearning dataset accuracy (UA) defined by $1-$\textit{accuracy of a model on the unlearning dataset}. This approach is common and has been used in different related works  \cite{fan2024challenging, bonato2024is, basak2024forget}. This metric reflects successful forgetting, where a higher UA indicates that the model is no longer able to correctly classify the unlearned samples, effectively performing no better than a random guess in an ideal scenario.

\medskip\noindent\textbf{B. Membership Inference Attack (MIA)}. 
MIAs, or membership inference attacks, originated in the field of privacy and security research. Their primary objective is to determine which specific samples were included in the training dataset.
If unlearning is successful, the unlearned model will not retain any evidence of the forgotten examples. As a result, MIAs will not be effective. The attacker will be unable to figure out that the forgotten set was originally included in the training set.
This approach has been used in the unlearning context \cite{fan2024challenging}.
We use the Area Under the ROC Curve (AUC) of the MIA to quantify the effectiveness of unlearning.
The MIA AUC measures how well an adversary can distinguish between training and non-training data points using model outputs. A value of 1.0 indicates perfect classification by the attacker, while 0.5 corresponds to random guessing. In the context of unlearning, our objective is to remove information about certain users such that the model no longer retains any identifiable signal from them. Therefore, a lower MIA AUC indicates better forgetting, as it reflects reduced confidence or accuracy in the attacker's ability to infer membership.
For a successfully unlearned model, the AUC should be close to 50\%, indicating that the attack performs no better than a random guess. Therefore, we report the difference from a random classifier; the lower the difference, the better the quality of unlearning.

\medskip\noindent\textbf{C. Unlearning Speedup}. We are measuring the wall-clock unlearning time to compare the speed of unlearning with the retraining time. We made sure that the running environment is consistent across the runs. We report the speed up relative to the retraining model:

\begin{equation}
\label{eq:speedup}
    \text{Speedup} = \frac{\text{Runtime of Retraining}}{\text{Runtime of the Unlearning}}.
\end{equation}

\subsection{RQ1: Designing Optimal Importance Scoring}
\label{sec:RQ_importance_Score}
Designing an optimal importance scoring mechanism involves several key considerations. We first discuss the criteria for choosing an importance score, then explore the methodologies for determining weighting parameters for a unified score, and finally analyze the effect of the importance score on model performance.

\subsubsection{Choosing the Importance Score}
\label{sec:choosing_imortance_score}

\begin{figure}[t]
    \centering
        \begin{subfigure}[b]{\textwidth}
            \includegraphics[width=\linewidth]{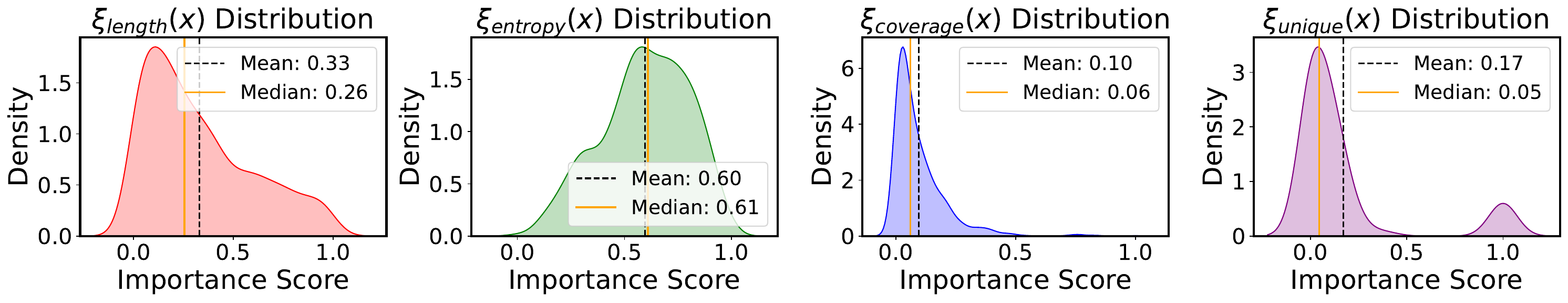}
            \caption{HO-Geolife Dataset Importance Density Plot.}
            \label{fig:importance_score_density_geolife}
        \end{subfigure}

        \begin{subfigure}[b]{\textwidth}
            \includegraphics[width=\linewidth]{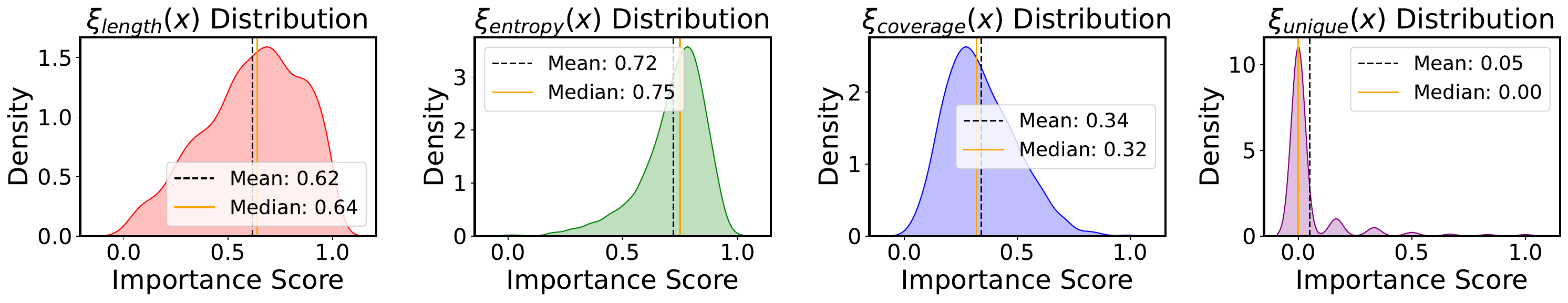}
            \caption{HO-Rome Dataset Importance Density Plot.}
            \label{fig:importance_score_density_rome}
        \end{subfigure}

        \begin{subfigure}[b]{\textwidth}
            \includegraphics[width=\linewidth]{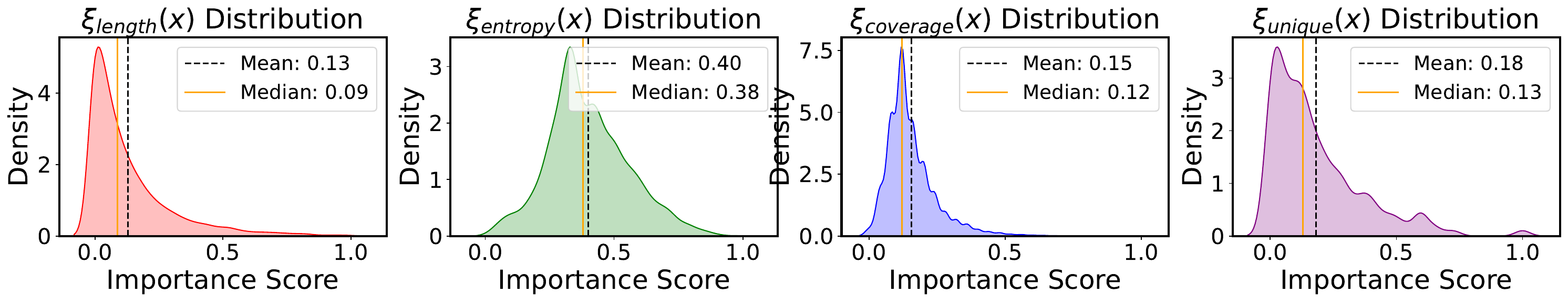}
            \caption{HO-NYC Dataset Importance Density Plot.}
            \label{fig:importance_score_density_nyc}
        \end{subfigure}
    
    \caption[Density plots show importance score distributions across four definitions]{Density plots depicting the distribution of normalized importance scores across datasets with different definitions from the left to the right, (1) Length of Trajectory, (2) Entropy, (3) Coverage Diversity, and (4) User Uniqueness. Each plot shows the spread of normalized importance scores, with the x-axis representing the normalized score and the y-axis representing the density of occurrences. Mean and median importance scores are indicated, highlighting differences in different importance across datasets.}
    \label{fig:importance_score_density}
\end{figure}

The choice of an importance score is critical to effectively remove the influence of specific data samples while maintaining the integrity of the remaining model. Different downstream tasks may necessitate different importance metrics, with the choice often depending on dataset characteristics or model behavior.
In our approach, we opted for the following importance scores in our experiments: (1) Coverage Diversity, (2) Information-theoretic, and (3) Unified Trajectory importance score including the length importance.
We discuss how we can choose the weighting scheme of the unified score function in Section~\ref{sec:unified_importance_weights}.
The importance score distributions for different datasets are presented in Figure~\ref{fig:importance_score_density}.
Each subplot visualizes the distribution of importance scores, with mean and median values marked to indicate central tendencies. 
The differences in distributions across datasets highlight variations in the significance of these metrics.

The selected importance score is designed to reflect each sample's role in influencing the decision boundaries, making it particularly useful for identifying which data points contribute most to model behavior. This approach enables us to prioritize the unlearning of high-impact samples, thus minimizing model degradation after the removal.

\begin{figure*}[t]
    \centering
    \begin{subfigure}[t]{0.5\textwidth}
        \centering
        \includegraphics[width=\linewidth,height=7cm]{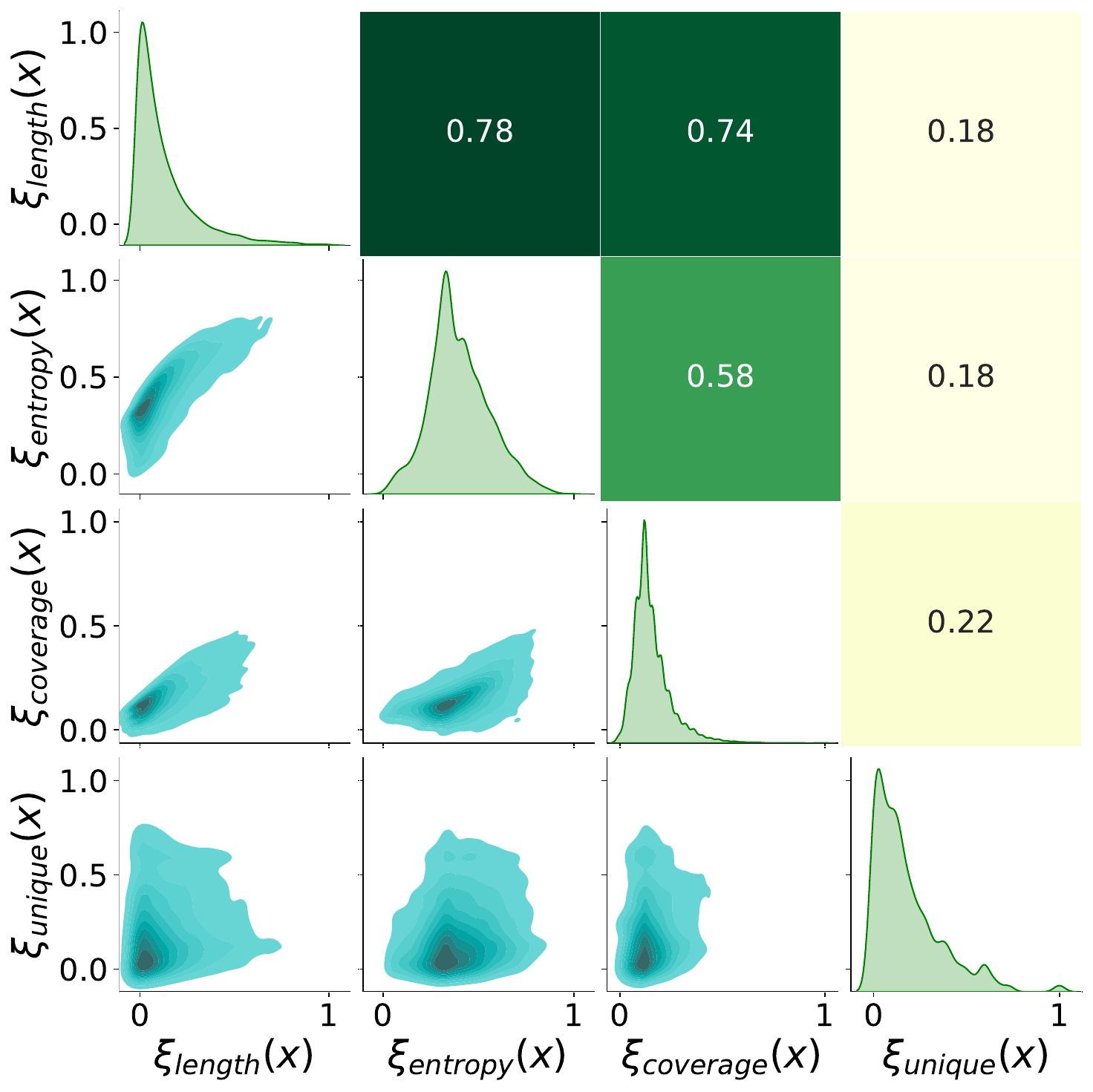}
        \caption{Ho-NYC Dataset}
        \label{fig:pair_corr_nyc}
    \end{subfigure}%
    ~ 
    \begin{subfigure}[t]{0.5\textwidth}
        \centering
        \includegraphics[width=\linewidth, height=7cm]{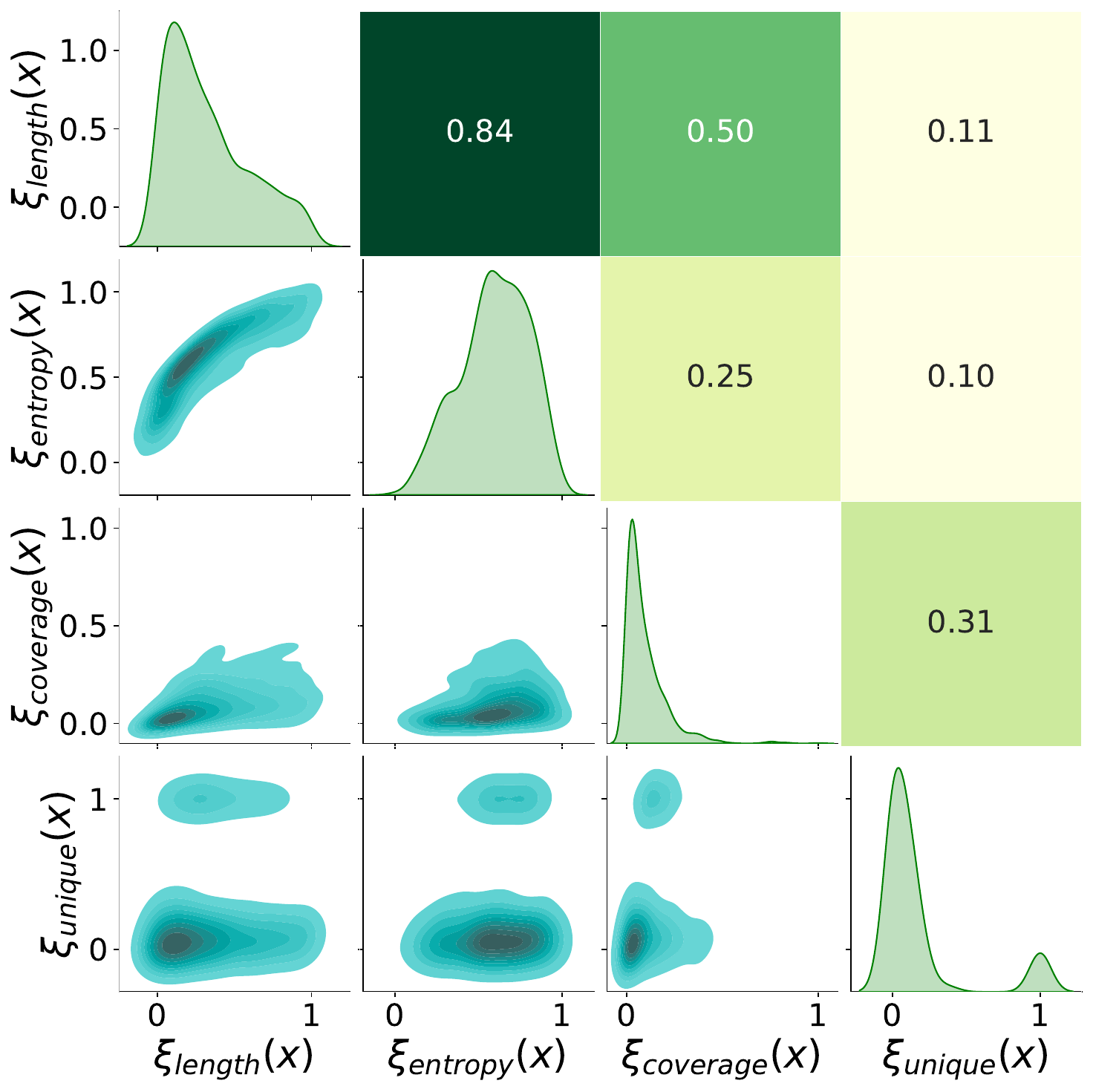}
        \caption{Ho-Geolife Dataset}
        \label{fig:pair_corr_geolife}
    \end{subfigure}
    
    \caption[Pairwise plots show importance distributions and correlations across datasets]{Pairwise comparison of four importance scores 
($\xi_{\text{length}}, \xi_{\text{entropy}}, \xi_{\text{coverage}}, \xi_{\text{unique}}$) 
on (a) Ho-NYC and (b) Ho-Geolife datasets. 
The figure is arranged as a symmetric matrix where each row and column correspond 
to one importance score. 
The diagonal cells show the marginal distribution of each score individually. 
The cells below the diagonal show the joint distribution between two scores 
using contour density plots, which reveal how the values co-vary. 
The cells above the diagonal display the correlation coefficient 
between the two scores, with the background color indicating the strength of correlation. 
This analysis highlights that $\xi_{\text{length}}$ and $\xi_{\text{entropy}}$ 
are highly correlated in both datasets, $\xi_{\text{coverage}}$ shows 
moderate correlation with them, and $\xi_{\text{unique}}$ is largely 
independent, particularly in Ho-Geolife where it exhibits a clear bimodal distribution.}
    \label{fig:pairwise_corr}
\end{figure*}

We compared different importance scores within a dataset in Figure \ref{fig:pairwise_corr}, we can see the positive correlation (upper triangle $> 0$) between length, entropy, and coverage diversity, but user uniqueness shows a much weaker correlation with other three importance scores and tends to cluster near zero, as depicted in Figure~\ref{fig:importance_score_density}. This is due to the strict definition of the uniqueness function in Equation~\ref{eq:user_unique_importance}, which requires that a token (block) appears exclusively in a single user's trajectory to be considered unique.
Given the nature of urban mobility, where users share common pathways, achieving high uniqueness is uncommon. As a result, we have decided not to use user uniqueness in our experiments, as its low values may not contribute meaningfully to the overall importance evaluation. However, we include it in our analysis to demonstrate that alternative definitions of importance exist, and this metric could still be useful in specific cases, such as datasets with sparse user distributions or disjoint mobility patterns across different parts of the city.

\subsubsection{Weighting Parameters for Unified Score}
\label{sec:unified_importance_weights}

We observe a strong positive correlation between the key importance scores length, coverage diversity, and entropy, indicating that even if we combine them into a single function, their contributions would not be contradictory.
Since these three importance scores align well, they provide a consistent way to evaluate importance without introducing conflicting influences. Their strong interdependence suggests that they collectively capture the richness and comprehensiveness of user trajectories, making them reliable measures for our use case.
When combining multiple scores into a single importance score, it is essential to determine the appropriate weighting parameters. This process can be approached through both \textbf{manual adjustments tailored to specific downstream} tasks and \textbf{unsupervised} optimization techniques.

\textbf{I) Manual Weight Selection for Downstream Tasks}
In many cases, the relative importance of the individual scores may be informed by the requirements of a specific downstream task. For example:
\begin{itemize}
    \item \textbf{Domain Knowledge:} Weights can be assigned manually based on expert knowledge or prior experience regarding the significance of each score.
    \item \textbf{Task-Specific Priorities:} Different tasks may prioritize certain scores over others. For instance, if interpretability is critical, simpler weighting schemes (e.g., equal weights) may be preferred, while performance-oriented tasks may rely on fine-tuned weights informed by data analysis.
\end{itemize}

\begin{figure*}[t]
    \centering
    \begin{subfigure}[t]{0.3\textwidth}
        \centering
        \includegraphics[width=\linewidth]{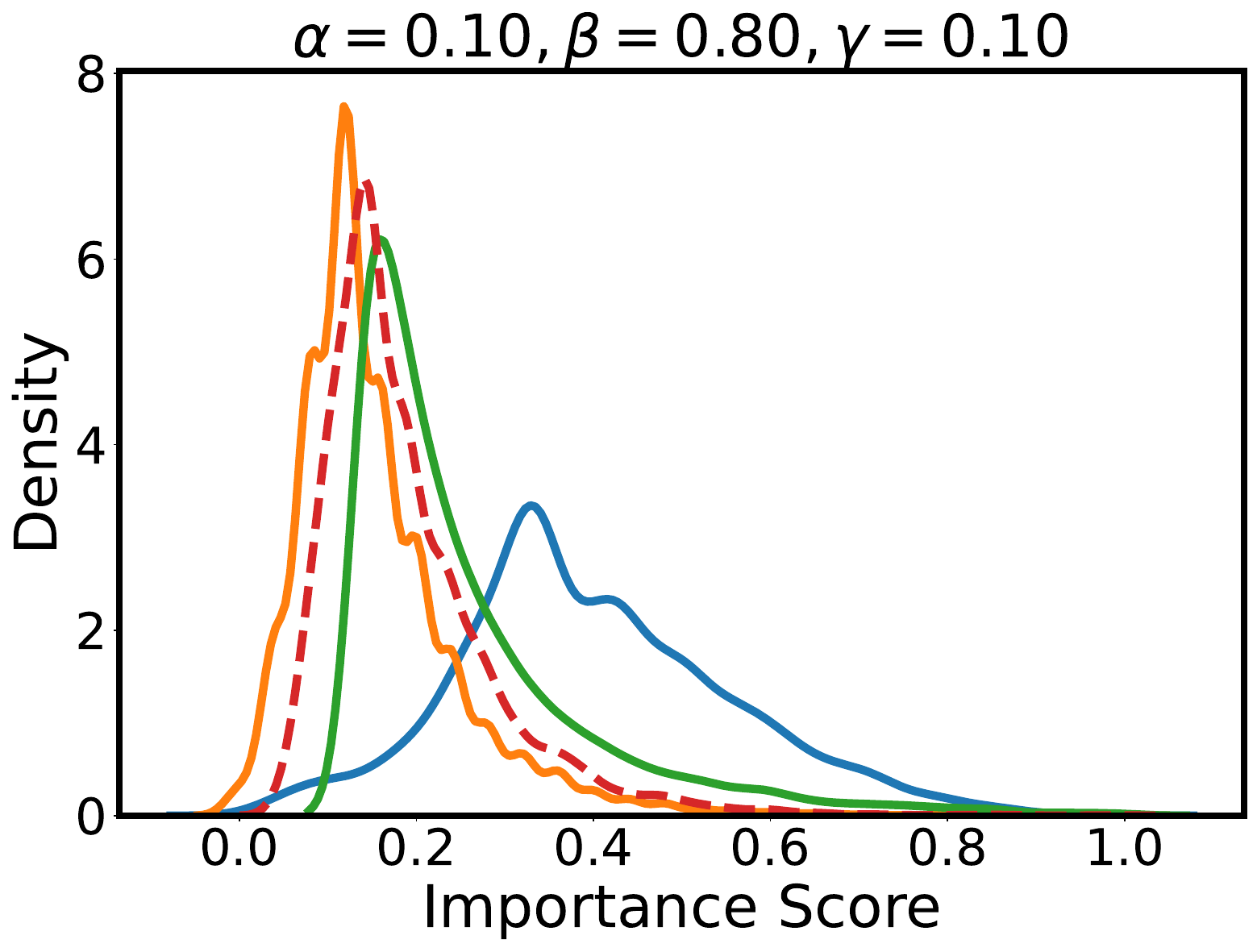}
        \caption{HO-NYC Dataset}
        \label{fig:unified_score_nyc}
    \end{subfigure}%
    ~ 
    \begin{subfigure}[t]{0.3\textwidth}
        \centering
        \includegraphics[width=\linewidth]{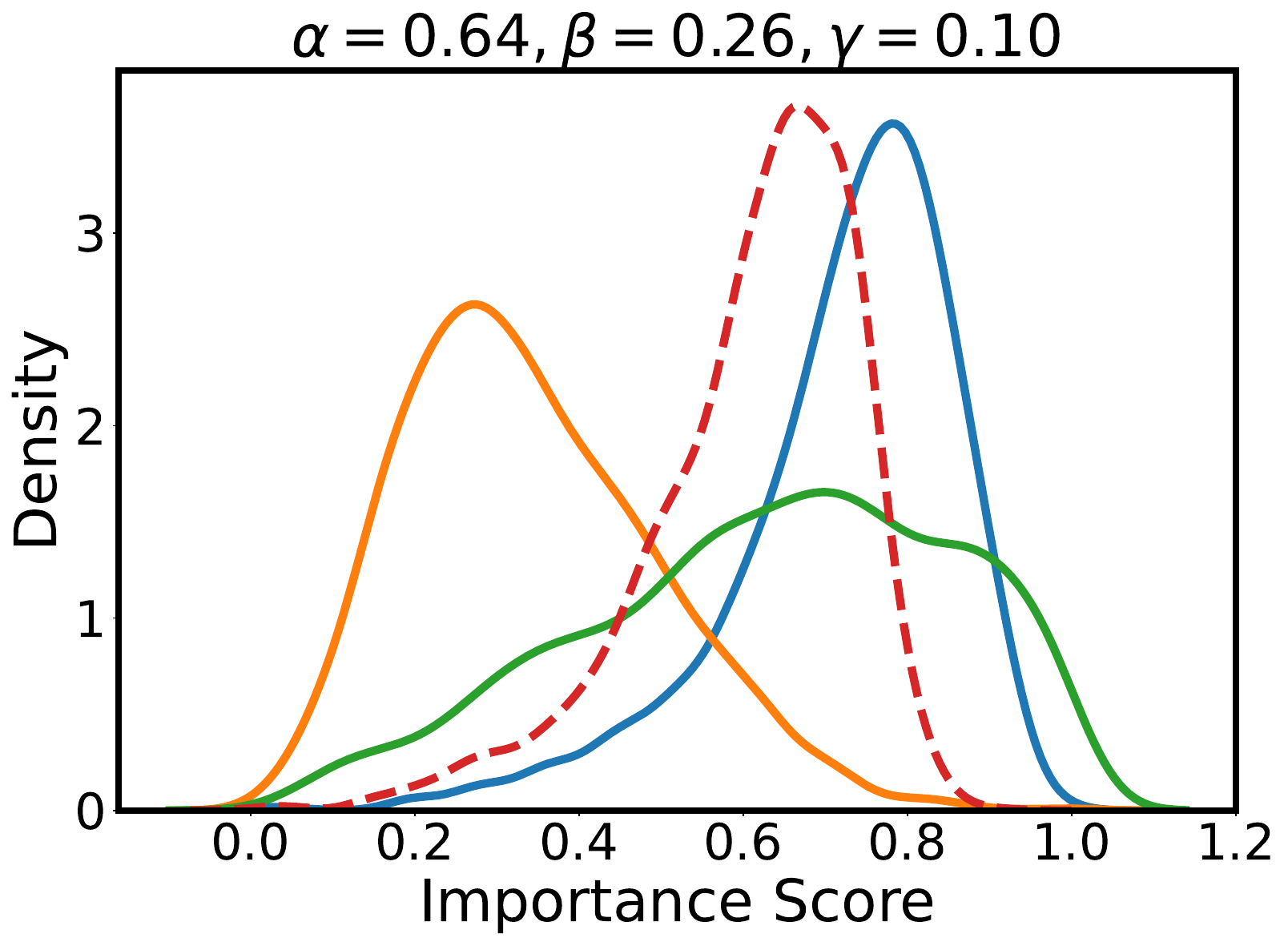}
        \caption{HO-Rome Dataset}
        \label{fig:unified_score_rome}
    \end{subfigure}%
    ~
    \begin{subfigure}[t]{0.3\textwidth}
        \centering
        \includegraphics[width=\linewidth]{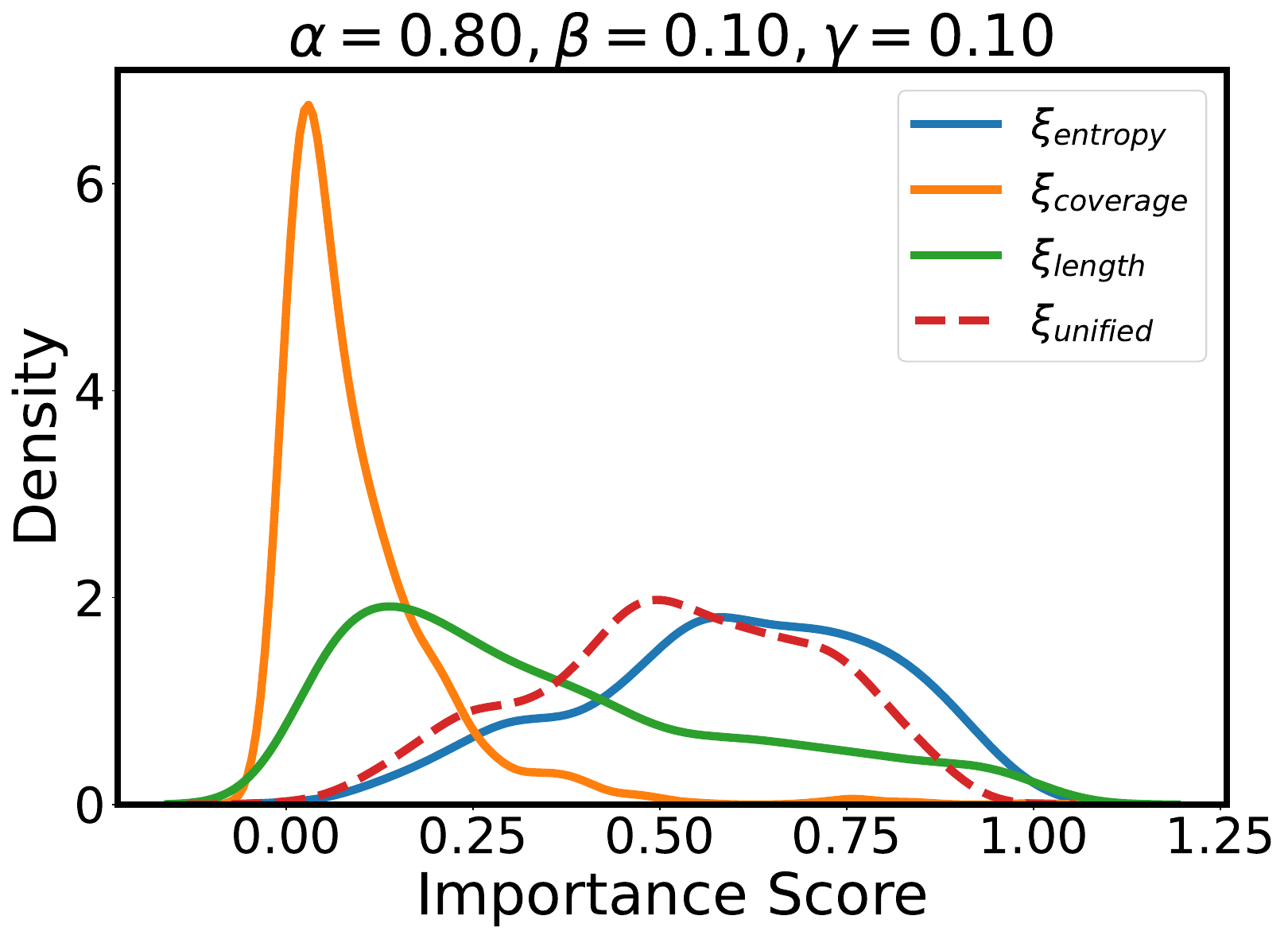}
        \caption{HO-Geolife Dataset}
        \label{fig:unified_score_geolife}
    \end{subfigure}
    
    \caption[Density distribution of individual scores and unified version]{The figure illustrates the density distribution of individual scores and the optimized unified score for different datasets. The optimization incorporates higher-order shape descriptors (skewness and kurtosis), ensuring that the composite score is spread out, symmetric, and robust to outliers.}
    \label{fig:unified_score}
\end{figure*}

\textbf{II) Unsupervised Optimization Techniques}
Unsupervised methods are particularly useful when there is no explicit ground truth or target variable to guide the weighting process. These techniques aim to optimize certain statistical properties of the unified score. For example:
\begin{itemize}
    \item \textbf{Variance Maximization:} Allocates weights to maximize the spread of the unified score across users or samples, ensuring that it captures as much variability as possible from the underlying data.
    \item \textbf{Entropy Maximization:} Encourages a broad and balanced distribution of the unified score by maximizing its entropy, promoting uniform contributions from the individual scores.
    \item \textbf{Asymmetry and Tailedness Measures:} Incorporates additional statistical measures such as skewness and kurtosis. These metrics ensure that the unified score is not only spread out but also symmetric and devoid of extreme outliers, 
    balancing its statistical properties. As shown in Figure~\ref{fig:unified_score}, the optimization effectively incorporates these metrics by minimizing the objective function:
    \begin{equation}
\text{Objective} = -w_1 \cdot V + w_2 \cdot |S| + w_3 \cdot K_e,
\label{eq:importance_weights_optimization_objective}
\end{equation}
\end{itemize}

\noindent In the objective function (Equation~\ref{eq:importance_weights_optimization_objective}), \(V\) represents the variance of the trajectory level unified scores, calculated using the combined score function in Equation~\ref{eq:single_importance_score} and illustrated in Figure~\ref{fig:unified_score}. The variance measures the spread of scores across samples, and maximizing it ensures greater diversity in the distribution. The term \(S\) denotes the skewness of the scores' distribution, which quantifies the asymmetry of the distribution. Positive skewness (\(S > 0\)) indicates a right-skewed distribution with a longer tail on the right, while negative skewness (\(S < 0\)) indicates a left-skewed distribution with a longer tail on the left. The optimization minimizes the absolute value of skewness, \(|S|\), in order to encourage symmetry in the distribution. 
The term \(K_e\) corresponds to the excess kurtosis of the unified scores, which is defined as kurtosis minus three so that a normal distribution is centered around zero. Positive values of excess kurtosis (\(K_e > 0\)) reflect heavier tails with more outliers, whereas negative values (\(K_e < 0\)) reflect lighter tails with fewer outliers. Minimizing \(K_e\) therefore reduces the influence of extreme outliers. Finally, the coefficients \(w_1, w_2,\) and \(w_3\) are fixed weights assigned to variance, skewness, and kurtosis, respectively, and they determine the relative importance of each term in the optimization. For simplicity, equal weights are chosen, such that \(w_1 = \tfrac{1}{3}, w_2 = \tfrac{1}{3}, w_3 = \tfrac{1}{3}\).

\smallskip\noindent By combining unsupervised techniques with manual adjustments, it is possible to create a flexible and robust method for determining the weighting parameters. This ensures that the unified score is both data-driven and aligned with the specific goals of the application.

\subsubsection{Effect of the importance score}
\label{sec:importance_score_impact}

Analysis of the \textsc{TraceHiding} variants, as detailed in Table~\ref{tab:th-variants}, reveals nuanced trade-offs and strengths among its different variants (Coverage Diversity, Entropy, Unified) when compared against a full Retraining baseline at 10\% uniform sampling. The bolded numbers indicate the variant closest to the Retraining method's performance for each metric. A key insight is that no single \textsc{TraceHiding} variant consistently mimics the Retraining baseline across all metrics and datasets. Specifically, the \textsc{Entropy} variant frequently shows performance closest to Retraining for Unlearning Accuracy (e.g., 70.20\% for HO-Geolife, 68.61\% for HO-NYC, 89.82\% for HO-Rome) and sometimes for Remaining Accuracy (80.90\% for HO-Geolife) and Test Accuracy (56.98\% for HO-Geolife, 73.36\% for HO-NYC). This suggests that \textsc{Entropy}-based \textsc{TraceHiding} is often effective at achieving unlearning outcomes and maintaining overall model behavior that closely resembles a model retrained from scratch. Conversely, the \textsc{Unified} variant demonstrates a notable strength in matching Retraining's Membership Inference Attack (MIA) accuracy (e.g., 43.79\% for HO-Geolife, 43.65\% for HO-NYC, 15.21\% for HO-Rome), indicating its potential in replicating the privacy characteristics of a retrained model. The \textsc{Coverage Diversity} variant, while generally competitive, is less frequently the closest to Retraining across these metrics. This comparison highlights that different \textsc{TraceHiding} strategies excel in different aspects of unlearning, with \textsc{Entropy} often balancing unlearning effectiveness with overall model utility, and \textsc{Unified} showing promise in privacy-related metrics, emphasizing the need to select a variant based on the specific unlearning goals and desired trade-offs.

\begin{table}[t]
\caption{\textsc{TraceHiding} variant comparison (10\,\%, uniform sampling).}
\label{tab:th-variants}
\centering
\begin{tabular*}{\linewidth}{@{\extracolsep{\fill}}llcccc}
\toprule
Dataset & \textbf{Variant} & UA & RA & TA & MIA\\
\midrule
\multirow{4}{*}{HO-Geolife}
 & Coverage Diversity   & 60.62 & 83.49 & 58.92 & 44.08\\ 
 & Entropy              & \textbf{70.20} & \textbf{80.90} & \textbf{56.98} & 44.61\\ 
 & Unified              & 59.56 & 83.66 & 58.96 & \textbf{43.79}\\ 
 & Retraining           & 83.05 & 68.39 & 49.30 & 33.08\\ 
\midrule
\multirow{4}{*}{HO-NYC}
 & Coverage Diversity   & 58.98 & 94.19 & 75.21 & 43.72\\ 
 & Entropy              & \textbf{68.61} & 92.98 & \textbf{73.36} & 43.80\\ 
 & Unified              & 66.04 & \textbf{93.53} & 73.95 & \textbf{43.65}\\ 
 & Retraining           & 80.07 & 93.34 & 73.57 & 38.22\\ 
\midrule
\multirow{4}{*}{HO-Rome}
 & Coverage Diversity   & 85.41 & \textbf{39.69} & 21.42 & 16.56\\ 
 & Entropy              & \textbf{89.82} & 32.46 & 18.57 & 18.80\\ 
 & Unified              & 83.88 & 38.69 & \textbf{21.03} & \textbf{15.21}\\ 
 & Retraining           & 87.67 & 43.31 & 20.64 & 10.91\\ 
\bottomrule
\end{tabular*}
\end{table}

\subsection{RQ2: Advancing Unlearning Methodologies}
\label{sec:RQ_unlearning}
In this section, we evaluate unlearning approaches across three critical aspects: their performance under uniform sampling, robustness across diverse datasets, and adaptability to different model architectures.

\subsubsection{Unlearning under Uniform Sampling}
\label{sec:eval_uniform_sampling}

Uniform sampling removes a random fraction of users
(1\,\%, 5\,\%, 10\,\%, 20\,\%).  
Table~\ref{tab:ua-rank-uniform} shows the \emph{average rank}
(lower = better) of each method’s \textbf{Unlearning Accuracy (UA)}
across all three datasets and the five model architectures.
Across all removal fractions, the \textsc{TraceHiding} family clearly dominates in Table~\ref{tab:ua-rank-uniform}: the entropy variant achieves the lowest mean rank (3.03) and improves as the removal percentage grows (rank drops from 3.73 at 5,\% to 2.67 at 10–20,\%), while the uniform‐weight variant, despite a slightly higher mean rank (3.46), exhibits the smallest rank volatility, making it the most predictable choice. \textsc{NegGrad} occupies a respectable mid-tier position (mean~3.80), but its enhanced version (\textsc{NegGrad$+$}) performs worse, hinting that the added loss for remembering the remaining data had adverse effect on unlearning accuracy. (\textsc{SCRUB}, \textsc{Bad-T}) and naive \textsc{Finetuning} cluster at the bottom, indicating that generic method without considering the importance or just finetuning heuristics translate poorly into unlearning accuracy.

\begin{table}[t]
\caption[Average unlearning method rank under uniform sampling]{The average rank of methods based on Unlearning Accuracy (UA) across different datasets and models under \textbf{uniform} sampling, where a lower rank indicates better performance. Results are shown for various sizes of unlearning data samples.}
\label{tab:ua-rank-uniform}
\centering
\begin{tabular*}{\linewidth}{@{\extracolsep{\fill}} l c c c c}
\toprule
Method & 1 \% & 5 \% & 10 \% & 20 \% \\
\midrule
\textsc{TraceHiding (Ent.)} & 3.03 & 3.73 & 2.67 & 2.67 \\
\textsc{TraceHiding (C.D.)} & 3.67 & 3.40 & 3.07 & 3.27 \\
\textsc{TraceHiding (Uni.)} & 3.57 & 3.47 & 3.53 & 3.27 \\
\textsc{NegGrad}           & 4.00 & 3.57 & 3.73 & 3.90 \\
\textsc{NegGrad+}           & 4.40 & 4.43 & 5.07 & 5.23 \\
\textsc{SCRUB}              & 5.47 & 5.43 & 5.97 & 5.27 \\
\textsc{Bad-T}              & 5.83 & 6.00 & 5.53 & 6.03 \\
\textsc{Finetuning}         & 6.03 & 5.97 & 6.43 & 6.37 \\
\bottomrule
\end{tabular*}
\end{table}


\subsubsection{Cross-Dataset Robustness}
\label{sec:cross_dataset_results}

Table~\ref{tab:mean-performance-10} reports the mean Unlearning Accuracy (UA) and Test Accuracy (TA) obtained after deleting \emph{10\,\%} of the training data \emph{per dataset}. Across all three datasets, the relative ordering of the methods is stable: the proposed \textsc{TraceHiding} variants remain among the top performers.
\begin{table}[t]
\caption[Ranks methods by accuracy and runtime after data deletion]{Mean Unlearning Accuracy (UA) and Test Accuracy (TA) at 10\,\% \emph{uniform} deletion, averaged over four models per dataset.  
Methods are ranked by their average distance to the \emph{Retraining} baseline (gold standard) on both UA and TA.  
The closest method is shown in \textbf{bold}; the second closest is \underline{underlined}.  
Runtime is expressed as a multiplicative factor with respect to full retraining.}
\label{tab:mean-performance-10}
\centering
\begin{tabular*}{\linewidth}{@{\extracolsep{\fill}} lcccccccc}
\toprule
\multirow{2}{*}{\textbf{Method}} &
\multicolumn{2}{c}{HO-Rome} &
\multicolumn{2}{c}{HO-Geolife} &
\multicolumn{2}{c}{HO-NYC} &
\multirow{2}{*}{Speedup}\\
\cmidrule(lr){2-3}\cmidrule(lr){4-5}\cmidrule(lr){6-7}
& UA & TA &
  UA & TA &
  UA & TA & \\
\midrule
Retraining & 87.67 & 20.64 & 83.05 & 49.30 & 80.07 & 73.57 & 1$\times$ \\
\midrule
\textsc{TraceHiding (Ent.)} & \textbf{89.82} & 18.57 & \textbf{70.20} & 56.98 & \textbf{68.61} & \textbf{73.36} & 39.97$\times$ \\
\textsc{TraceHiding (Uni.)} & 83.88 & \textbf{21.03} & 59.56 & 58.96 & \underline{66.04} & \underline{73.95} & 39.78$\times$ \\
\textsc{TraceHiding (C.D.)} & \underline{85.41} & 21.42 & 60.62 & 58.92 & 58.98 & 75.21 & 39.31$\times$ \\
\textsc{SCRUB} & 84.32 & 17.44 & 57.81 & 57.17 & 61.96 & 65.92 & 37.90$\times$ \\
\textsc{Bad-T} & 81.94 & \underline{19.88} & 59.07 & 57.49 & 51.12 & 74.29 & 32.78$\times$ \\
\textsc{NegGrad+} & 90.41 & 16.30 & 62.99 & \underline{54.06} & 57.98 & 58.05 & 49.61$\times$ \\
\textsc{NegGrad} & 94.11 & 11.09 & \underline{67.29} & \textbf{46.63} & 60.96 & 44.12 & 53.86$\times$ \\
\textsc{Finetuning} & 76.50 & 23.19 & 37.61 & 63.32 & 12.23 & 81.50 & 13.72$\times$ \\
\bottomrule
\end{tabular*}
\end{table}
Table~\ref{tab:mean-performance-10} shows that effective machine unlearning can be achieved at a fraction of retraining cost. \textsc{TraceHiding~(Ent.)} is Pareto--optimal, deviating on average by only $\approx6$\,pp from the retraining gold standard while delivering a $40\times$ speed-up.  The other \textsc{TraceHiding} variants follow closely, confirming that the method family consistently balances privacy (UA) and utility (TA).  In contrast, gradient–reversal approaches (\textsc{NegGrad+}) push runtime gains to $50\times$ but at the expense of large TA drops.  \textsc{Finetuning} sits at the opposite extreme: it retains high TA yet leaks deleted information (UA falls to $12\%$ on \emph{HO-NYC}) and is the slowest alternative ($14\times$). Overall, high-fidelity unlearning is readily attainable on simpler domains such as \emph{HO-Rome}, whereas complex, high-utility datasets like \emph{HO-NYC} still expose a tension between speed, privacy, and performance.

\subsubsection{Model Architecture Sensitivity}
\label{sec:model_sesitivity}

\begin{table}[t]
\caption[RNN, Transformer, and GCN Unlearning Accuracy at 10\,\% Deletion]{Comparative Unlearning Accuracy at 10\% Deletion for RNN, Transformer, and GCN Architectures under Various Unlearning Methods (Uniform Sampling). Best and second-best accuracies for each model are indicated by \textbf{bold} and \underline{underline}, respectively.}
\label{tab:ua-by-model}
\centering
\begin{tabular*}{\linewidth}{@{\extracolsep{\fill}}lccccc}
\toprule
\textbf{Method} & GRU & LSTM & BERT & ModernBERT & GCN-TULHOR \\
\midrule
\textsc{Bad-T}              & 51.57 & 47.52 & 65.62 & 60.47 & 95.04 \\
\textsc{Finetuning}         & 46.08 & 44.03 & 35.49 & 39.34 & 45.63 \\
\textsc{NegGrad}            & 57.88 & \textbf{56.83} & \underline{92.46} & 63.43 & \textbf{100.00} \\
\textsc{NegGrad+}           & 57.23 & 53.20 & 81.25 & 60.62 & \textbf{100.00} \\
\textsc{SCRUB}              & 54.09 & 50.19 & 79.96 & 55.91 & \textbf{100.00} \\
\textsc{TraceHiding (C.D.)} & \underline{60.52} & 55.62 & 83.74 & 67.45 & 74.36 \\
\textsc{TraceHiding (Ent.)} & \textbf{61.39} & \underline{56.07} & \textbf{93.48} & \textbf{81.37} & 88.72 \\
\textsc{TraceHiding (Uni.)} & 59.90 & 53.15 & 81.44 & \underline{70.33} & 84.33 \\
\bottomrule
\end{tabular*}
\end{table}

The presented table, Table~\ref{tab:ua-by-model}, evaluates the unlearning accuracy of various methods applied to different machine learning architecturesl, Recurrent Neural Networks (RNNs) like GRU and LSTM, Transformer models such as BERT and ModernBERT, and the Graph Convolutional Network (GCN-TULHOR), all under a fixed 10\% data deletion sample size using uniform sampling. Analyzing the unlearning methods, \textsc{TraceHiding (Ent.)} consistently emerges as a highly effective approach across diverse architectures. It achieves the best performance for GRU (61.39\%), BERT (93.48\%), and ModernBERT (81.37\%), and secures the second-best accuracy for LSTM (56.07\%) and GCN-TULHOR (88.72\%), demonstrating its robust applicability. Similarly, \textsc{NegGrad} exhibits strong efficacy, particularly excelling with LSTM (achieving the best 56.83\%) and contributing to the perfect 100\% unlearning scores for GCN-TULHOR, while also being the second-best method for BERT (92.46\%). \textsc{NegGrad+} and \textsc{SCRUB} also achieve perfect 100\% unlearning accuracy for GCN-TULHOR, indicating their strong performance in specific architectural contexts. In contrast, \textsc{Finetuning} consistently underperforms all other methods, yielding the lowest unlearning accuracies across all models (ranging from 35.49\% to 46.08\%), which underscores its general inadequacy as a standalone unlearning strategy. \textsc{Bad-T} shows moderate performance, while \textsc{TraceHiding (C.D.)} and \textsc{TraceHiding (Uni.)} offer competitive, though generally not leading, results. The data suggests that while some methods like \textsc{TraceHiding (Ent.)} and \textsc{NegGrad} are broadly effective, the optimal unlearning method can still be architecture-dependent, as evidenced by GCN-TULHOR's unique ability to achieve perfect unlearning with multiple methods. It is crucial to note, however, that unlearning accuracy is not the sole objective; the preservation of model utility on retained data is equally vital for practical applications.

\subsection{RQ3: Impact of Sampling Strategies}
\label{sec:RQ_deletion_sampling}
In this section, we evaluate how sampling strategies affect unlearning performance. First, we assess robustness under \textit{targeted sampling}, where high-information users are removed. Next, we analyze the effect of \textit{deletion sample size} on utility and privacy. These results reveal the strengths and limitations of methods in adversarial and practical settings.
\subsubsection{Unlearning under Targeted Sampling}
\label{sec:eval_targeted_sampling}

Targeted sampling removes the most important users, ranked by entropy. Table~\ref{tab:delta-mia} shows the change in MIA AUC (relative to uniform sampling) under a 10\% deletion scenario. A lower MIA AUC indicates better forgetting, meaning the attack model is less able to distinguish the unlearned dataset.
Since we are removing high-information users, we expect models to struggle more with unlearning in this setting. The metric in Table~\ref{tab:delta-mia} allows us to compare the robustness of different unlearning methods under more adversarial deletion conditions.
\begin{table}[t]
\caption[MIA AUC change at 10\,\% unlearning]{Change in MIA AUC ($\Delta$ = \textit{targeted} – \textit{uniform})  
         at 10\,\% unlearning (mean over datasets/models; $\downarrow$ better).}
\label{tab:delta-mia}
\centering
\begin{tabular*}{\linewidth}{@{\extracolsep{\fill}} l c c c}
\toprule
Method & MIA AUC (Targeted) & MIA AUC (Uniform) & $\Delta$ \\
\midrule
\textsc{TraceHiding (Ent.)} & 31.53 & 35.73 & -4.21 \\
\textsc{TraceHiding (C.D.)} & 31.86 & 34.79 & -2.93 \\
\textsc{NegGrad}            & 32.51 & 33.58 & -1.07 \\
\textsc{Finetuning}         & 37.27 & 36.78 & 0.49 \\
\textsc{NegGrad+}           & 33.38 & 32.30 & 1.08 \\
\textsc{Bad-T}              & 36.32 & 34.64 & 1.68 \\
\textsc{SCRUB}              & 36.54 & 34.04 & 2.49 \\
\textsc{TraceHiding (Uni.)} & 37.08 & 34.22 & 2.86 \\
\bottomrule
\end{tabular*}
\end{table}
Table~\ref{tab:delta-mia} demonstrates that the \textsc{TraceHiding} variants perform best under targeted sampling. Specifically, \textsc{TraceHiding (Ent.)} shows the largest reduction in MIA AUC with a $\Delta$ of -4.21\%, indicating superior forgetting when high-information users are removed. \textsc{TraceHiding (C.D.)} also performs well with a -2.93\% change.
\textsc{NegGrad} shows a modest improvement (-1.07\%), while \textsc{Finetuning} slightly worsens under targeted sampling, with a positive change of 0.49\%. \textsc{NegGrad+}, \textsc{Bad-T}, \textsc{SCRUB}, and \textsc{TraceHiding (Uni.)} all have positive $\Delta$ values, meaning their performance degrades under targeted sampling. \textsc{SCRUB} and \textsc{TraceHiding (Uni.)} in particular suffer the most, with increases of 2.49\% and 2.86\% in MIA AUC, respectively. This suggests that these methods are more sensitive to the deletion of high-information users and less effective in such scenarios. Notably, while \textsc{TraceHiding (Uni.)} is a variant of the trace hiding approach, it performs poorly under targeted sampling, suggesting that the unified importance score may be less effective at identifying and mitigating the impact of high-information users in adversarial deletion scenarios.
Overall, targeted unlearning stresses the unlearning methods differently, revealing \textsc{TraceHiding} variants, especially the entropy-based one, as the most resilient to adversarial deletions.


\subsubsection{Effect of the Deletion Sample Size}
\label{sec:effect_sample_size}

Our analysis reveals a distinct trade-off between maintaining model utility and the efficacy of the unlearning process. As shown in Table~\ref{tab:sample-size-trend}, the \textsc{TraceHiding (Ent.)} method consistently preserves model utility more effectively than \textsc{SCRUB}. Across nearly all datasets and sample sizes, \textsc{TraceHiding (Ent.)} yields superior Test Accuracy (TA) and Remaining Accuracy (RA), indicating a less destructive impact on the model's general performance and knowledge learned from the remaining data.

From a privacy perspective, measured by the (MIA), the \textsc{SCRUB} method demonstrates a general superiority. Lower MIA scores indicate a more successful defense against attacks aiming to determine if a specific data point was part of the model's training set. Across the HO-Geolife and HO-NYC datasets, \textsc{SCRUB} consistently achieves lower (better) MIA scores than \textsc{TraceHiding (Ent.)}. For example, on HO-Geolife, \textsc{SCRUB}'s advantage in MIA mitigation is evident across all unlearning percentages. While both methods show improved privacy protection (lower MIA scores) as the proportion of unlearned data increases, \textsc{SCRUB}'s more aggressive unlearning procedure appears more effective at obfuscating the membership status of the forgotten samples. This highlights a critical insight: the factors that lead to a reduction in utility in \textsc{SCRUB} are directly contributing to its enhanced privacy protection.

\begin{table}[t]
\caption[Comparison of unlearning methods' performance metrics]{A quantitative comparison of four metrics across the datasets and varying sample sizes (1\%, 5\%, 10\%, 20\%) for two strong candidates for unlearning method: \textsc{TraceHiding (Ent.)} and \textsc{SCRUB}. The \textbf{bold} values indicate the performance closer to a retrained model, signifying better outcomes for each respective metric.}
\label{tab:sample-size-trend}
\centering

\begin{tabular*}{\linewidth}{@{\extracolsep{\fill}} cl
                *{4}{c} 
                *{4}{c} 
                }
\toprule
\multirow{2}{*}{\textbf{Dataset}} & \multirow{2}{*}{\textbf{Metric}} 
    & \multicolumn{4}{c}{\textsc{TraceHiding (Ent.)}}
    & \multicolumn{4}{c}{\textsc{SCRUB}} \\
\cmidrule(lr){3-6} \cmidrule(lr){7-10}
    & 
    & 1\,\% & 5\,\% & 10\,\% & 20\,\%
    & 1\,\% & 5\,\% & 10\,\% & 20\,\% \\
\midrule
\multirow{4}{*}{HO-Rome}
    & UA & \textbf{84.80} & \textbf{87.62} & \textbf{89.82} & 92.09 & 88.11 & 86.30 & 84.32 & \textbf{88.10} \\
    & RA & \textbf{42.06} & \textbf{36.94} & \textbf{32.46} & \textbf{26.65} & 33.13 & 29.42 & 29.50 & 26.56 \\
    & TA & \textbf{22.85} & \textbf{20.74} & \textbf{18.57} & 15.08 & 19.18 & 17.60 & 17.44 & \textbf{15.37} \\
    & MIA & \textbf{33.39} & \textbf{22.17} & 18.80 & \textbf{14.96} & 35.28 & 24.60 & \textbf{17.09} & 15.17 \\
\midrule
\multirow{4}{*}{HO-Geolife}
    & UA & \textbf{76.38} & \textbf{67.24} & \textbf{70.20} & \textbf{69.68} & 73.99 & 64.86 & 57.81 & 59.51 \\
    & RA & 83.26 & 82.85 & 80.90 & 80.65 & \textbf{80.07} & \textbf{78.55} & \textbf{77.52} & \textbf{75.67} \\
    & TA & 61.35 & \textbf{62.14} & \textbf{56.98} & 55.10 & \textbf{60.48} & 60.34 & 57.17 & \textbf{54.41} \\
    & MIA & 47.58 & 44.07 & 44.61 & 42.80 & \textbf{46.26} & \textbf{39.84} & \textbf{41.02} & \textbf{41.15} \\
\midrule
\multirow{4}{*}{HO-NYC}
    & UA & \textbf{50.77} & 54.18 & \textbf{68.61} & \textbf{74.43} & 48.61 & \textbf{58.67} & 61.96 & 67.50 \\
    & RA & \textbf{95.04} & \textbf{93.87} & \textbf{92.98} & \textbf{92.51} & 87.34 & 81.48 & 81.36 & 80.83 \\
    & TA & \textbf{79.92} & \textbf{77.75} & \textbf{73.36} & \textbf{68.27} & 73.89 & 67.99 & 65.92 & 61.26 \\
    & MIA & 49.30 & 46.81 & \textbf{43.80} & 41.49 & \textbf{48.17} & \textbf{46.71} & 44.02 & \textbf{40.94} \\
\bottomrule
\end{tabular*}
\end{table}

\subsection{RQ4: Balancing Accuracy and Efficiency}
\label{sec:RQ_tradeoffs_speed}
\label{sec:speed_up}
Achieving optimal computational efficiency while maintaining high levels of accuracy is a persistent challenge in the field of machine learning.
When considering computational methods, one must often balance between accuracy and efficiency. In this subsection, we examine this trade-off in the context of two specific methods: SCRUB and TraceHiding (Ent.).
Their performance in this regard is detailed in Table~\ref{tab:speedup-comparison}, neither method holds a universal advantage, with the choice depending on the unlearning task's scale. \textsc{SCRUB} generally offers a significant speedup for smaller unlearning requests (e.g., 1\% and 5\%), achieving up to 85.0$\times$ the speed of retraining on the HO-Rome dataset. However, its efficiency tends to decrease as the percentage of data to be unlearned grows. In contrast, \textsc{TraceHiding (Ent.)} demonstrates more consistent, and in some cases superior, speedup for larger unlearning tasks, particularly on the HO-NYC dataset. When cross-referencing with our utility analysis (Table~\ref{tab:sample-size-trend}), a clear conclusion emerges: \textsc{SCRUB} provides a rapid solution for privacy-critical scenarios that demand aggressive unlearning, but this speed comes at the cost of model utility. \textsc{TraceHiding (Ent.)}, while not always the fastest, presents a more balanced profile, offering competitive computational performance without a substantial degradation in model accuracy. This makes it a compelling choice for applications where maintaining high utility is as critical as the efficiency of the unlearning process itself.

\begin{table}[t]
\caption[Speedup comparison of unlearning methods]{Comparison of speedup ($\times$ faster than retraining) achieved by two strong unlearning candidates, \textsc{TraceHiding (Ent.)} and \textsc{SCRUB}, across three datasets and varying sample sizes (1\%, 5\%, 10\%, 20\%). \textbf{Bold} values highlight the higher speedup between methods for a given configuration.}
\label{tab:speedup-comparison}
\centering

\begin{tabular*}{\linewidth}{@{\extracolsep{\fill}}clcccc}
\toprule
\textbf{Dataset} & \textbf{Method} & \textbf{1\%} & \textbf{5\%} & \textbf{10\%} & \textbf{20\%} \\
\midrule
\multirow{2}{*}{HO-Rome}
    & \textsc{TraceHiding (Ent.)} & 46.8$\times$ & 43.0$\times$ & 37.4$\times$ & \textbf{43.6$\times$} \\
    & \textsc{SCRUB}              & \textbf{85.0$\times$} & \textbf{50.6$\times$} & \textbf{38.8$\times$} & 32.3$\times$ \\
\midrule
\multirow{2}{*}{HO-Geolife}
    & \textsc{TraceHiding (Ent.)} & 16.6$\times$ & 17.3$\times$ & \textbf{20.6$\times$} & \textbf{17.5$\times$} \\
    & \textsc{SCRUB}              & \textbf{29.1$\times$} & \textbf{21.0$\times$} & 16.2$\times$ & 14.7$\times$ \\
\midrule
\multirow{2}{*}{HO-NYC}
    & \textsc{TraceHiding (Ent.)} & \textbf{56.1$\times$} & \textbf{72.5$\times$} & 74.5$\times$ & 50.3$\times$ \\
    & \textsc{SCRUB}              & 54.7$\times$ & 71.3$\times$ & \textbf{80.3$\times$} & \textbf{73.0$\times$} \\
\bottomrule
\end{tabular*}
\end{table}

\section{Conclusions}
\label{sec:conclusion}

This work addressed the challenge of \textbf{machine unlearning for trajectory data} in privacy-sensitive mobility analytics (e.g., requests from users to erase their data). We introduced \textsc{TraceHiding}, the first \textit{importance-aware} unlearning framework for trajectory classification tasks such as Trajectory–User Linking (TUL). The \textbf{key idea} of our approach was to selectively guide the unlearning process by quantifying the significance of individual data samples, thereby aiming for a more efficient and effective removal of user-specific information while preserving overall model utility.
In the rest of this section, we present the \textbf{key empirical findings} from our evaluations across datasets, models, and deletion scenarios, highlight the \textbf{core technical contributions} of the \textsc{TraceHiding} framework, outline promising \textbf{avenues for future research}, and discuss the \textbf{broader implications} of our results for advancing privacy-preserving machine learning and responsible AI in mobility analytics and related domains.

\subsection{Key Empirical Findings}

Our comprehensive experimental evaluation yielded several key empirical findings:

\smallskip\noindent\textbf{RQ1: Importance Score Impact:} The choice and formulation of the importance score were shown to be critical. The entropy-based importance score, \textsc{TraceHiding (Ent.)}, generally offered the best balance in achieving high UA and maintaining TA, closely mirroring retraining results, especially for datasets like HO-Geolife and HO-NYC. Other scores, like the unified score or coverage diversity, sometimes provided slight advantages in MIA resilience, though differences in MIA were often negligible among \textsc{TraceHiding} variants. The information in Section~\ref{sec:choosing_imortance_score} and analysis in Section~\ref{sec:importance_score_impact} detailed how different formulations affected performance.

\smallskip\noindent\textbf{RQ2: Enhanced Capabilities:} The \textsc{TraceHiding} method, particularly the entropy-based variant (\textsc{TraceHiding (Ent.)}), demonstrated superior or competitive performance in Unlearning Accuracy (UA) compared to baseline methods like \textsc{NegGrad}, \textsc{NegGrad+}, \textsc{SCRUB}, \textsc{Bad-T}, and \textsc{Finetuning} across various datasets and model architectures under uniform user deletion scenarios. It often provided the closest approximation to full retraining in terms of Unlearning Accuracy (UA) and Test Accuracy (TA) while being significantly faster.

\smallskip\noindent\textbf{RQ3: Sampling Strategy Effects:} When evaluating unlearning under targeted sampling, \textsc{TraceHiding (Ent.)} and \textsc{TraceHiding (C.D.)} proved more resilient, showing better forgetting (lower or improved MIA AUC) compared to uniform sampling. This contrasted with several other methods whose effectiveness diminished under such adversarial deletion conditions.

\smallskip\noindent\textbf{RQ4: Unlearning Accuracy vs. Computational Cost Trade-offs:} \textsc{TraceHiding} methods, particularly \textsc{TraceHiding (Ent.)}, generally achieved a good balance by providing UA and TA close to retraining but with approximately 40x speedup. Compared to \textsc{NegGrad+}, \textsc{TraceHiding (Ent.)} often excelled in UA, TA, and speedup for larger deletion percentages, while \textsc{NegGrad+} sometimes offered better MIA resilience and speed for small deletion requests (e.g., 1,\%).

\smallskip\noindent\textbf{Scalability.} Moreover, \textsc{TraceHiding} is designed with scalability in mind. Since the importance scores are computed solely from static characteristics of the training data (e.g., entropy, trajectory length, and coverage diversity), they can be precomputed during data preprocessing and stored efficiently. In settings involving large training data, these scores typically do not need to be recomputed unless there is a significant shift in the data distribution. Instead, the unlearning process can retrieve them on demand using well-established data structures, such as hash maps or tree-based indices, based on trajectory or user identifiers. As a result, the scalability challenge shifts from computational overhead to efficient data retrieval and storage, a well-understood problem in systems and database design.

\subsection{Core Technical Contributions}

In this work, we present \textsc{TraceHiding}, an importance-aware unlearning framework that leverages hierarchical importance scores to selectively target high-impact trajectories during the unlearning process. To achieve this, we introduce scalable, model-agnostic metrics at the token, trajectory, and user levels—encompassing factors such as coverage diversity, entropy, length, and uniqueness—and unify them into normalized importance scores. Building on these scores, we adapt a teacher--student distillation approach to simultaneously maximize forgetting on the unlearning set while preserving knowledge on the remaining data, with per-sample weighting guided by importance. We validate our framework through extensive experiments on three large-scale trajectory datasets—\textsc{HO-Rome}, \textsc{HO-Geolife}, and \textsc{HO-NYC}—and across diverse model architectures, including \textsc{GRU}, \textsc{LSTM}, \textsc{BERT}, \textsc{ModernBERT}, and \textsc{GCN-TULHOR}, demonstrating that \textsc{TraceHiding} outperforms existing machine unlearning baselines in terms of unlearning accuracy, utility preservation, and runtime efficiency. To support reproducibility and facilitate further research, we also release all source code, preprocessing pipelines, and benchmark datasets.

\subsection{Future Research Directions}

While \textsc{TraceHiding} has shown promising results in trajectory-based unlearning, several directions remain open for future research. These directions explore extensions to new trajectory data mining tasks, richer data representations, real-time scenarios, and broader social considerations such as fairness and regulatory compliance.

\subsubsection{Beyond TUL: Other Predictive Tasks}  
Future work could explore how the \textsc{TraceHiding} framework generalizes to other location-based  generative or predictive tasks such as next-location prediction, transportation mode inference, or mobility anomaly detection. These tasks differ in their learning objectives and may require adapting the loss function or importance metrics. Testing across such use cases can offer deeper insight into the versatility and robustness of importance-aware unlearning.

\subsubsection{Richer Importance Signals}  
The current implementation of importance scores focuses on entropy, length, and diversity. Future research might incorporate semantic context, such as Points of Interest (POI) types, user demographics, or even textual annotations, to better estimate the privacy sensitivity of samples. Moreover, integrating gradient-based influence functions could help capture a more causal relationship between individual data points and model predictions, enabling even finer control during unlearning.

\subsubsection{Streaming and Continual Settings}  
Deploying unlearning methods like \textsc{TraceHiding} in real-world applications requires addressing the challenges of streaming and continual learning. In such settings, user trajectories arrive in real time, necessitating fast, memory-efficient updates and the ability to handle concept drift. Maintaining unlearning guarantees under resource-constrained or dynamic environments will require new algorithmic strategies and optimizations.

\subsubsection{Fairness and Compliance Analysis}  
Finally, any practical deployment of machine unlearning must ensure that the process does not inadvertently introduce biases or disproportionately affect certain user groups. Future research should involve fairness audits to monitor group-level performance shifts post-unlearning. Additionally, aligning unlearning procedures with evolving legal frameworks such as the GDPR and CCPA is essential for real-world applicability in sensitive domains like healthcare or finance.

\medskip\noindent\textbf{Important Note.} It is important to note that these future directions constitute substantial extensions and are therefore beyond the scope of the current work. They are included here for completeness and to help chart a path for further research on this important yet largely overlooked topic.

\subsection{Broad Implications}
\textsc{TraceHiding} advances the state of the art in \textbf{privacy-preserving mobility analytics} by showing that accurate, efficient, and targeted unlearning is possible without full retraining. The hierarchical importance-driven approach bridges the gap between data protection mandates and operational feasibility, enabling compliance with ``right-to-be-forgotten'' regulations while safeguarding model utility. Beyond mobility, the framework’s principles are transferable to other spatiotemporal and sequence-learning domains, such as healthcare, finance, or recommendation systems, where selective forgetting is critical for \textbf{responsible and trustworthy AI}.

\clearpage

\bibliographystyle{ACM-Reference-Format}
\bibliography{refs}

\appendix
\clearpage
\noindent {\LARGE\textbf{APPENDIX}}

\section{Results Under Uniform Random Sampling} \label{appndx:eval_uniform_sampling}

In this section, we present detailed results evaluating unlearning methods under the setting of uniform random sampling of users for deletion. Specifically, users are selected independently and uniformly at random from the dataset to simulate deletion requests. This evaluation setting provides a controlled and unbiased scenario to measure the unlearning performance across a representative subset of the user population. It allows us to assess the robustness, consistency, and utility preservation of the unlearning approach without introducing bias from specific user attributes or behaviors.

\begin{table}[!htbp]
\centering
\caption[The results for the HO-Rome dataset using uniform sampling]{The results for the HO-Rome dataset using uniform sampling. All numbers are presented as percentages.}\label{tab:results_rome_uniform_sampling}

\resizebox{\textwidth}{!}{
\begin{tabular}{lcccccccccccccccc}
\toprule
\rowcolor{lightgray}
\multicolumn{17}{@{}c@{}}{\textbf{HO‑Rome Dataset}} \\
\midrule
Sample Size & \multicolumn{4}{c}{1 \%} & \multicolumn{4}{c}{5 \%} & \multicolumn{4}{c}{10 \%} & \multicolumn{4}{c}{20 \%} \\
\cmidrule(lr){2-5} \cmidrule(lr){6-9} \cmidrule(lr){10-13} \cmidrule(lr){14-17}
Methods & UA & RA & TA & MIA & UA & RA & TA & MIA & UA & RA & TA & MIA & UA & RA & TA & MIA \\
\midrule

\multicolumn{17}{c}{\textbf{GRU}}\\ \midrule
\rowcolor{gold!30}
\textsc{Retraining} & 100  &  0.72  &  0.73  &  0  &  100  &  0.66  &  0.64  &  0.61  &  100  &  1.33  &  1.2  &  0.73  &  100  &  0.83  &  0.61  &  3.14 \\
\textsc{Finetuning} & 98.52  &  5.74  &  6.17  &  24.3  &  92.28  &  5.61  &  6.23  &  12.84  &  94.2  &  5.72  &  6.23  &  6.07  &  94.9  &  5.82  &  6.17  &  4.58 \\
\textsc{NegGrad} & 100  &  5.58  &  6.05  &  24.1  &  99.84  &  4.84  &  5.12  &  12.15  &  98.09  &  5.15  &  5.12  &  5.99  &  96.38  &  5.92  &  5.91  &  4.71 \\
\textsc{NegGrad+} & 100  &  5.51  &  5.56  &  24.32  &  99.84  &  4.9  &  5.23  &  11.95  &  97.44  &  5.59  &  5.64  &  6.08  &  96.5  &  5.97  &  6.08  &  4.45 \\
\textsc{Bad-T} & 98.52  &  5.71  &  6.14  &  24.29  &  92.12  &  5.56  &  6.2  &  12.83  &  94.13  &  5.65  &  6.35  &  6.1  &  94.9  &  5.78  &  6.26  &  4.58 \\
\textsc{SCRUB} & 99.26  &  5.72  &  6.14  &  24.28  &  92.28  &  5.57  &  6.2  &  12.9  &  94.12  &  5.67  &  6.23  &  5.96  &  94.97  &  5.8  &  6.17  &  4.41 \\
\hdashline
\textsc{TraceHiding (Ent.)} & 99.26  &  5.69  &  6.17  &  23.67  &  95.19  &  5.45  &  5.88  &  11.94  &  94.83  &  5.63  &  5.85  &  5.45  &  95.14  &  5.8  &  6.02  &  4.42 \\
\textsc{TraceHiding (C.D.)} & 99.26  &  5.71  &  6.08  &  23.73  &  95.5  &  5.57  &  5.99  &  11.44  &  95.35  &  5.71  &  6.2  &  4.86  &  95.22  &  5.77  &  5.99  &  4.15 \\
\textsc{TraceHiding (Uni.)} & 99.26  &  5.73  &  6.14  &  22.6  &  95.98  &  5.63  &  6.05  &  12.71  &  94.98  &  5.82  &  6.14  &  5.87  &  95.18  &  5.78  &  5.99  &  4.06 \\
\multicolumn{17}{c}{\textbf{LSTM}} \\ \midrule
\rowcolor{gold!30}
\textsc{Retraining} & 100  &  6.37  &  3.98  &  4.25  &  100  &  0.73  &  0.7  &  5.89  &  100  &  0.79  &  0.7  &  2.7  &  100  &  0.85  &  0.67  &  5.85 \\
\textsc{Finetuning} & 98.52  &  11.1  &  6.84  &  22.42  &  90.56  &  11.08  &  6.84  &  14.57  &  89.08  &  11.02  &  6.73  &  7.69  &  90.14  &  11.35  &  6.81  &  8.24 \\
\textsc{NegGrad} & 99.26  &  11.18  &  6.93  &  20.69  &  96.85  &  11.26  &  6.75  &  14.52  &  95.22  &  11.18  &  6.78  &  7.93  &  94.72  &  11.73  &  6.35  &  8.46 \\
\textsc{NegGrad+} & 99.26  &  11.24  &  6.9  &  22.23  &  97.01  &  11.3  &  6.99  &  14.3  &  94.92  &  11.49  &  6.64  &  8  &  94.53  &  11.89  &  6.26  &  8.42 \\
\textsc{Bad-T} & 98.52  &  11.1  &  6.87  &  22.41  &  90.4  &  11.13  &  6.93  &  14.53  &  89.16  &  11.16  &  6.99  &  7.65  &  89.96  &  11.53  &  6.99  &  8.29 \\
\textsc{SCRUB} & 98.52  &  11.1  &  6.73  &  22.42  &  90.56  &  11.07  &  6.84  &  14.59  &  89.08  &  11.05  &  6.73  &  7.67  &  90.07  &  11.38  &  6.81  &  8.26 \\
\hdashline
\textsc{TraceHiding( Ent.)} & 98.52  &  11.08  &  6.75  &  22.3  &  94.47  &  11.19  &  6.61  &  13.68  &  91.21  &  11.17  &  6.7  &  6.61  &  95.6  &  11.92  &  6.05  &  7.09 \\
\textsc{TraceHiding (C.D.)} & 98.52  &  11.13  &  6.78  &  22.55  &  94.49  &  11.17  &  6.67  &  12.16  &  91.52  &  11.38  &  6.81  &  6.76  &  95.15  &  11.65  &  6.23  &  8.43 \\
\textsc{TraceHiding (Uni.)} & 98.52  &  11.08  &  6.78  &  21.57  &  95.26  &  11.2  &  6.75  &  12.02  &  91.72  &  11.04  &  6.78  &  6.27  &  94.64  &  11.63  &  6.08  &  7.86 \\
\multicolumn{17}{c}{\textbf{BERT}} \\
\midrule
\rowcolor{gold!30}
\textsc{Retraining} & 100  &  88.74  &  39.65  &  31.35  &  100  &  87.76  &  38.42  &  26.96  &  100  &  84.55  &  36.61  &  20.7  &  100  &  85.36  &  32.34  &  18.12 \\
\textsc{Finetuning} & 35.65  &  81.54  &  38.04  &  48.9  &  60.63  &  82.44  &  38.1  &  43.6  &  58.48  &  82.29  &  37.66  &  36.82  &  54.16  &  82.92  &  37.02  &  35.39 \\
\textsc{NegGrad} & 86.2  &  76.09  &  36.87  &  48.35  &  80.6  &  52.58  &  29.77  &  39.77  &  97.59  &  7.54  &  4.74  &  36.75  &  99.28  &  0.64  &  0.64  &  27.5 \\
\textsc{NegGrad+} & 69.72  &  80.49  &  38.04  &  49.23  &  72.05  &  73.06  &  36.32  &  42.1  &  82.58  &  60.91  &  31.08  &  32.03  &  88.72  &  41.01  &  21.67  &  31.7 \\
\textsc{Bad-T} & 75.19  &  79.36  &  37.98  &  49.25  &  65.36  &  78.78  &  37.25  &  42.84  &  62.31  &  78.82  &  36.64  &  40.65  &  53.69  &  78.19  &  35.67  &  42.86 \\
\textsc{SCRUB} & 79.54  &  77.72  &  37.08  &  47.77  &  82.94  &  67.38  &  34.27  &  42.74  &  70.4  &  67.86  &  34.33  &  38.36  &  86.53  &  53.7  &  25.94  &  32.48 \\
\hdashline
\textsc{TraceHiding (Ent.)} & 86.39  &  78.17  &  37.08  &  45.6  &  85.59  &  68.25  &  34.36  &  38.79  &  89.56  &  55.98  &  29.94  &  36.3  &  94.91  &  43.64  &  21.73  &  33.55 \\
\textsc{TraceHiding (C.D.)} & 78.61  &  80.49  &  37.81  &  48.11  &  72.24  &  76.4  &  37.16  &  42.85  &  83.77  &  71.14  &  34.33  &  33.48  &  82.5  &  70.85  &  31.29  &  31.03 \\
\textsc{TraceHiding (Uni.)} & 95.28  &  79.97  &  37.92  &  48.1  &  76.05  &  74.77  &  35.67  &  43.12  &  75.8  &  70.13  &  34.36  &  34.1  &  80.55  &  64.91  &  30.41  &  31.83 \\
\multicolumn{17}{c}{\textbf{ModernBERT}} \\
\midrule
\rowcolor{gold!30}
\textsc{Retraining} & 100  &  69.38  &  39.94  &  34.27  &  100  &  67.02  &  38.77  &  26.36  &  100  &  63.88  &  36.78  &  13.08  &  100  &  57  &  32.4  &  18.76 \\
\textsc{Finetuning} & 62.22  &  59.56  &  37.49  &  4.89  &  65.61  &  59.16  &  37.72  &  25.28  &  64.51  &  61.21  &  36.9  &  28.43  &  64.04  &  61.07  &  35.76  &  19.5 \\
\textsc{NegGrad} & 72.59  &  62.25  &  40.38  &  45.43  &  77.91  &  62.5  &  39.47  &  15.67  &  79.66  &  63.03  &  38.25  &  27.06  &  78.83  &  62.54  &  36.11  &  11.43 \\
\textsc{NegGrad+} & 76.67  &  61.98  &  40.12  &  13.7  &  75.29  &  61.91  &  39.27  &  28.12  &  77.07  &  61.92  &  37.6  &  16.97  &  75.31  &  62.18  &  36.32  &  29.04 \\
\textsc{Bad-T} & 75.28  &  62.55  &  40.2  &  37.3  &  70.39  &  62.97  &  38.68  &  34.97  &  68.8  &  63.88  &  37.95  &  17.74  &  66.37  &  64.04  &  35.38  &  27.45 \\
\textsc{SCRUB} & 63.24  &  61.21  &  39.56  &  48.14  &  65.74  &  59.98  &  38.36  &  35  &  67.97  &  60.12  &  37.72  &  19.16  &  68.93  &  59.19  &  36.23  &  17.7 \\
\hdashline
\textsc{TraceHiding (Ent.)} & 47.59  &  58.63  &  38.25  &  34.36  &  70.78  &  56.69  &  36.87  &  25.83  &  77.64  &  55.9  &  35.03  &  26.79  &  76.23  &  53.72  &  32.63  &  11.9 \\
\textsc{TraceHiding (C.D.)} & 62.13  &  58.54  &  38.36  &  34.72  &  67.93  &  58.59  &  37.92  &  25.38  &  70.03  &  57.23  &  35.56  &  20.43  &  70.44  &  58.36  &  34.74  &  28.12 \\
\textsc{TraceHiding (Uni.)} & 44.91  &  58.37  &  38.48  &  46.08  &  66.75  &  57.36  &  36.67  &  32.71  &  69.8  &  57.38  &  35.94  &  10.34  &  73.95  &  57.1  &  34.18  &  11.02 \\
\multicolumn{17}{c}{\textbf{GCN-TULHOR}} \\ \midrule
\rowcolor{gold!30}
\textsc{Retraining} & 23.7  &  65.48  &  28.22  &  46.82  &  36  &  65.84  &  28.1  &  27.36  &  38.33  &  65.98  &  27.92  &  17.37  &  34.35  &  65.48  &  27.75  &  18.93 \\
\textsc{Finetuning} & 76.48  &  69.03  &  30.7  &  41.84  &  78.68  &  67.44  &  28.8  &  19.27  &  76.24  &  67.61  &  28.45  &  20.2  &  77.29  &  70.65  &  27.66  &  18.68 \\
\textsc{NegGrad} & 100  &  1.11  &  0.91  &  29.42  &  100  &  0.49  &  0.44  &  7.5  &  100  &  0.54  &  0.56  &  6.38  &  100  &  0.58  &  0.47  &  4.76 \\
\textsc{NegGrad+} & 100  &  1.34  &  1.08  &  34.37  &  100  &  0.62  &  0.56  &  6.96  &  100  &  0.68  &  0.56  &  5.59  &  100  &  0.67  &  0.56  &  5.32 \\
\textsc{Bad-T} & 100  &  35.64  &  18.39  &  37.17  &  97.37  &  21.44  &  11.55  &  19.17  &  95.28  &  20.5  &  11.46  &  16.87  &  94.32  &  21.77  &  12.13  &  16.14 \\
\textsc{SCRUB} & 100  &  9.92  &  6.4  &  33.77  &  100  &  3.08  &  2.34  &  17.79  &  100  &  2.78  &  2.19  &  14.31  &  100  &  2.7  &  1.73  &  13.02 \\
\hdashline
\textsc{TraceHiding (Ent.)} & 92.22  &  56.71  &  25.99  &  41.03  &  92.06  &  43.12  &  20  &  20.62  &  95.84  &  33.62  &  15.35  &  18.83  &  98.56  &  18.19  &  8.95  &  17.83 \\
\textsc{TraceHiding (C.D.)} & 87.04  &  56.86  &  26.61  &  41.35  &  88.13  &  52.63  &  23.8  &  18.4  &  86.4  &  53.01  &  24.18  &  17.28  &  86.28  &  52.13  &  22.63  &  16.81 \\
\textsc{TraceHiding (Uni.)} & 89.26  &  55.85  &  26.11  &  41.25  &  91.87  &  49.28  &  22.98  &  21.63  &  87.12  &  49.06  &  21.93  &  19.49  &  88.79  &  47.7  &  21.14  &  17.12 \\
\bottomrule
\end{tabular}
}
\end{table}

\begin{table}[!htbp]

\centering
\caption[The results for the HO-Geolife dataset using uniform sampling]{The results for the HO-Geolife dataset using uniform sampling. All numbers are presented as percentages.}\label{tab:results_geolife_uniform_sampling}

\resizebox{\textwidth}{!}{
\begin{tabular}{lcccccccccccccccc}
\toprule
\rowcolor{lightgray}
\multicolumn{17}{@{}c@{}}{\textbf{HO‑Geolife Dataset}} \\
\midrule
Sample Size & \multicolumn{4}{c}{1 \%} & \multicolumn{4}{c}{5 \%} & \multicolumn{4}{c}{10 \%} & \multicolumn{4}{c}{20 \%} \\
\cmidrule(lr){2-5} \cmidrule(lr){6-9} \cmidrule(lr){10-13} \cmidrule(lr){14-17}
Methods & UA & RA & TA & MIA & UA & RA & TA & MIA & UA & RA & TA & MIA & UA & RA & TA & MIA \\
\midrule

\multicolumn{17}{c}{\textbf{GRU}} \\ \midrule
\rowcolor{gold!30}
\textsc{Retraining} & 100  &  66.17  &  54.8  &  35.9  &  100  &  72.61  &  60.85  &  36.15  &  100  &  52.13  &  41.21  &  25.43  &  100  &  50.56  &  37.58  &  23.71 \\
\textsc{Finetuning} & 44.73  &  84.96  &  64.98  &  47.99  &  38.12  &  85.47  &  65.2  &  46.58  &  25.82  &  84.88  &  65.34  &  46.73  &  16.19  &  84.88  &  65.48  &  46.12 \\
\textsc{NegGrad} & 79.15  &  83.94  &  62.56  &  48.02  &  58.58  &  84.96  &  64.7  &  45.99  &  45.64  &  83.14  &  62.28  &  46.01  &  41.47  &  83.46  &  62.42  &  45.41 \\
\textsc{NegGrad+} & 56.83  &  84.63  &  64.27  &  48.04  &  54.73  &  85.23  &  65.41  &  46.08  &  46.59  &  85.52  &  64.13  &  46.16  &  37.29  &  85.36  &  63.99  &  45.55 \\
\textsc{Bad-T} & 47.88  &  84.08  &  64.84  &  47.98  &  38.24  &  84.38  &  65.34  &  46.46  &  28.81  &  83.95  &  64.41  &  46.42  &  21.73  &  83.96  &  64.06  &  45.82 \\
\textsc{SCRUB} & 50.29  &  84.37  &  64.77  &  48.03  &  38.33  &  84.53  &  65.2  &  46.48  &  34.73  &  84.34  &  64.2  &  46.38  &  30.21  &  84.03  &  63.42  &  45.55 \\
\hdashline
\textsc{TraceHiding (Ent.)} & 60.61  &  84.37  &  64.41  &  47.98  &  45.58  &  84.59  &  65.05  &  46.47  &  50.46  &  85.08  &  63.2  &  46.25  &  39.52  &  84.25  &  62.85  &  45.63 \\
\textsc{TraceHiding (C.D.)} & 62.42  &  84.45  &  64.06  &  48.05  &  45.32  &  84.53  &  65.2  &  46.41  &  48.85  &  84.43  &  62.49  &  46.25  &  39.07  &  84.33  &  62.28  &  45.64 \\
\textsc{TraceHiding (Uni.)} & 50.68  &  84.39  &  64.98  &  47.98  &  44.64  &  84.48  &  65.27  &  46.31  &  46.16  &  85.03  &  63.2  &  46.3  &  42.05  &  84.22  &  62.28  &  45.37 \\
\multicolumn{17}{c}{\textbf{LSTM}} \\ \midrule
\rowcolor{gold!30}
\textsc{Retraining} & 100  &  62.36  &  50.46  &  29.48  &  100  &  66.42  &  53.67  &  32  &  100  &  31.08  &  26.05  &  17.08  &  100  &  31.62  &  23.56  &  11.03 \\
\textsc{Finetuning} & 42.84  &  79.64  &  60.64  &  46.46  &  38.24  &  80.49  &  61.14  &  42.74  &  28.7  &  79.46  &  60.57  &  43.95  &  21.33  &  79.11  &  61  &  43.6 \\
\textsc{NegGrad} & 43.68  &  79.27  &  60.57  &  46.17  &  42.8  &  79.78  &  60.36  &  42.45  &  48.53  &  78.18  &  58.65  &  43.36  &  41.37  &  78.05  &  59.64  &  42.95 \\
\textsc{NegGrad+} & 45.63  &  79.27  &  60.57  &  46.24  &  44.48  &  80.01  &  60.57  &  42.42  &  39.97  &  79.87  &  60  &  43.76  &  33.27  &  79.18  &  60.93  &  43.1 \\
\textsc{Bad-T} & 40.5  &  79.16  &  60.57  &  46.33  &  37.86  &  79.71  &  60.85  &  42.51  &  29.05  &  78.75  &  60.78  &  43.87  &  22.28  &  78.04  &  60.71  &  43.65 \\
\textsc{SCRUB} & 42.84  &  79.23  &  60.57  &  46.37  &  37.09  &  79.95  &  60.78  &  42.56  &  30  &  78.97  &  60.5  &  43.79  &  23.22  &  78.53  &  60.64  &  43.48 \\
\hdashline
\textsc{TraceHiding (Ent.)} & 43.9  &  79.23  &  60.71  &  46.19  &  39.61  &  80.08  &  60.93  &  42.5  &  42.15  &  79.08  &  59.22  &  43.48  &  43.26  &  78.79  &  59.15  &  43.06 \\
\textsc{TraceHiding (C.D.)} & 43.35  &  79.31  &  60.57  &  45.92  &  41.39  &  79.9  &  60.78  &  41.77  &  39.18  &  79.47  &  59.5  &  43.67  &  48.86  &  78.85  &  56.87  &  43.4 \\
\textsc{TraceHiding (Uni.)} & 43.9  &  79.33  &  60.28  &  46.38  &  40.62  &  79.97  &  60.85  &  41.91  &  39.69  &  79.32  &  59.64  &  43.3  &  51.65  &  78.8  &  57.79  &  43.11 \\
\multicolumn{17}{c}{\textbf{BERT}} \\ \midrule
\rowcolor{gold!30}
\textsc{Retraining} & 100  &  86.93  &  65.12  &  42.88  &  100  &  87.3  &  67.76  &  36.9  &  100  &  86.52  &  59.36  &  39.18  &  100  &  84.32  &  56.51  &  39.8 \\
\textsc{Finetuning} & 63.07  &  87.69  &  66.98  &  48.59  &  49.69  &  87.89  &  66.9  &  48.28  &  40.6  &  88.17  &  65.27  &  47.78  &  36.8  &  88.65  &  64.98  &  47.1 \\
\textsc{NegGrad} & 93.14  &  85.85  &  63.35  &  47.67  &  78.3  &  87.15  &  64.63  &  48.36  &  79.8  &  69.52  &  50.25  &  45.49  &  88.9  &  59.6  &  41.21  &  45.89 \\
\textsc{NegGrad+} & 95.43  &  86.68  &  63.84  &  48.39  &  77.81  &  87.04  &  65.05  &  47.69  &  66.98  &  81.43  &  57.86  &  45.82  &  83.23  &  77.66  &  52.81  &  45.16 \\
\textsc{Bad-T} & 70.07  &  86.26  &  63.35  &  48.64  &  69.22  &  86.52  &  65.05  &  48.33  &  82.52  &  83.34  &  54.66  &  47.06  &  83.41  &  84.01  &  52.67  &  46.89 \\
\textsc{SCRUB} & 89  &  86.61  &  63.2  &  48.28  &  76.01  &  84.9  &  64.34  &  47.39  &  71.6  &  81.62  &  56.87  &  46.25  &  77.86  &  78.78  &  53.38  &  45.48 \\
\hdashline
\textsc{TraceHiding (Ent.)} & 99.13  &  83.17  &  61.14  &  48.64  &  100  &  80.31  &  60.93  &  47.43  &  93.77  &  75.23  &  52.46  &  45.07  &  93.86  &  74.93  &  48.68  &  44.44 \\
\textsc{TraceHiding (C.D.)} & 74.61  &  86.31  &  64.41  &  49.32  &  83.11  &  85.62  &  65.05  &  48.11  &  78.11  &  85.2  &  58.65  &  47.05  &  75.06  &  86.01  &  56.65  &  45.9 \\
\textsc{TraceHiding (Uni.)} & 92.86  &  86.36  &  62.28  &  49.13  &  90.59  &  86.77  &  64.91  &  47.34  &  70.27  &  83.01  &  57.72  &  47.01  &  68.99  &  84.81  &  58.36  &  46.03 \\
\multicolumn{17}{c}{\textbf{ModernBERT}} \\ \midrule
\rowcolor{gold!30}
\textsc{Retraining} & 100  &  83.5  &  60  &  26.73  &  100  &  82.35  &  62.35  &  16.46  &  100  &  81.12  &  53.67  &  35.57  &  99.84  &  77.29  &  48.9  &  25.01 \\
\textsc{Finetuning} & 73.87  &  81.75  &  61.85  &  30.19  &  73.51  &  83.01  &  61.14  &  12.62  &  51.11  &  83.32  &  61.35  &  39.52  &  51.97  &  84.02  &  59.15  &  24.73 \\
\textsc{NegGrad} & 92.97  &  80.21  &  58.65  &  28.73  &  78.17  &  82.03  &  61.42  &  21.31  &  62.5  &  81.77  &  57.22  &  45.87  &  67.82  &  81.1  &  54.31  &  43.45 \\
\textsc{NegGrad+} & 93.16  &  81.59  &  58.86  &  23.07  &  77.53  &  82.31  &  61.49  &  20.51  &  61.41  &  82.89  &  58.22  &  29.14  &  62.23  &  83.83  &  57.15  &  44.07 \\
\textsc{Bad-T} & 70.59  &  80.89  &  58.72  &  16.28  &  67.86  &  81.46  &  60.5  &  30.77  &  63.69  &  79.98  &  55.16  &  33.28  &  57.87  &  81.92  &  54.59  &  44.69 \\
\textsc{SCRUB} & 90.74  &  80.88  &  58.79  &  47  &  76.09  &  82.34  &  61.07  &  21.1  &  52.72  &  80.99  &  59.07  &  28.64  &  66.37  &  81.09  &  54.95  &  34.4 \\
\hdashline
\textsc{TraceHiding (Enr.)} & 93.12  &  80.72  &  59  &  47.63  &  73.3  &  80.4  &  60.07  &  37.34  &  85.76  &  79.1  &  54.59  &  43.73  &  85.53  &  80.46  &  50.96  &  38.32 \\
\textsc{TraceHiding (C.D.)} &  92.43  &  81.2  &  59.22  &  28.08  &  68.74  &  80.85  &  60  &  29.85  &  72.19  &  79.94  &  55.44  &  38.11  &  66.42  &  80.55  &  53.81  &  26.97 \\
\textsc{TraceHiding (Uni.)} & 91.18  &  80.37  &  59.29  &  23.72  &  67.38  &  80.96  &  61.07  &  46.02  &  65.11  &  81.69  &  56.65  &  38.1  &  77.75  &  81.77  &  52.67  &  38.69 \\
\multicolumn{17}{c}{\textbf{GCN-TULHOR}} \\ \midrule
\rowcolor{gold!30}
\textsc{Retraining} & 11.94  &  90.36  &  65.84  &  49.24  &  16.62  &  90.97  &  65.91  &  48.84  &  15.27  &  91.08  &  66.19  &  48.15  &  11.13  &  90.56  &  65.98  &  47.6 \\
\textsc{Finetuning} & 45.65  &  93.03  &  65.84  &  49.23  &  53.8  &  92.74  &  65.05  &  48.74  &  41.81  &  93.56  &  64.06  &  47.85  &  41.45  &  93.62  &  64.06  &  47.7 \\
\textsc{NegGrad} & 100  &  24.76  &  21  &  39.48  &  100  &  6.14  &  5.62  &  30.2  &  100  &  5.61  &  4.77  &  32.67  &  100  &  5.75  &  4.63  &  27.58 \\
\textsc{NegGrad+} & 100  &  33.98  &  28.61  &  41  &  98.9  &  27.68  &  26.05  &  31.63  &  100  &  37.05  &  30.11  &  36.66  &  100  &  42.91  &  32.31  &  35.4 \\
\textsc{Bad-T} & 92.61  &  82.73  &  59.64  &  42.35  &  89.84  &  79.56  &  58.15  &  42.56  &  91.3  &  77.71  &  52.46  &  42.53  &  89.3  &  78.11  &  53.67  &  39.64 \\
\textsc{SCRUB} & 97.07  &  69.24  &  55.09  &  41.64  &  96.79  &  61.05  &  50.32  &  41.68  &  100  &  61.65  &  45.2  &  40.05  &  99.9  &  55.93  &  39.64  &  36.86 \\
\hdashline
\textsc{TraceHiding (Ent.)} & 85.15  &  88.8  &  61.49  &  47.48  &  77.7  &  88.88  &  63.7  &  46.61  &  78.87  &  86.01  &  55.44  &  44.53  &  86.24  &  84.8  &  53.88  &  42.57 \\
\textsc{TraceHiding (C.D.)} & 78.09  &  89.32  &  61.71  &  48.62  &  66.83  &  89.95  &  64.77  &  47.96  &  64.76  &  88.45  &  58.51  &  45.34  &  63.22  &  89.36  &  57.37  &  45.14 \\
\textsc{TraceHiding (Uni.)} & 79.2  &  88.73  &  61  &  48.37  &  71.43  &  89.29  &  63.84  &  48.01  &  76.6  &  89.26  &  57.58  &  44.22  &  75.21  &  89.09  &  56.58  &  44.67 \\
\bottomrule
\end{tabular}
}
\end{table}

\begin{table}[!htbp]
\centering
\caption[The results for the HO-NYC dataset using uniform sampling]{The results for the HO-NYC dataset using uniform sampling. All numbers are presented as percentages.}\label{tab:results_NYC_uniform_sampling}

\resizebox{\textwidth}{!}{
\begin{tabular}{lcccccccccccccccc}
\toprule
\rowcolor{lightgray}
\multicolumn{17}{@{}c@{}}{\textbf{HO‑NYC Dataset}} \\
\midrule
Sample Size & \multicolumn{4}{c}{1 \%} & \multicolumn{4}{c}{5 \%} & \multicolumn{4}{c}{10 \%} & \multicolumn{4}{c}{20 \%} \\
\cmidrule(lr){2-5} \cmidrule(lr){6-9} \cmidrule(lr){10-13} \cmidrule(lr){14-17}
Methods & UA & RA & TA & MIA & UA & RA & TA & MIA & UA & RA & TA & MIA & UA & RA & TA & MIA \\
\midrule

\multicolumn{17}{c}{\textbf{GRU}} \\ \midrule
\rowcolor{gold!30}
\textsc{Retraining} & 100  &  89.41  &  75.42  &  46.98  &  100  &  87.43  &  72.22  &  42.42  &  100  &  81.72  &  65.22  &  40.5  &  100  &  80.84  &  58.67  &  39.72 \\
\textsc{Finetuning} & 15.64  &  92.18  &  77.55  &  49.61  &  15.22  &  92.82  &  77.54  &  47.6  &  18.22  &  93.15  &  77.24  &  43.67  &  14.79  &  92.86  &  77.21  &  40.24 \\
\textsc{NegGrad} & 32.77  &  87.46  &  74.24  &  49.38  &  34.1  &  87.18  &  73.61  &  47.08  &  29.92  &  87.49  &  73.28  &  43.06  &  28.85  &  86.35  &  71.19  &  39.68 \\
\textsc{NegGrad+} & 23.64  &  87.67  &  74.64  &  49.35  &  31.69  &  88.06  &  74.15  &  47.15  &  27.65  &  88.7  &  74.06  &  43.14  &  27.9  &  89  &  72.74  &  39.8 \\
\textsc{Bad-T} & 26.07  &  87.61  &  74.57  &  49.42  &  32.13  &  87.34  &  73.66  &  47.11  &  31.78  &  87.81  &  73.02  &  43.04  &  31.29  &  87.08  &  71.12  &  39.56 \\
\textsc{SCRUB} & 26.8  &  87.66  &  74.59  &  49.47  &  28.99  &  87.67  &  74.1  &  47.26  &  33.41  &  88.16  &  73.15  &  43.07  &  33.98  &  87.54  &  70.55  &  39.72 \\
\hdashline
\textsc{TraceHiding (Ent.)} & 33.38  &  87.68  &  74.43  &  49.37  &  31.08  &  87.53  &  73.9  &  47.23  &  38.88  &  88.25  &  72.46  &  42.95  &  45.71  &  86.66  &  68.17  &  39.63 \\
\textsc{TraceHiding (C.D.)} & 29.99  &  87.61  &  74.55  &  49.4  &  42.78  &  87.41  &  73.13  &  46.88  &  37.36  &  88.12  &  72.69  &  43.29  &  54.47  &  86.2  &  66.54  &  39.52 \\
\textsc{TraceHiding (Uni.)} & 33.9  &  87.73  &  74.44  &  49.35  &  36.81  &  87.56  &  73.55  &  47.13  &  38.57  &  88.05  &  72.57  &  42.8  &  43.57  &  87.06  &  69.16  &  39.86 \\
\multicolumn{17}{c}{\textbf{LSTM}} \\ \midrule
\rowcolor{gold!30}
\textsc{Retraining} & 100  &  91.23  &  76.35  &  48.4  &  100  &  87.34  &  71.98  &  42.25  &  100  &  85.96  &  66.54  &  40.62  &  100  &  66.33  &  48.72  &  36.24 \\
\textsc{Finetuning} & 13.12  &  93.14  &  79.63  &  49.59  &  13.05  &  93.79  &  79.61  &  47.73  &  14.31  &  93.89  &  79.64  &  44.76  &  12.07  &  93.84  &  79.19  &  39.91 \\
\textsc{NegGrad} & 25.28  &  90.58  &  77.42  &  49.35  &  24.77  &  90.47  &  77.05  &  47.42  &  26.74  &  90.62  &  76.42  &  44.39  &  24.17  &  90.02  &  75.28  &  39.44 \\
\textsc{NegGrad+} & 22.24  &  90.74  &  77.6  &  49.35  &  22.84  &  90.99  &  77.56  &  47.47  &  24.71  &  91.44  &  77.14  &  44.31  &  23.96  &  91.56  &  76.15  &  39.41 \\
\textsc{Bad-T} & 19.3  &  90.68  &  77.63  &  49.59  &  24.03  &  90.84  &  77.2  &  47.39  &  24.35  &  90.91  &  76.51  &  44.33  &  24.42  &  90.78  &  75  &  39.26 \\
\textsc{SCRUB} & 25.53  &  90.69  &  77.46  &  49.57  &  26.77  &  90.77  &  77.1  &  47.37  &  31.48  &  90.98  &  76.08  &  44.28  &  28.15  &  90.62  &  74.83  &  39.11 \\
\hdashline
\textsc{TraceHiding (Ent.)} & 28.05  &  90.69  &  77.6  &  49.5  &  28.66  &  90.71  &  77.02  &  47.33  &  34.86  &  90.94  &  75.64  &  44.51  &  36.88  &  90.26  &  73.13  &  38.91 \\
\textsc{TraceHiding (C.D.)} & 26.16  &  90.71  &  77.64  &  49.54  &  26.91  &  90.71  &  77.14  &  47.34  &  36.17  &  90.88  &  75.6  &  44.1  &  39.62  &  90.29  &  72.7  &  38.68 \\
\textsc{TraceHiding (Uni.)} & 37.8  &  90.66  &  77.32  &  49.48  &  36.8  &  90.7  &  76.79  &  47.22  &  28.05  &  91.06  &  76.32  &  44.38  &  40.77  &  90.03  &  72.63  &  38.97 \\
\multicolumn{17}{c}{\textbf{BERT}} \\ \midrule
\rowcolor{gold!30}
\textsc{Retraining} & 100  &  99.76  &  85.62  &  42.25  &  100  &  99.75  &  82.83  &  37.4  &  100  &  99.62  &  78.22  &  33.35  &  100  &  99.4  &  71.52  &  32.94 \\
\textsc{Finetuning} & 10.58  &  99.35  &  86.33  &  49.99  &  8.38  &  99.37  &  86.2  &  49.73  &  7.39  &  99.44  &  86.11  &  49.23  &  5.96  &  99.45  &  85.42  &  47.81 \\
\textsc{NegGrad} & 46.01  &  98.66  &  84.72  &  50  &  90.69  &  46.42  &  39.62  &  49.44  &  100  &  0.55  &  0.5  &  45.9  &  100  &  1.1  &  0.92  &  38.44 \\
\textsc{NegGrad+} & 42.53  &  99.69  &  85.82  &  49.99  &  63.82  &  91.08  &  76.35  &  49.57  &  94.18  &  78.26  &  61.82  &  48.52  &  98.63  &  71.23  &  51.35  &  47.47 \\
\textsc{Bad-T} & 55.44  &  99.65  &  85.46  &  49.77  &  77.86  &  98.28  &  81.19  &  49.76  &  52.03  &  99.23  &  80.56  &  48.87  &  46.95  &  99.49  &  77.33  &  48.21 \\
\textsc{SCRUB} & 54.15  &  99.42  &  85.55  &  49.49  &  90.32  &  93.99  &  77.69  &  49.79  &  97.86  &  90.4  &  70.36  &  49.27  &  98.98  &  93.94  &  66.22  &  48.89 \\
\hdashline
\textsc{TraceHiding (Ent.)} & 71.95  &  99.41  &  84.99  &  50  &  96.93  &  97.75  &  80.32  &  49.16  &  97.12  &  96.36  &  75  &  48.17  &  98.92  &  97.28  &  69.2  &  48.21 \\
\textsc{TraceHiding (C.D.)} & 63.45  &  99.58  &  85.57  &  49.26  &  79.25  &  99.11  &  82.55  &  49.19  &  89.33  &  99.05  &  77.8  &  47.92  &  91.36  &  98.71  &  70.8  &  46.79 \\
\textsc{TraceHiding (Uni.)} & 68.76  &  99.47  &  85.21  &  49.82  &  94.93  &  98  &  80.58  &  49.39  &  98.25  &  97.84  &  75.63  &  48.59  &  98.36  &  97.33  &  69  &  47.91 \\
\multicolumn{17}{c}{\textbf{ModernBERT}} \\ \midrule
\rowcolor{gold!30}
\textsc{Retraining} & 100  &  99.86  &  81.42  &  38.66  &  100  &  99.81  &  78.5  &  36.57  &  100  &  99.79  &  73.25  &  30.64  &  100  &  99.63  &  66.62  &  27.09 \\
\textsc{Finetuning} & 2.11  &  99.76  &  82.07  &  50  &  2.89  &  99.66  &  81.87  &  48.55  &  2.38  &  99.79  &  81.74  &  45.84  &  2.27  &  99.71  &  80.86  &  41.62 \\
\textsc{NegGrad} & 39.72  &  99.92  &  81.92  &  49.84  &  33.36  &  99.85  &  80.78  &  47.93  &  48.12  &  87.89  &  69.75  &  44.97  &  35.23  &  93.56  &  71.57  &  40.19 \\
\textsc{NegGrad+} & 37.35  &  99.92  &  81.92  &  49.97  &  31.94  &  99.69  &  80.9  &  47.6  &  43.36  &  97.36  &  76.1  &  44.71  &  33.24  &  97.92  &  74.73  &  40.81 \\
\textsc{Bad-T} & 43.31  &  99.93  &  81.81  &  49.76  &  34.41  &  99.93  &  81.26  &  47.62  &  48.92  &  99.91  &  78.01  &  43.77  &  55.37  &  99.88  &  73.49  &  40.08 \\
\textsc{SCRUB} & 36.59  &  99.9  &  82.06  &  49.99  &  47.27  &  98.63  &  78.94  &  48.16  &  47.03  &  97.84  &  75.99  &  45.6  &  76.38  &  93.55  &  64.78  &  44.55 \\
\hdashline
\textsc{TraceHiding (Ent.)} & 51.6  &  99.8  &  81.71  &  49.97  &  23.25  &  99.25  &  80.84  &  48.39  &  80.71  &  96.5  &  71.75  &  45.99  &  94.84  &  96.68  &  65.22  &  45.09 \\
\textsc{TraceHiding (C.D.)} & 54.88  &  99.85  &  81.61  &  49.99  &  42.55  &  99.03  &  79.66  &  47.94  &  60.14  &  98.5  &  75.42  &  45.72  &  88.69  &  97.37  &  66.2  &  42.53 \\
\textsc{TraceHiding (Uni.)} & 40.65  &  99.77  &  81.61  &  49.99  &  39.35  &  99.31  &  80.28  &  47.97  &  76.08  &  97.9  &  72.95  &  45.5  &  94.78  &  96.52  &  65  &  44.65 \\
\multicolumn{17}{c}{\textbf{GCN-TULHOR}} \\ \midrule
\rowcolor{gold!30}
\textsc{Retraining} & 0.57  &  99.65  &  84.82  &  49.98  &  0.57  &  99.66  &  84.75  &  48.22  &  0.33  &  99.64  &  84.63  &  46.02  &  0.31  &  99.63  &  84.63  &  39.84 \\
\textsc{Finetuning} & 18.98  &  99.64  &  83.72  &  49.97  &  19.97  &  99.63  &  83.28  &  47.37  &  18.84  &  99.69  &  82.79  &  43.18  &  19.61  &  99.7  &  80.98  &  40.46 \\
\textsc{NegGrad} & 100  &  0.52  &  0.51  &  38.73  &  100  &  0.78  &  0.74  &  34.85  &  100  &  0.72  &  0.65  &  27.83  &  100  &  1.08  &  0.9  &  22.35 \\
\textsc{NegGrad+} & 100  &  1.43  &  1.51  &  38.73  &  100  &  1.26  &  1.16  &  39.52  &  100  &  1.38  &  1.14  &  33.64  &  100  &  1.58  &  1.29  &  28.99 \\
\textsc{Bad-T} & 99.08  &  89.5  &  75.08  &  43.52  &  99.45  &  80.88  &  66.64  &  40.9  &  98.54  &  79.03  &  63.35  &  37.47  &  96.84  &  80.78  &  61.02  &  38.32 \\
\textsc{SCRUB} & 100  &  59.04  &  49.78  &  42.33  &  100  &  36.35  &  32.11  &  40.96  &  100  &  39.43  &  34.04  &  37.86  &  100  &  38.5  &  29.91  &  32.43 \\
\hdashline
\textsc{TraceHiding (Ent.)} & 68.9  &  97.6  &  80.85  &  47.69  &  90.96  &  94.12  &  76.66  &  41.94  &  91.46  &  92.86  &  71.95  &  37.35  &  95.81  &  91.68  &  65.62  &  35.61 \\
\textsc{TraceHiding (C.D.)} & 64.23  &  98.25  &  82.08  &  47.38  &  72.63  &  95.75  &  78.38  &  41.85  &  71.91  &  94.41  &  74.54  &  37.58  &  76.4  &  94.04  &  69.8  &  36.17 \\
\textsc{TraceHiding (Uni.)} & 79.76  &  97.48  &  80.69  &  46.7  &  88.64  &  94.85  &  77.31  &  41.52  &  89.25  &  92.78  &  72.31  &  36.95  &  93.78  &  92.13  &  66.32  &  35.72 \\
\bottomrule
\end{tabular}
}
\end{table}

\section{Results Under Targeted Sampling} \label{appndx:eval_targeted_sampling}

In this section, we present results from evaluating unlearning methods under a targeted sampling strategy, where users with high informational content are selected for deletion. To identify such users, we compute the entropy of their trajectories based on token level representations, capturing the degree of variability and unpredictability in their behavior. Lower entropy values indicate users whose data contributes more significantly to the model's knowledge. By focusing on these low-entropy users, we simulate a more challenging unlearning scenario where the model must forget influential or informative data. This targeted approach provides insight into the performance of unlearning methods when facing deletion requests that potentially have a larger impact on model utility and generalization.

\begin{table}[!htbp]
\centering
\caption[The results for the HO-Rome dataset using targeted sampling]{The results for the HO-Rome dataset using targeted sampling. All numbers are presented as percentages.}\label{tab:results_rome_targeted_sampling}

\resizebox{\textwidth}{!}{
\begin{tabular}{lcccccccccccccccc}

\toprule
\rowcolor{lightgray}
\multicolumn{17}{@{}c@{}}{\textbf{HO‑Rome Dataset}} \\
\midrule
Sample Size & \multicolumn{4}{c}{1 \%} & \multicolumn{4}{c}{5 \%} & \multicolumn{4}{c}{10 \%} & \multicolumn{4}{c}{20 \%} \\
\cmidrule(lr){2-5} \cmidrule(lr){6-9} \cmidrule(lr){10-13} \cmidrule(lr){14-17}
Methods & UA & RA & TA & MIA & UA & RA & TA & MIA & UA & RA & TA & MIA & UA & RA & TA & MIA \\
\midrule

\multicolumn{17}{c}{\textbf{GRU}} \\ \midrule
\rowcolor{gold!30}
\textsc{Retraining} & 100  &  17.54  &  14.12  &  12.04  &  100  &  28.16  &  20.18  &  14.24  &  100  &  3.23  &  2.89  &  10.72  &  100  &  1.27  &  0.88  &  21.41 \\
\textsc{Finetuning} & 98.1  &  5.7  &  6.05  &  19.76  &  98.42  &  5.8  &  6.11  &  12.64  &  96.77  &  5.98  &  6.17  &  6.64  &  92.69  &  5.66  &  6.14  &  0.95 \\
\textsc{NegGrad} & 100  &  3.18  &  3.19  &  17.99  &  100  &  1.69  &  1.96  &  9.27  &  99.92  &  1.42  &  1.52  &  2.49  &  99.26  &  0.96  &  1.23  &  2.81 \\
\textsc{NegGrad+} & 100  &  3.57  &  3.63  &  13.07  &  100  &  3.64  &  3.16  &  6.12  &  99.67  &  3.87  &  3.57  &  4.57  &  98.52  &  3.11  &  2.84  &  3.55 \\
\textsc{Bad-T} & 98.1  &  5.69  &  5.99  &  19.61  &  99.05  &  5.67  &  5.82  &  13.36  &  96.91  &  5.73  &  5.94  &  7.05  &  95.37  &  5.04  &  5.09  &  5.42 \\
\textsc{SCRUB} & 99.31  &  5.63  &  5.58  &  20.64  &  98.89  &  5.63  &  5.7  &  11.18  &  98.35  &  5.65  &  5.38  &  6.74  &  96.59  &  4.59  &  4.68  &  3.06 \\
\hdashline
\textsc{TraceHiding (Ent.)} & 98.79  &  5.7  &  6.05  &  18.61  &  98.57  &  5.87  &  6.05  &  11.82  &  96.93  &  5.8  &  6.02  &  6.71  &  94.6  &  4.71  &  5.26  &  3.07 \\
\textsc{TraceHiding (C.D.)} & 97.49  &  5.68  &  6.11  &  20.37  &  98.58  &  5.79  &  6.14  &  12.57  &  96.69  &  5.88  &  5.96  &  6.24  &  94.99  &  4.85  &  5.38  &  3.56 \\
\textsc{TraceHiding (Uni.)} & 98.79  &  5.71  &  6.14  &  20.35  &  98.26  &  5.74  &  6.08  &  12.67  &  97.08  &  5.81  &  5.99  &  6.51  &  94.79  &  4.93  &  5.41  &  2.8 \\
\multicolumn{17}{c}{\textbf{LSTM}} \\ \midrule
\rowcolor{gold!30}
\textsc{Retraining} & 100  &  8.66  &  5.56  &  9.69  &  100  &  5.84  &  3.95  &  11.97  &  100  &  0.86  &  0.67  &  26.18  &  100  &  0.95  &  0.7  &  40.94 \\
\textsc{Finetuning} & 97.93  &  11.14  &  6.75  &  21.68  &  86.44  &  11.23  &  6.96  &  16.89  &  92.2  &  11.53  &  7.08  &  10.55  &  87.79  &  11.11  &  6.96  &  7.79 \\
\textsc{NegGrad} & 100  &  8.46  &  5.85  &  10.62  &  99.83  &  5.23  &  3.27  &  14.16  &  99.92  &  3.54  &  2.19  &  5.75  &  99.81  &  2.75  &  0.99  &  6.41 \\
\textsc{NegGrad+} & 100  &  9.47  &  6.14  &  17.47  &  99.84  &  8.99  &  5.47  &  11.6  &  99.85  &  7.94  &  4.77  &  11.05  &  98.86  &  9.08  &  4.5  &  4.36 \\
\textsc{Bad-T} & 95.03  &  11.14  &  6.78  &  22.64  &  95.38  &  10.98  &  6.75  &  18.21  &  94.66  &  11.38  &  6.4  &  11.95  &  93.18  &  10.5  &  6.08  &  14.33 \\
\textsc{SCRUB} & 96.41  &  11.22  &  6.96  &  21.54  &  97.55  &  10.63  &  6.32  &  15.31  &  95.03  &  10.67  &  6.17  &  10.3  &  95.41  &  10.14  &  5.47  &  6.33 \\
\hdashline
\textsc{TraceHiding (Ent.)} & 97.93  &  11.1  &  6.84  &  21.39  &  87.6  &  11.07  &  6.96  &  16.84  &  94.42  &  11.33  &  6.58  &  8.12  &  92.34  &  11.05  &  5.91  &  5.59 \\
\textsc{TraceHiding (C.D.)} & 96.55  &  11.19  &  6.81  &  21.84  &  87.79  &  11.12  &  7.11  &  16.09  &  93.74  &  11.36  &  6.96  &  10.32  &  91.27  &  11.14  &  6.02  &  6.42 \\
\textsc{TraceHiding (Uni.)} & 96.55  &  11.09  &  6.81  &  22.25  &  87.73  &  11.14  &  6.96  &  16.46  &  93.98  &  11.35  &  6.81  &  8.92  &  91.8  &  11.03  &  5.94  &  4.95 \\
\multicolumn{17}{c}{\textbf{BERT}} \\ \midrule
\rowcolor{gold!30}
\textsc{Retraining} & 100  &  88.89  &  39.18  &  28.56  &  100  &  86.7  &  37.37  &  24.4  &  100  &  86.91  &  36.67  &  21.38  &  100  &  81.34  &  32.66  &  20.12 \\
\textsc{Finetuning} & 63.03  &  81.87  &  38.54  &  48.86  &  55.95  &  81.01  &  37.63  &  43.21  &  58.48  &  82.51  &  38.19  &  37.62  &  55.65  &  83.21  &  36.35  &  33.82 \\
\textsc{NegGrad} & 77.28  &  74.93  &  36.75  &  46.45  &  75.64  &  52.88  &  28.36  &  41.34  &  97.42  &  10.98  &  7.16  &  35.09  &  100  &  1.21  &  0.73  &  27.29 \\
\textsc{NegGrad+} & 69.13  &  77.67  &  37.51  &  45.72  &  75.75  &  70.64  &  34.77  &  39.26  &  80.16  &  58  &  30.15  &  32.72  &  79.33  &  50.08  &  26.17  &  30.5 \\
\textsc{Bad-T} & 71.9  &  78.71  &  37.66  &  48.11  &  60.75  &  77.99  &  35.94  &  40.61  &  52.21  &  79.7  &  37.02  &  41.03  &  54.15  &  78.92  &  35.38  &  43.29 \\
\textsc{SCRUB} & 79.46  &  79.87  &  38.27  &  47.53  &  83.21  &  66.19  &  31.73  &  43.57  &  82.9  &  65.24  &  32.02  &  37.01  &  76.68  &  63.94  &  29.24  &  32.93 \\
\hdashline
\textsc{TraceHiding (Ent.)} & 82.58  &  75.71  &  37.28  &  46.19  &  81.45  &  66.26  &  33.33  &  40  &  86.84  &  58.58  &  29.77  &  35.86  &  94.18  &  49  &  23.39  &  30.22 \\
\textsc{TraceHiding (C.D.)} & 84.56  &  79.01  &  37.92  &  47.11  &  77.55  &  76.01  &  36.58  &  39.89  &  85.77  &  72.97  &  34.47  &  33.77  &  82.9  &  71.68  &  31.49  &  30.89 \\
\textsc{TraceHiding (Uni.)} & 91.14  &  78.09  &  37.72  &  47.59  &  83.22  &  73.7  &  35.82  &  38.44  &  80.84  &  72.09  &  34.68  &  34.29  &  77.98  &  70.16  &  31.17  &  30.04 \\
\multicolumn{17}{c}{\textbf{ModernBERT}} \\ \midrule
\rowcolor{gold!30}
\textsc{Retraining} & 100  &  67.59  &  40.03  &  34.75  &  100  &  67.12  &  38.36  &  25.44  &  100  &  62.76  &  37.13  &  24.57  &  100  &  56.81  &  31.52  &  17.6 \\
\textsc{Finetuning} & 66.14  &  59.72  &  37.49  &  40.88  &  61.12  &  59.67  &  35.82  &  33.65  &  63.41  &  59.95  &  36.32  &  30.33  &  65.09  &  60.55  &  35.67  &  26.55 \\
\textsc{NegGrad} & 76.76  &  62.51  &  40.23  &  40.3  &  74.91  &  62.78  &  39.5  &  34.59  &  74.13  &  62.63  &  38.92  &  30.75  &  78.91  &  61.97  &  35.67  &  24.89 \\
\textsc{NegGrad+} & 74.6  &  62.29  &  40.23  &  40.98  &  70.8  &  61.83  &  39.39  &  34.92  &  72.25  &  61.56  &  37.78  &  27.21  &  77.25  &  62.36  &  34.82  &  13.49 \\
\textsc{Bad-T} & 82.38  &  62.61  &  40.29  &  14.22  &  63.24  &  62.96  &  39.21  &  36.29  &  68.97  &  63.23  &  37.4  &  29.99  &  64.8  &  64.72  &  35.47  &  13 \\
\textsc{SCRUB} & 55.8  &  59.38  &  38.65  &  41.72  &  69.03  &  59.64  &  38.89  &  35.6  &  66.33  &  59.42  &  37.31  &  31.35  &  70.26  &  60.44  &  35.58  &  9.56 \\
\hdashline
\textsc{TraceHiding (Ent.)} & 48.27  &  57.52  &  37.49  &  41.92  &  63.65  &  58.81  &  37.81  &  35.29  &  69.27  &  55.27  &  35.35  &  30.09  &  81.05  &  51.4  &  30.99  &  9.51 \\
\textsc{TraceHiding (C.D.)} & 62.54  &  58.47  &  38.65  &  41.9  &  68.23  &  57.56  &  37.4  &  31.49  &  65.31  &  57.72  &  36.35  &  29.88  &  70.97  &  58.5  &  35  &  9.66 \\
\textsc{TraceHiding (Uni.)} & 71.83  &  59.25  &  38.45  &  41.26  &  66.12  &  57.59  &  37.49  &  27.82  &  67.5  &  56.33  &  36.4  &  31.53  &  74.2  &  57.16  &  33.27  &  10.84 \\
\multicolumn{17}{c}{\textbf{GCN-TULHOR}} \\ \midrule
\rowcolor{gold!30}
\textsc{Retraining} & 48.13  &  65.71  &  27.92  &  38.39  &  31.29  &  65.55  &  28.01  &  31.16  &  32.26  &  65.23  &  27.84  &  20.02  &  34.06  &  65.49  &  28.36  &  15.9 \\
\textsc{Finetuning} & 84.78  &  66.86  &  29.36  &  31.57  &  76.03  &  67.85  &  29.33  &  25.93  &  77.52  &  68.92  &  28.86  &  20  &  77.74  &  70.97  &  27.31  &  15.49 \\
\textsc{NegGrad} & 100  &  0.81  &  0.88  &  19.88  &  100  &  0.52  &  0.56  &  12.21  &  100  &  0.51  &  0.5  &  6.21  &  100  &  0.58  &  0.5  &  2.23 \\
\textsc{NegGrad+} & 100  &  0.89  &  0.7  &  22.68  &  100  &  0.75  &  0.67  &  8.77  &  100  &  0.78  &  0.64  &  2.83  &  100  &  0.69  &  0.58  &  3.33 \\
\textsc{Bad-T} & 100  &  30.1  &  16.32  &  31.46  &  97.87  &  23.19  &  11.93  &  22.26  &  97.35  &  19.72  &  11.49  &  18.22  &  94.1  &  22.1  &  12.11  &  14.08 \\
\textsc{SCRUB} & 100  &  7.19  &  4.74  &  23.98  &  100  &  3.57  &  2.84  &  18.55  &  100  &  2.88  &  2.31  &  14.38  &  99.86  &  2.09  &  1.61  &  8.89 \\
\hdashline
\textsc{TraceHiding (Ent.)} & 86.29  &  56.82  &  26.11  &  29.9  &  91.1  &  44.09  &  20.99  &  26.37  &  95.66  &  34.67  &  16.9  &  17.92  &  97.69  &  25.9  &  12.6  &  13.6 \\
\textsc{TraceHiding (C.D.)} & 88.67  &  58.44  &  26.46  &  31.19  &  81.94  &  53.57  &  25.09  &  24.85  &  86.14  &  51.28  &  22.98  &  19.93  &  85.23  &  53.74  &  22.78  &  15.19 \\
\textsc{TraceHiding (Uni.)} & 93.13  &  57.87  &  26.46  &  31.71  &  83.4  &  50.14  &  23.36  &  24.67  &  88.48  &  48.5  &  21.99  &  18.5  &  86.5  &  46.15  &  20.76  &  13.72 \\
\bottomrule
\end{tabular}
}
\end{table}

\begin{table}[!htbp]
\centering
\caption[The results for the HO-Geolife dataset using targeted sampling]{The results for the HO-Geolife dataset using targeted sampling. All numbers are presented as percentages.}\label{tab:results_geolife_targeted_sampling}

\resizebox{\textwidth}{!}{
\begin{tabular}{lcccccccccccccccc}
\toprule
\rowcolor{lightgray}
\multicolumn{17}{@{}c@{}}{\textbf{HO‑Geolife Dataset}} \\
\midrule
Sample Size & \multicolumn{4}{c}{1 \%} & \multicolumn{4}{c}{5 \%} & \multicolumn{4}{c}{10 \%} & \multicolumn{4}{c}{20 \%} \\
\cmidrule(lr){2-5} \cmidrule(lr){6-9} \cmidrule(lr){10-13} \cmidrule(lr){14-17}
Methods & UA & RA & TA & MIA & UA & RA & TA & MIA & UA & RA & TA & MIA & UA & RA & TA & MIA \\
\midrule

\multicolumn{17}{c}{\textbf{GRU}} \\ \hline
\rowcolor{gold!30}
\textsc{Retraining} & 100  &  75.84  &  60.78  &  34.49  &  100  &  66.53  &  53.81  &  37.99  &  100  &  54.74  &  42.35  &  30.34  &  100  &  41.42  &  32.03  &  27.56 \\
\textsc{Finetuning} & 50.86  &  84.7  &  65.27  &  48.15  &  26.93  &  84.84  &  65.27  &  47.83  &  10.61  &  83.3  &  65.55  &  48.87  &  23.22  &  85.93  &  65.55  &  45.24 \\
\textsc{NegGrad} & 68.65  &  83.66  &  64.13  &  48.07  &  54.66  &  84.13  &  63.56  &  47.64  &  27.98  &  81.61  &  63.84  &  48.42  &  40.71  &  85.29  &  64.2  &  44.42 \\
\textsc{NegGrad+} & 68.16  &  83.96  &  64.06  &  48.17  &  49.58  &  84.69  &  64.2  &  47.7  &  19.34  &  83.3  &  65.2  &  48.62  &  38.65  &  86.79  &  65.05  &  44.58 \\
\textsc{Bad-T} & 50.44  &  83.86  &  64.98  &  48.11  &  27.51  &  83.82  &  64.56  &  47.79  &  12.33  &  82.1  &  65.05  &  48.75  &  24.95  &  84.99  &  63.99  &  45.09 \\
\textsc{SCRUB} & 54.06  &  83.82  &  64.91  &  48.15  &  29.83  &  84  &  65.12  &  47.81  &  14.17  &  82.34  &  65.12  &  48.69  &  32.04  &  85.2  &  64.41  &  44.92 \\
\hdashline
\textsc{TraceHiding (Ent.)} & 58.8  &  83.84  &  64.77  &  4.78  &  51.83  &  84.35  &  64.06  &  10.69  &  49.85  &  82.01  &  60.85  &  19.98  &  44.25  &  86.51  &  63.84  &  23.59 \\
\textsc{TraceHiding (C.D.)} & 60.13  &  83.92  &  64.91  &  4.8  &  54.94  &  84.54  &  63.77  &  10.6  &  44.15  &  81.68  &  61.42  &  20.22  &  41.49  &  85.27  &  64.27  &  23.43 \\
\textsc{TraceHiding (Uni.)} & 58.03  &  83.84  &  64.84  &  48.14  &  48.48  &  84.23  &  64.7  &  47.59  &  25.93  &  81.98  &  64.48  &  48.4  &  41.3  &  86.04  &  64.34  &  44.36 \\
\multicolumn{17}{c}{\textbf{LSTM}} \\ \hline
\rowcolor{gold!30}
\textsc{Retraining} & 100  &  65.94  &  55.66  &  31.27  &  100  &  60.07  &  50.46  &  33.5  &  100  &  39.87  &  33.24  &  20.64  &  100  &  33.34  &  26.33  &  13.31 \\
\textsc{Finetuning} & 33.56  &  79.31  &  60.5  &  47.59  &  36.4  &  79.77  &  61.21  &  46.57  &  19.12  &  78.3  &  60.85  &  45.88  &  27.47  &  80.54  &  61.21  &  41.1 \\
\textsc{NegGrad} & 34.05  &  78.89  &  60.36  &  47.66  &  48.2  &  78.61  &  60.36  &  46.2  &  33.69  &  77.55  &  59.79  &  45.81  &  38.15  &  80.2  &  59.57  &  40.76 \\
\textsc{NegGrad+} & 34.05  &  79.01  &  60.5  &  47.64  &  42.26  &  79.18  &  60.71  &  46.36  &  36.99  &  78.12  &  58.51  &  45.8  &  38.06  &  81.18  &  60.07  &  40.98 \\
\textsc{Bad-T} & 33.56  &  78.84  &  60.5  &  47.71  &  36.2  &  78.9  &  60.85  &  46.6  &  20.47  &  77.68  &  60.78  &  45.73  &  27.48  &  79.74  &  60.78  &  41.07 \\
\textsc{SCRUB} & 33.8  &  78.91  &  60.64  &  47.74  &  37.35  &  79.13  &  61  &  46.53  &  20.59  &  77.83  &  60.64  &  46.29  &  29.83  &  80.12  &  60.85  &  41.12 \\
\hdashline
\textsc{TraceHiding (Ent.)} & 44.57  &  78.9  &  60  &  5.43  &  47.03  &  79.12  &  59.79  &  10.16  &  34.06  &  77.76  &  59  &  19.73  &  51.89  &  80.54  &  57.86  &  19.76 \\
\textsc{TraceHiding (C.D.)} & 41.82  &  79.01  &  59.86  &  5.56  &  55.29  &  78.97  &  60  &  10.1  &  38.24  &  77.73  &  58.58  &  19.82  &  48.69  &  80.57  &  57.94  &  19.96 \\
\textsc{TraceHiding (Uni.)} & 32.8  &  78.97  &  60.57  &  47.64  &  46  &  79.12  &  60.64  &  46.47  &  32.4  &  77.65  &  59.72  &  46.2  &  43.04  &  80.56  &  59.72  &  40.91 \\
\multicolumn{17}{c}{\textbf{BERT}} \\ \hline
\rowcolor{gold!30}
\textsc{Retraining} & 100  &  86.05  &  67.19  &  33.2  &  100  &  85.7  &  64.91  &  42.02  &  100  &  83.82  &  58.72  &  41.12  &  100  &  86.44  &  59.79  &  37.87 \\
\textsc{Finetuning} & 46.26  &  87.24  &  66.83  &  49.5  &  44.62  &  87.35  &  65.48  &  48.48  &  24.14  &  85.53  &  65.41  &  49.37  &  37.02  &  89.55  &  66.12  &  47.67 \\
\textsc{NegGrad} & 67.37  &  86.02  &  65.77  &  49.07  &  79.43  &  80.19  &  59.86  &  48.26  &  81.68  &  69.91  &  49.11  &  47.7  &  90.26  &  64.67  &  46.48  &  45.37 \\
\textsc{NegGrad+} & 67.25  &  86.43  &  65.62  &  49.39  &  81.16  &  83.16  &  61.21  &  47.95  &  84.88  &  79.11  &  55.8  &  47.41  &  89.6  &  79.08  &  54.23  &  44.69 \\
\textsc{Bad-T} & 55.55  &  86.2  &  65.27  &  49.54  &  67.34  &  82.8  &  60.71  &  48.46  &  78.39  &  80.95  &  55.02  &  48.01  &  82.63  &  84.04  &  53.81  &  46.41 \\
\textsc{SCRUB} & 65.78  &  86.16  &  65.05  &  49.61  &  70.23  &  82.33  &  60.57  &  47.34  &  74.92  &  79.51  &  55.02  &  48.21  &  92.78  &  78  &  52.31  &  45.3 \\
\hdashline
\textsc{TraceHiding (Ent.)} &  90.8  &  84.76  &  65.2  &  48.91  &  84.4  &  80.36  &  59.72  &  46.99  &  83.93  &  78.71  &  54.73  &  46.95  &  99.32  &  72.97  &  49.89  &  43.37 \\
\textsc{TraceHiding (C.D.)} & 74.41  &  86.14  &  65.62  &  48.67  &  60.74  &  85.74  &  63.63  &  48.7  &  75.79  &  83.07  &  57.15  &  47.96  &  68.88  &  86.05  &  59.36  &  44.63 \\
\textsc{TraceHiding (Uni.)} & 91.82  &  84.93  &  65.2  &  49.01  &  68.03  &  85  &  61.99  &  47.83  &  82.47  &  81.08  &  55.87  &  47.7  &  80.05  &  83.83  &  56.87  &  44.5 \\
\multicolumn{17}{c}{\textbf{ModernBERT}} \\ \hline
\rowcolor{gold!30}
\textsc{Retraining} & 99.76  &  82.9  &  62.56  &  8.49  &  100  &  82.74  &  59.57  &  25.29  &  99.68  &  79.98  &  53.52  &  14.69  &  100  &  81.04  &  53.88  &  27.09 \\
\textsc{Finetuning} & 47.87  &  81.13  &  61.42  &  11.28  &  58.5  &  82.74  &  61.64  &  25.22  &  33.83  &  80.86  &  60  &  34.05  &  44.77  &  84.17  &  61.42  &  32.26 \\
\textsc{NegGrad} & 58.53  &  80.34  &  60.71  &  18.58  &  67.29  &  81.38  &  59.36  &  17.17  &  50.51  &  78.62  &  56.23  &  23.72  &  60.59  &  82.5  &  57.65  &  35.17 \\
\textsc{NegGrad+} & 54.63  &  81.13  &  61.49  &  24.93  &  64.07  &  82.01  &  60.64  &  33.97  &  47.65  &  80.78  &  58.43  &  23.32  &  59.48  &  83.78  &  58.79  &  43.65 \\
\textsc{Bad-T} & 56.64  &  80.81  &  60.78  &  29.42  &  61.76  &  80.49  &  59.22  &  24.69  &  57.8  &  78.7  &  55.87  &  29.28  &  53.26  &  81.63  &  56.87  &  41.85 \\
\textsc{SCRUB} & 61.61  &  80.09  &  60.85  &  10.36  &  61.27  &  80.96  &  59.86  &  24.76  &  63.67  &  78.81  &  55.3  &  41.54  &  62.79  &  82.78  &  57.37  &  26.96 \\
\hdashline
\textsc{TraceHiding (Ent.)} & 58.48  &  79.01  &  60.64  &  39.56  &  65.08  &  79.21  &  58.86  &  24.14  &  70.24  &  77.61  &  54.66  &  22.39  &  69.07  &  80.73  &  57.51  &  40.74 \\
\textsc{TraceHiding (C.D.)} & 45.39  &  79  &  61  &  39.69  &  54.41  &  80.96  &  60.71  &  41.15  &  52.33  &  77.59  &  56.16  &  22.42  &  61.04  &  82.47  &  58.79  &  28.3 \\
\textsc{TraceHiding (Uni.)} & 47.41  &  79.6  &  60.28  &  10.49  &  70.38  &  79.82  &  58.93  &  25.18  &  64.88  &  77.76  &  55.37  &  47.14  &  66.09  &  81.51  &  56.23  &  42.7 \\
\multicolumn{17}{c}{\textbf{GCN-TULHOR}} \\ \midrule
\rowcolor{gold!30}
\textsc{Retraining} & 9.4  &  90.35  &  66.19  &  49.9  &  8.46  &  90.16  &  65.69  &  49.4  &  5.26  &  89.77  &  66.12  &  49.1  &  13.89  &  91.04  &  65.77  &  46.63 \\\textsc{Finetuning} & 39.95  &  92.22  &  65.69  &  49.94  &  40.58  &  92.59  &  65.34  &  49.28  &  24.58  &  92.28  &  64.63  &  48.33  &  36.86  &  94.36  &  65.48  &  46.32 \\
\textsc{NegGrad} & 97.5  &  25.77  &  21.71  &  31.34  &  100  &  5.67  &  5.77  &  37.93  &  100  &  7.05  &  6.19  &  34.68  &  100  &  7.53  &  5.98  &  24.45 \\
\textsc{NegGrad+} & 100  &  31.5  &  28.33  &  33.78  &  100  &  36.17  &  32.53  &  33.96  &  100  &  38.44  &  31.6  &  40.05  &  100  &  37.55  &  29.61  &  32.93 \\
\textsc{Bad-T} & 74.77  &  86.21  &  63.49  &  44.83  &  90.63  &  81.13  &  56.87  &  44.89  &  86.61  &  76.97  &  54.02  &  44.09  &  95.9  &  79.17  &  53.52  &  37.7 \\
\textsc{SCRUB} & 89.25  &  70.13  &  55.87  &  43.69  &  100  &  60.09  &  47.47  &  41.96  &  99.51  &  57.92  &  44.2  &  42.82  &  99.88  &  55.94  &  41.64  &  35.21 \\
\hdashline
\textsc{TraceHiding (Ent.)} & 56.11  &  89.09  &  63.35  &  45.29  &  70.82  &  87.08  &  60.36  &  46.7  &  73.24  &  86.13  &  56.09  &  46.15  &  75.78  &  84.4  &  56.87  &  40.48 \\
\textsc{TraceHiding (C.D.)} &  53.99  &  88.6  &  64.2  &  48.69  &  58.18  &  89.33  &  61.57  &  48.17  &  65.49  &  86.77  &  58.36  &  47.65  &  58.43  &  88.78  &  59.86  &  44.07 \\
\textsc{TraceHiding (Uni.)} & 55.21  &  88.7  &  63.49  &  49.8  &  68.46  &  88.45  &  61.49  &  48.32  &  75.76  &  86.81  &  58.08  &  47.17  &  71.02  &  88.93  &  58.93  &  42.68 \\
\bottomrule
\end{tabular}
}
\end{table}

\begin{table}[!htbp]
\centering
\caption[The results for the HO-NYC dataset using targeted sampling]{The results for the HO-NYC dataset using targeted sampling. All numbers are presented as percentages.}\label{tab:results_NYC_targeted_sampling}

\resizebox{\textwidth}{!}{
\begin{tabular}{lcccccccccccccccc}
\toprule
\rowcolor{lightgray}
\multicolumn{17}{@{}c@{}}{\textbf{HO‑NYC Dataset}} \\
\midrule
Sample Size & \multicolumn{4}{c}{1 \%} & \multicolumn{4}{c}{5 \%} & \multicolumn{4}{c}{10 \%} & \multicolumn{4}{c}{20 \%} \\
\cmidrule(lr){2-5} \cmidrule(lr){6-9} \cmidrule(lr){10-13} \cmidrule(lr){14-17}
Methods & UA & RA & TA & MIA & UA & RA & TA & MIA & UA & RA & TA & MIA & UA & RA & TA & MIA \\
\midrule

\multicolumn{17}{c}{\textbf{GRU}} \\ \midrule
\rowcolor{gold!30}
\textsc{Retraining} & 100  &  88.83  &  75.07  &  45.92  &  100  &  84.15  &  68.96  &  43.72  &  100  &  84.54  &  66.36  &  41.76  &  100  &  78.79  &  56  &  39.73 \\
\textsc{Finetuning} & 15.51  &  91.97  &  77.28  &  49.95  &  13.74  &  92.59  &  77.33  &  48.11  &  14.43  &  92.8  &  77.02  &  45.04  &  15.18  &  93.18  &  76.96  &  39.78 \\
\textsc{NegGrad} & 31.33  &  87.45  &  74.52  &  49.9  &  31.1  &  87.09  &  73.39  &  47.84  &  30.97  &  87.02  &  72.49  &  44.63  &  28.17  &  86.56  &  71.11  &  39.37 \\
\textsc{NegGrad+} & 36.67  &  87.64  &  74.64  &  49.91  &  28.3  &  88.01  &  74.01  &  47.9  &  30.95  &  88.21  &  73.48  &  44.68  &  27.19  &  89.06  &  72.76  &  39.43 \\
\textsc{Bad-T} & 26.65  &  87.45  &  74.52  &  49.9  &  33.25  &  87.2  &  73.02  &  47.76  &  33.59  &  87.31  &  72.39  &  44.77  &  35.34  &  87.2  &  70.71  &  39.13 \\
\textsc{SCRUB} & 29.28  &  87.55  &  74.49  &  49.87  &  34.13  &  87.52  &  73.41  &  48.02  &  34.52  &  87.61  &  72.61  &  44.7  &  33.34  &  87.72  &  70.95  &  39.18 \\
\hdashline
\textsc{TraceHiding (Ent.)} & 37.16  &  87.66  &  74.46  &  49.92  &  42.01  &  87.3  &  72.83  &  47.9  &  41.32  &  87.4  &  71.7  &  44.73  &  38.38  &  86.97  &  69.42  &  39.04 \\
\textsc{TraceHiding (C.D.)} & 33.98  &  87.7  &  74.6  &  49.87  &  38.22  &  87.34  &  72.97  &  47.89  &  43.38  &  87.39  &  71.56  &  44.65  &  36.91  &  87.47  &  69.93  &  39.06 \\
\textsc{TraceHiding (Uni.)} & 31.23  &  87.61  &  74.46  &  49.91  &  36.96  &  87.45  &  73.1  &  47.97  &  36.65  &  87.47  &  72.31  &  44.85  &  43.86  &  87.27  &  68.55  &  38.79 \\
\multicolumn{17}{c}{\textbf{LSTM}} \\ \midrule
\rowcolor{gold!30}
\textsc{Retraining} & 100  &  91.78  &  76.31  &  47.19  &  100  &  88.41  &  71.1  &  42.18  &  100  &  85.84  &  66.33  &  40.46  &  100  &  84.21  &  58.23  &  39.79 \\
\textsc{Finetuning} & 6.14  &  93.6  &  79.97  &  49.49  &  9.46  &  93.48  &  79.94  &  48.36  &  11.72  &  93.65  &  79.5  &  44.75  &  12.86  &  93.75  &  78.99  &  40.71 \\
\textsc{NegGrad} & 20.9  &  90.54  &  77.44  &  49.2  &  26.35  &  90.07  &  76.65  &  47.87  &  23.46  &  90.21  &  76.33  &  44.33  &  24.24  &  90.16  &  74.91  &  40.07 \\
\textsc{NegGrad+} & 18.39  &  90.7  &  77.82  &  49.2  &  28.15  &  90.83  &  76.83  &  47.94  &  24.49  &  91.23  &  76.74  &  44.25  &  24.78  &  91.81  &  75.71  &  40.04 \\
\textsc{Bad-T} & 11.92  &  90.62  &  77.76  &  49.32  &  28.09  &  90.5  &  76.6  &  47.94  &  24.46  &  90.55  &  76.33  &  44.12  &  25.87  &  90.87  &  74.8  &  39.75 \\
\textsc{SCRUB} & 19.47  &  90.72  &  77.65  &  49.27  &  24.75  &  90.45  &  77.06  &  48.14  &  26.46  &  90.54  &  75.86  &  44.35  &  31.06  &  90.84  &  74.01  &  39.81 \\
\hdashline
\textsc{TraceHiding (Ent.)} & 25.4  &  90.64  &  77.52  &  49.26  &  34.09  &  90.31  &  76.11  &  48.13  &  34.65  &  90.29  &  74.97  &  44.25  &  39.56  &  90.48  &  72.38  &  39.78 \\
\textsc{TraceHiding(C.D)} & 28.53  &  90.62  &  77.58  &  49.28  &  38.41  &  90.28  &  75.94  &  47.9  &  29.81  &  90.45  &  75.67  &  44.25  &  42.24  &  90.28  &  71.67  &  39.75 \\
\textsc{TraceHiding (Uni.)} & 30.23  &  90.64  &  77.41  &  49.26  &  30.25  &  90.37  &  76.46  &  48.05  &  30.69  &  90.43  &  75.46  &  44.44  &  40.66  &  90.32  &  72.15  &  39.57 \\
\multicolumn{17}{c}{\textbf{BERT}} \\ \midrule
\rowcolor{gold!30}
\textsc{Retraining} & 100  &  99.79  &  85.73  &  41.64  &  100  &  99.56  &  81.74  &  37.82  &  100  &  99.64  &  78.16  &  33.72  &  100  &  99.44  &  69.77  &  32.4 \\
\textsc{Finetuning} & 6.84  &  99.33  &  86.52  &  50  &  6.31  &  99.46  &  86.29  &  49.73  &  5.19  &  99.46  &  86.12  &  49.38  &  6.38  &  99.51  &  85.48  &  29.45 \\
\textsc{NegGrad} & 63.19  &  99.73  &  85.95  &  49.99  &  89.09  &  54.19  &  46.45  &  49.58  &  100  &  1.17  &  0.99  &  45.51  &  100  &  0.49  &  0.37  &  37.5 \\
\textsc{NegGrad+} & 15.79  &  99.79  &  86.62  &  50  &  72.17  &  90.83  &  73.93  &  49.48  &  91.77  &  82.35  &  62.04  &  48.92  &  98  &  74.28  &  51.87  &  48.07 \\
\textsc{Bad-T} & 64.05  &  99.67  &  85.95  &  49.96  &  77.55  &  98.31  &  79.96  &  49.86  &  54.02  &  99.28  &  80.03  &  49.24  &  48.08  &  99.55  &  76.26  &  34.82 \\
\textsc{SCRUB} & 61.54  &  99.45  &  85.72  &  49.99  &  91.33  &  91.41  &  74.14  &  49.68  &  98.66  &  92.89  &  71.52  &  49.25  &  99.42  &  91.47  &  63.23  &  48.72 \\
\hdashline
\textsc{TraceHiding (Ent.)} & 90.78  &  98.77  &  84.45  &  0.99  &  96.66  &  98.89  &  80.24  &  9.28  &  98.93  &  97.79  &  75.72  &  18.36  &  98.88  &  97.47  &  67.59  &  25.56 \\
\textsc{TraceHiding (C.D.)} & 80.24  &  99.17  &  85.03  &  49.99  &  80.91  &  99.03  &  81.35  &  49.52  &  88.56  &  98.94  &  77.74  &  48.31  &  94.62  &  98.7  &  69.27  &  32.98 \\
\textsc{TraceHiding (Uni.)} & 90.78  &  98.77  &  84.45  &  49.99  &  96.66  &  98.89  &  80.24  &  49.54  &  98.93  &  97.79  &  75.72  &  48.43  &  98.88  &  97.47  &  67.59  &  30.03 \\
\multicolumn{17}{c}{\textbf{ModernBERT}} \\ \midrule
\rowcolor{gold!30}
\textsc{Retraining} & 100  &  99.89  &  81.53  &  40.85  &  100  &  99.81  &  76.89  &  36.08  &  100  &  99.74  &  73.05  &  31.21  &  100  &  99.72  &  65.03  &  22.26 \\
\textsc{Finetuning} & 3.15  &  99.74  &  81.8  &  50  &  1.88  &  99.71  &  81.91  &  49.16  &  2.82  &  99.7  &  81.48  &  44.25  &  2.86  &  99.83  &  81.23  &  41.66 \\
\textsc{NegGrad} & 42.44  &  99.9  &  82.09  &  50  &  37.48  &  99.46  &  79.66  &  48.33  &  26.88  &  99.37  &  79.38  &  44.55  &  47.58  &  80.08  &  60.54  &  32.77 \\
\textsc{NegGrad+} & 40.36  &  99.92  &  82.1  &  50  &  34.43  &  99.74  &  80.49  &  48.88  &  36.99  &  98.91  &  77.93  &  43.79  &  38.75  &  96.59  &  72.64  &  24.47 \\
\textsc{Bad-T} & 41.54  &  99.93  &  82.2  &  49.99  &  43.44  &  99.9  &  79.94  &  48.6  &  44.11  &  99.88  &  78.32  &  43.38  &  60.67  &  99.84  &  71.47  &  28.31 \\
\textsc{SCRUB} & 54.14  &  99.82  &  81.81  &  50  &  35.91  &  99.65  &  80.06  &  49.03  &  31.85  &  99.04  &  78.23  &  45.49  &  75.94  &  94.15  &  64.41  &  45.19 \\
\hdashline
\textsc{TraceHiding (Ent.)} & 38.78  &  99.84  &  81.97  &  49.92  &  51.43  &  98.28  &  77.61  &  48.41  &  78.1  &  96.82  &  71.84  &  45.46  &  96.54  &  96.6  &  62.81  &  44.97 \\
\textsc{TraceHiding (C.D.)} & 46.58  &  99.85  &  81.91  &  50  &  40.99  &  99.23  &  79.09  &  48.39  &  67.96  &  98.05  &  73.86  &  44.82  &  89.57  &  97.38  &  64.72  &  30.4 \\
\textsc{TraceHiding (Uni.)} & 45.85  &  99.88  &  81.98  &  50  &  47.52  &  98.93  &  78.85  &  48.8  &  72.95  &  97.32  &  73.15  &  44.94  &  96.29  &  96.65  &  63.15  &  45.23 \\
\multicolumn{17}{c}{\textbf{GCN-TULHOR}} \\ \midrule
\rowcolor{gold!30}
\textsc{Retraining} & 0.22  &  99.65  &  84.72  &  49.99  &  0.56  &  99.66  &  84.82  &  48.16  &  0.41  &  99.62  &  84.68  &  45.49  &  0.45  &  99.65  &  84.72  &  39.6 \\
\textsc{Finetuning} & 21.93  &  99.56  &  83.58  &  49.91  &  17.75  &  99.7  &  82.88  &  47.44  &  19.86  &  99.57  &  82.11  &  44.03  &  19.8  &  99.77  &  81.08  &  39.07 \\
\textsc{NegGrad} & 100  &  0.8  &  0.76  &  43.47  &  100  &  0.9  &  0.8  &  39.24  &  100  &  0.72  &  0.66  &  28  &  98.92  &  0.46  &  0.62  &  16.4 \\
\textsc{NegGrad+} & 100  &  2.82  &  2.61  &  44.67  &  100  &  2.04  &  1.94  &  38.75  &  100  &  1.66  &  1.44  &  35.57  &  100  &  2.37  &  1.85  &  31.64 \\
\textsc{Bad-T} & 99.74  &  91.53  &  77.05  &  44.33  &  99.6  &  79.76  &  65.98  &  40.66  &  98.78  &  79.64  &  63.8  &  39.21  &  97.54  &  81.97  &  58.92  &  38.43 \\
\textsc{SCRUB} & 100  &  59.14  &  51.24  &  44.41  &  99.85  &  39.25  &  34.67  &  41.86  &  100  &  39.02  &  32.94  &  36.93  &  100  &  42.23  &  32.43  &  33.16 \\
\hdashline
\textsc{TraceHiding (Ent.)} & 92.91  &  95.54  &  77.84  &  46.75  &  87.54  &  94.42  &  76.25  &  42.79  &  93.33  &  92.92  &  71.89  &  36.16  &  96.2  &  91.96  &  64.03  &  35.36 \\
\textsc{TraceHiding (C.D.)} & 82.03  &  97.46  &  81.47  &  46.57  &  68.49  &  95.08  &  77.13  &  43.92  &  70  &  94.53  &  74.98  &  37.7  &  78.01  &  94.16  &  68.17  &  35.54 \\
\textsc{TraceHiding (Uni.)} & 88.87  &  96.34  &  79.15  &  47.08  &  85.82  &  94.6  &  76.26  &  42.77  &  91.28  &  92.67  &  71.82  &  36.74  &  94.44  &  92.84  &  64.96  &  34.57 \\
\bottomrule
\end{tabular}
}
\end{table}

\end{document}